\documentclass{article}

\usepackage[preprint,nonatbib]{neurips_2026}

\usepackage[utf8]{inputenc} % allow utf-8 input
\usepackage[T1]{fontenc}    % use 8-bit T1 fonts
\usepackage{hyperref}       % hyperlinks
\usepackage{url}            % simple URL typesetting
\usepackage{booktabs}       % professional-quality tables
\usepackage{amsfonts}       % blackboard math symbols
\usepackage{nicefrac}       % compact symbols for 1/2, etc.
\usepackage{microtype}      % microtypography
\usepackage{xcolor}         % colors
\usepackage{tabularx}
\usepackage{listings}
\usepackage{amsmath} 
\usepackage[most]{tcolorbox}
\usepackage{fvextra}
\usepackage{array}
\makeatletter
\tcbset{
  promptboxstyle/.style={}
}

\newcommand{\promptboxsetup}[1]{%
  \tcbset{promptboxstyle/.style={#1}}%
}

\newenvironment{promptbox}
  {\VerbatimEnvironment
   \begin{tcolorbox}[
     breakable,
     auto outer arc,
     boxrule=0.5pt,
     colback=TableLightBlue,
     colframe=TableDarkBlue,
     left=1mm,
     right=1mm,
     top=1mm,
     bottom=1mm,
     promptboxstyle
   ]
   \begin{Verbatim}[
     breaklines=true,
     breakanywhere=true,
     breaksymbolleft={},
     formatcom=\fontsize{8pt}{8pt}\selectfont
   ]}
  {\end{Verbatim}
   \end{tcolorbox}
   \tcbset{promptboxstyle/.style={}}}
\makeatother

\usepackage{multirow}
\usepackage{CJKutf8}
\usepackage{subfigure}
\usepackage{caption}
\usepackage{float}
\usepackage{graphicx}
\usepackage{amsmath}
\usepackage{array}
\usepackage{makecell}
\usepackage{amsfonts} 
\usepackage{booktabs}
\usepackage{enumitem}
\usepackage[table,xcdraw]{xcolor}
\usepackage[normalem]{ulem}
\definecolor{TableBlue}{HTML}{E4EAF7}
\definecolor{TableLightBlue}{HTML}{F1F7FB}
\definecolor{TableDarkBlue}{HTML}{3985C5}
\definecolor{TableGreen}{HTML}{ECF6E5}
\definecolor{TableLightGreen}{HTML}{F2F8EC}
\definecolor{TableDarkGreen}{HTML}{80B442}
\definecolor{TablePurple}{HTML}{E5E5FA}
\definecolor{TableLightPurple}{HTML}{F1F1FC}
\definecolor{TableDarkPurple}{HTML}{7E31C5}
\definecolor{TableYellow}{HTML}{FEFEDA}
\definecolor{TableOrange}{HTML}{FFE6C8}
\definecolor{TableRed}{HTML}{ffe6e4}

\usepackage[most]{tcolorbox}
\usepackage{tabularx}
\usepackage{booktabs}
\usepackage{array}
\usepackage{ragged2e}
\usepackage{xcolor}
\usepackage{enumitem}

\newcolumntype{Y}{>{\RaggedRight\arraybackslash}X}

\definecolor{FailBlue}{RGB}{232,240,254}
\definecolor{FailBorder}{RGB}{91,134,229}
\definecolor{FailGray}{RGB}{245,246,250}
\definecolor{FailGold}{RGB}{255,244,214}
\definecolor{FailRed}{RGB}{210,60,60}
\definecolor{FailGreen}{RGB}{46,125,50}

\newcommand{\scorepill}[1]{%
  \tcbox[
    on line,
    boxrule=0pt,
    arc=2mm,
    left=1mm,right=1mm,top=0.4mm,bottom=0.4mm,
    colback=FailGold,
    colframe=FailGold
  ]{\textbf{#1}}%
}

\newtcolorbox{failurecasebox}[2][]{
  enhanced,
  breakable,
  colback=white,
  colframe=FailBorder,
  boxrule=0.6pt,
  arc=1.5mm,
  left=1.5mm,right=1.5mm,top=1mm,bottom=1mm,
  title={\textbf{Failure Case #2} \hfill \textsc{DI Track}},
  coltitle=black,
  colbacktitle=FailBlue,
  fonttitle=\small,
  #1
}

\newtcolorbox{innerblock}[1][]{
  enhanced,
  breakable,
  colback=FailGray,
  colframe=white,
  boxrule=0pt,
  arc=1mm,
  left=1mm,right=1mm,top=0.8mm,bottom=0.8mm,
  #1
}
\newcommand{\greencheck}{{\color{green!70!black}$\boldsymbol{\checkmark}$}}
\newcommand{\redcross}{{\color{red!70!black}$\boldsymbol{\times}$}}
\definecolor{ourrow}{RGB}{230, 240, 255}

\title{FinReasoning: A Hierarchical Benchmark for Reliable Financial Research Reporting}

\begin{document}

\author{
\mdseries\large
Yiyun Zhu$^{1,*}$, Yidong Jiang$^{1,*}$, Ziwen Xu$^{1}$, Yinsheng Yao$^{1}$, Dawei Cheng$^{1,\dag}$, 
\\
\large
Jinru Ding$^{2}$ and Jie Xu$^{2}$
\\
\small
$^1$ School of Computer Science and Technology, Tongji University, Shanghai, China \\
$^2$ Shanghai Artificial Intelligence Laboratory, Shanghai, China
\\
\small
\{2533994, dcheng\}@tongji.edu.cn \quad
\{dingjinru, xujie\}@pjlab.org.cn
}

\maketitle

\begingroup
\renewcommand{\thefootnote}{}
\footnotetext{$*$ These authors contributed equally to this work.}
\footnotetext{$\dag$ Corresponding author.}
\endgroup

\begin{abstract}
Large language models (LLMs) are increasingly deployed in financial research workflows, where their role is evolving from single-model assistance for human analysts toward autonomous collaboration among multiple agents.
Yet real-world deployments still expose factual errors, numerical inconsistencies, and shallow analysis, which can distort assessments of corporate fundamentals and trigger severe economic losses. 
While existing benchmarks have begun to evaluate such failures, they score all aspects of the generated analysis in one pass, failing to  distinguish whether a model fails at foundational stages like auditing and correction, or underperforms at generating research-grade insights. Consequently, it obscures capability bottlenecks and the specialized strengths essential for multi-agent role assignment.
To address these gaps, we introduce FinReasoning, a hierarchical benchmark that decomposes the core capabilities of financial research into semantic consistency, data alignment, and deep insight. We further propose a fine-grained evaluation framework that strengthens hallucination-correction assessment and incorporates a 12-indicator rubric for core analytical skills. 
FinReasoning reveals clear capability stratification across model types.
Closed-source models (like Doubao-Seed-1.8) perform strongly overall and are better suited for core reasoning agents in multi-agent financial systems; open-source general models (like Qwen3-235B) show clear capability divergence and consistently underperform on Semantic Consistency, making them less suited for quality-sensitive generation tasks; financial-domain models (like Fin-R1) generate moderate insights but lack foundational auditing skills. Our work has already been deployed in pilot tests across several real-world scenarios. The resource is available at https://github.com/TongjiFinLab/FinReasoning.
\end{abstract}

\section{Introduction}
%The core paradigm of large language models is shifting from writing assistance to more complex decision support and deep business applications. In financial workflows, LLMs are increasingly integrated into automated generation of research reports. Institutions such as Union Bank of Switzerland \cite{redmond2025ubsai} and Citadel \cite{reuters2025citadelai} have begun using LLMs to draft equity research reports and market analyses, allowing analysts to process data, develop viewpoints, and shape narrative structures far more efficiently. 

Financial LLM applications are undergoing a paradigm shift from single-model assistance to autonomous multi-agent collaboration. Institutions such as Union Bank of Switzerland \cite{redmond2025ubsai} and Citadel \cite{reuters2025citadelai} have integrated LLMs into research pipelines for drafting equity reports and market analyses, while recent agent frameworks enable autonomous portfolio management and trading. Meanwhile, emerging multi-agent financial systems \cite{kou2024automate,yu2025finmem,li2025hedgeagents} increasingly distribute responsibilities across specialized agents for strategy generation, risk assessment, data verification, and decision synthesis.
% 加一些学术上的citation，例如ACL上的关于金融的以及多智能体的，体现在学术和工业上都很重要
% Therefore, LLMs are shifting from auxiliary analysis tools to primary generators of research content.

\begin{figure}[h]
    \centering
    \includegraphics[width=1\textwidth]{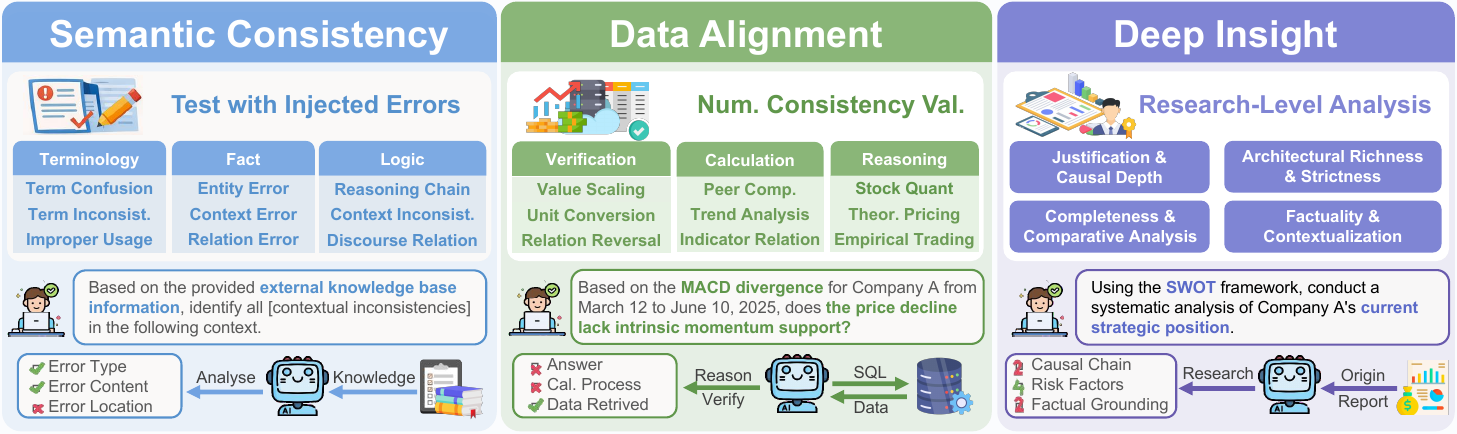}
    \caption{An overview of the taxonomy of FinReasoning benchmark, which is organized into three complementary tracks. Semantic Consistency track evaluates text-grounded error detection and correction. Data Alignment track assesses structured data grounding and executable rule-based reasoning. Deep Insight track measures research-grade analyses. Each track is constructed with multi-level tasks to diagnose LLMs' capabilities from factual reliability to analytical reasoning.}
    \label{fig:constr}
\vspace{-10pt}
\end{figure}

However, recent financial benchmarks like PHANTOM and CNFinBench\cite{ji2025phantom,ding2025cnfinbench,hu-etal-2025-fintrust,islam2023financebench,10.1145/3768292.3770417} reveal that current LLMs pose significant risks in financial writing, including factual errors, fabricated citations, and shallow analysis. 
Real-world reports have repeatedly revealed similar problems. Two government reports Deloitte prepared for Australian and Canadian authorities were found to contain fabricated references and incorrect citations linked to generative-AI hallucinations \cite{theindependent2025deloitte}. In financial research, where numerical precision and temporal consistency are essential, such failures are particularly concerning. These errors can distort views of corporate fundamentals, affecting investment decisions and triggering severe economic losses.
This raises a deeper challenge beyond error detection: evaluating whether models can sustain reliable reasoning throughout the entire analytical process. Many existing benchmarks typically treat hallucination as a binary outcome, without assessing a model’s ability to repair and maintain overall coherence after detection \cite{hu-etal-2025-fintrust,ji2025phantom,zhang2025faith}. Benchmarks closer to financial report generation mainly rely on aggregate accuracy and lack independent evaluation of key abilities like structured exposition, failing to expose analytical bottlenecks \cite{lai-etal-2025-sec, benyoash-etal-2025-secque,islam2023financebench}.

% Consequently, it is imperative to construct a systematic benchmark for reliably evaluating core capabilities of research-report generation process.

% 加一下CNFinBench的引用？模仿一下那一段intro的说法，把社会影响力都写上，应该是rebuttal上的
% 挖一挖为什么需要是不同层级的评估？要深一点。证券分析师的能力评估是不是层级的，专业领域的CFA的知识体系是不是层级的（是层级的，会什么要层级呢？可以说inspired by，因为这种专业知识体系，不同主体/人/大模型，在不同的层级模块上，能力是不一样的。那为什么要分模块能力要测出来呢？因为可以便于下游扩展。可能就是在某一个模块能力强，有些擅长管理，有些擅长详细的数据分析，可以更好地支持下游应用发展和多智能体应用），从这个里面去找依据，由此突出层级化的重要性

Financial reasoning is not a single, unified skill, but a structured process composed of multiple interdependent stages. In professional practice, analysts are evaluated across staged competencies—from ensuring semantic and factual correctness, to grounding intermediate steps in structured data, and ultimately to synthesizing evidence into high-level analytical judgments. 
% This layered organization is also reflected in formal knowledge systems such as the CFA curriculum, where foundational quantitative and accounting skills support higher-level valuation and reasoning. 
As a result, capabilities are rarely uniform across levels: both human analysts and models may perform well in certain stages while remaining limited in others. When these heterogeneous abilities are collapsed into a single aggregate metric, critical differences in capability profiles are obscured, making it difficult to diagnose reasoning failures or support downstream system design. Therefore, evaluating financial reasoning requires a hierarchical framework that can disentangle stage-specific competencies.

To address these gaps, we introduce FinReasoning, an open evaluation resource and benchmark to evaluate three core capabilities underlying financial research workflows. Figure~\ref{fig:constr} illustrates the distribution of tasks, which remains relatively balanced across tracks. Semantic Consistency inspects whether a model can maintain terminological precision, factual correctness, and reasoning continuity in long-form financial text. 
It evaluates the model's error diagnosis and correction quality through correctness, reasoning soundness, relevance, and completeness. 
Data Alignment examines whether a model can anchor its analysis in structured financial data by performing numerical verification, indicator computation, and rule-based reasoning. Its execution success rate and field/date selection accuracy identify where the reasoning pipeline breaks. 
Deep Insight assesses whether a model can transform verified evidence into research-grade analysis, using an expert rubric with 4 dimensions and 12 indicators to assess different aspects. This benchmark reveals the capability of LLMs at different financial reasoning stages, and has already been piloted in several real-world applications, including multi-agent retrieval and analytical systems in banking scenarios.
Our contributions are as follows:
\begin{itemize}[leftmargin=*]
\item We release FinReasoning as a hierarchical evaluation resource for assessing LLMs’ reliability in financial report generation, with tasks aligned to analyst workflows across semantic consistency, data alignment, and deep analytical insight.

%\item We introduce a fine-grained evaluation framework that enhances hallucination-correction evaluation and adds a 12-indicator rubric for analytical skills, revealing strong pattern detection yet weak correction and large disparities in deep causal reasoning.

\item We provide a fine-grained evaluation framework with documented protocols and a 12-indicator analytical rubric, enhancing hallucination-correction assessment and enabling structured, reproducible diagnosis of model strengths and failure modes.

\item We evaluate 19 LLMs and uncover systematic capability stratification across reasoning stages. Models with similar overall performance exhibit different strengths, demonstrating that financial reasoning is not a monolithic ability. These findings highlight the diagnostic value of hierarchical evaluation and provide actionable guidance for model selection and system design.
% 这里要同步改
%\item We evaluate 19 LLMs and characterize their capability profiles: closed-source models remain robust, while financial fine-tuned models struggle with semantic reliability and structured data reasoning, pointing to the need for future adaptation strategies.
\end{itemize}

\section{Related Work}

\subsection{Reasoning Benchmarks in General Domain}
% 把hierarchical也都要加，最后一句要加我们的工作和现有工作的区别，要deep一点，citation也要完整
The evaluation paradigm in the general domain is evolving from single accuracy metrics toward multidimensional reliability validation~\cite{li-etal-2024-dawn,zhu-etal-2021-tat,dong2024bamboo,bai2025longbench}. Within this landscape, assessing a model's reasoning integrity increasingly necessitates a focus on hallucination detection, numerical precision, and deep logical analysis. 
In the area of hallucination and consistency, HalluLens~\cite{bang-etal-2025-hallulens} establishes a rigorous classification system distinguishing between context-conflicting and fact-conflicting, highlighting the primary limitation of current models due to the ambiguity of external knowledge boundaries. In the field of structured and numerical reasoning, RealHiTBench~\cite{wu-etal-2025-realhitbench} introduces complex hierarchical tables and reveals the structural blindness of top models when handling non-flat data, strongly arguing that relying solely on text serialization fails to meet high-precision reasoning requirements. This finding complements earlier work on modal alignment in TAT-DQA~\cite{zhu2022towards} and StructText~\cite{kashyap2150structtext}. Regarding deep logical analysis, while FOLIO~\cite{han-etal-2024-folio} and ProntoQA~\cite{saparov2023language} respectively evaluate first-order logic and multi-step deductive chains, LogicBench~\cite{parmar-etal-2024-logicbench} exposes critical failures in complex negations and parametric interference.

\subsection{Financial Benchmarks}
% CNFinBench论文的引用加一下，然后是否能够支撑智能体？这个评测也要加上来
In the financial domain, early evaluation efforts focused on fragmented capabilities~\cite{lei2023cfbenchmark, li2025cfbenchmark}. TAT-QA~\cite{zhu-etal-2021-tat} and FinQA~\cite{chen-etal-2021-finqa} evaluate if models can perform numerical reasoning over hybrid financial data. To better emulate the complexity of real-world scenarios, recent benchmarks incorporate long financial disclosures and hallucination detection. SEC-QA~\cite{lai-etal-2025-sec} and SECQUE~\cite{benyoash-etal-2025-secque} leverage U.S. Securities and Exchange Commission filings to test insight generation and risk assessment, while FinAuditing~\cite{wang2025finauditing} performs structure-aware logical validation across documents based on international accounting and disclosure principles. Meanwhile, FAITH~\cite{zhang2025faith} targets intrinsic hallucinations in financial tables, and FinTrust~\cite{hu-etal-2025-fintrust} establishes a framework for evaluating fiduciary duty and alignment. 
Despite these advances, existing benchmarks still focus on isolated aspects of the financial workflow, failing to provide a unified and fine-grained evaluation of these capabilities within a coherent analytical framework. Although FinMaster~\cite{jiang2025finmasterholisticbenchmarkmastering} introduces a workflow-oriented setting grounded in real investment analysis, it prioritizes broad task coverage and offers only limited evaluation of reliability, traceability, and consistency in complex multi-step reasoning.
As a result, there remains a gap for a hierarchical evaluation that synthesizes financial logic with rigorous staged reliability validation. In Table~\ref{tab:benchmark_comparison}, we present a comprehensive comparison
of FinReasoning with representative Financial
benchmarks in this area. 

\vspace{-5pt}
\begin{table}[htbp]
\centering
\scriptsize
\caption{Comparison of FinReasoning with existing financial benchmarks. The criteria columns of the table include: (1) task type (objective vs. subjective/open-ended), (2) core competence (calculation, consistency, long text), (3) evaluation protocol (staged evaluation, LLM judge, expert rubric), (4) total evaluation samples, and (5) baselines tested.}
\label{tab:benchmark_comparison}
\resizebox{\textwidth}{!}{
\setlength{\tabcolsep}{2pt}
\begin{tabular}{lccccccccccc}
\toprule
\multirow{2}{*}{Benchmarks} 
& \multicolumn{2}{c}{Task Type} 
& \multicolumn{3}{c}{Core Competence} 
& \multicolumn{3}{c}{Evaluation Protocol} 
% & \multirow{2}{*}{QA Num.}
& \multirow{2}{*}{Baselines} \\
\cmidrule(lr){2-3} \cmidrule(lr){4-6} \cmidrule(lr){7-9}
& Obj. & Subj. & Calculation & Consistency & Long Text & Staged Eval. & LLM Judge & Expert Rubric & & \\
\midrule
TAT-QA \cite{zhu-etal-2021-tat} & \greencheck & \redcross & \greencheck & \redcross & \redcross & \redcross & \redcross & \redcross & 5 \\
FinQA \cite{chen-etal-2021-finqa} & \greencheck & \redcross & \greencheck & \redcross & \redcross & \redcross & \redcross & \redcross & 11 \\
SEC-QA \cite{lai-etal-2025-sec} & \redcross & \greencheck & \redcross & \redcross & \greencheck & \redcross & \greencheck & \redcross & 4 \\
FinAuditing \cite{wang2025finauditing} & \greencheck & \redcross & \redcross & \greencheck & \greencheck & \greencheck & \redcross & \redcross & 13 \\
FAITH \cite{zhang2025faith} & \greencheck & \redcross & \redcross & \greencheck & \redcross & \greencheck & \redcross & \redcross & 16 \\
FinMaster \cite{jiang2025finmasterholisticbenchmarkmastering} & \greencheck & \greencheck & \greencheck & \greencheck & \greencheck & \redcross & \greencheck & \redcross & 7 \\
\midrule
\rowcolor{ourrow}
\textbf{FinReasoning (Ours)} & \greencheck & \greencheck & \greencheck & \greencheck & \greencheck & \greencheck & \greencheck & \greencheck & \textbf{19} \\
\bottomrule
\end{tabular}}
\end{table}
\vspace{-10pt}

\section{FinReasoning}
\subsection{Data Source and Statistics}
FinReasoning is built upon real-world financial materials sourced from a leading domestic financial institution and publicly authorized for use. The prompt templates, code, data, resource usage instructions, and other related information are all available at our project url. For textual resources, the dataset aggregates 677 pieces of financial literature, over 10,000 publicly available news reports, and 901 Chinese financial research reports, covering a broad spectrum of market events, industry narratives, and analytical writing styles. For structured numerical data, we incorporate A-share market data from 2023 to 2025, including fundamental company indicators and daily trading information.
FinReasoning contains 4,800 high-quality financial evaluation samples. Table~\ref{tab:stastics} reports the detailed statistics. Detailed statistics are provided in Appendix~\ref{sec:detailed_statistics}.
\begin{figure}[t]
    \centering
    \includegraphics[width=1\textwidth]{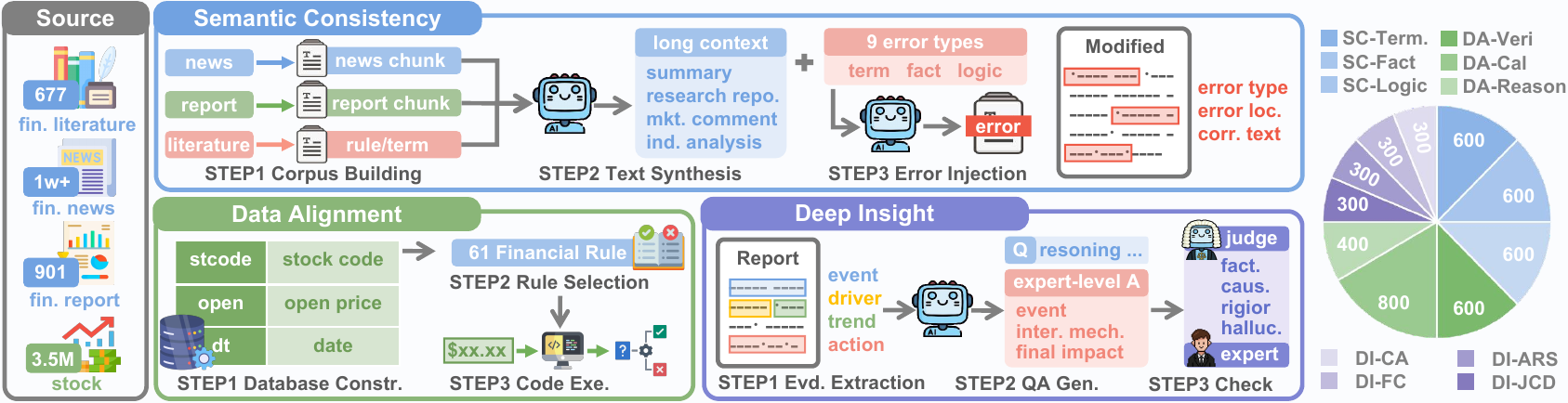}
    \caption{An overview of the FinReasoning benchmark, presenting its data sources, the construction procedures for each task, and the data statistics including the number of sub-tasks within each track.}
    \label{fig:FinReasoning}
\vspace{-10pt}
\end{figure}

\vspace{-5pt}
\begin{table}[h]
    \centering
    \caption{An overview of FinReasoning statistics, including the number of samples and the average, maximum, and minimum lengths of questions and answers.}
    \label{tab:stastics}
    \scriptsize
    \resizebox{\textwidth}{!}{
    \setlength\tabcolsep{10pt}
    \begin{tabular}{c|c|ccc|ccc}
        \midrule
        \multirow{2}{*}{Track} & \multirow{2}{*}{number} & \multicolumn{3}{c|}{Length (Q)} & \multicolumn{3}{c}{Length (A)} \\
        \cmidrule{3-8}
        & & avg. & max & min & avg. & max & min \\\midrule
        Semantic Consistency & 1800 & 8783 & 12183 & 5475 & 6767 & 12317 & 4326 \\
        Data Alignment      & 1800 & 57   & 220   & 23   & 1    & 1     & 1    \\
        Deep Insight         & 1200 & 138  & 347   & 33   & 1483 & 2823  & 653  \\
        \hline
    \end{tabular}
    }
\end{table}
\vspace{-5pt}

\subsection{Task Definition and Construction}
As shown in Figure~\ref{fig:FinReasoning}, FinReasoning decomposes the workflow of research report generation into three sequential capability modules that together reflect the full cognitive process of financial analysts: reading, verification, reasoning, and writing. We present a detailed data collection process and benchmark taxonomy in Appendix~\ref{sec:hierarchical_task_distribution}. DeepSeek-V3 \cite{deepseekai2025deepseekv3technicalreport} is used for QA generation across all tasks, with all outputs subjected to rigorous expert validation. The generation prompts are provided in Appendix~\ref{sec:data_generation_prompts}, and detailed validation procedure is described in Appendix~\ref{sec:manual_data_construction}.

\paragraph{Semantic Consistency (SC).} 
SC targets whether a model can maintain coherent semantics in long-form financial text by identifying and correcting  terminology misuse, factual inaccuracies, and breaks in reasoning chains. This capability aligns with the proofreading stage of analyst workflows, where one must understand long-text structure, compare information across segments, and accurately locate potential errors.

In constructing the dataset, we first conduct a pilot study to ground our error taxonomy in real-world LLM behavior. Specifically, we sample 200 long-form financial texts and prompt five representative models—DeepSeek-V3, Qwen3-235B, GPT-5, Intern-S1, and Fin-R1—to generate corresponding QA pairs or continuations. Through expert review, we identify, categorize, and distill recurring hallucination patterns observed across these outputs, yielding a systematic taxonomy of terminology misuse, factual inaccuracies, and logical inconsistencies commonly produced by LLMs in financial contexts. Guided by these empirically derived error categories, we then design nine controllable error-injection mechanisms that closely mirror the distribution and typology of real model hallucinations. This design allows us to introduce 1–3 diagnostically meaningful perturbations into each sample without disrupting overall fluency, thereby simulating authentic LLM behavior under controlled conditions. The original and perturbed texts are paired with structured annotations specifying error type, location, and corrected content, all verified through expert review. This approach ensures our benchmark diagnoses errors that reflect actual model weaknesses, directly aligning with the proofreading demands of real-world financial auditing scenarios.

\paragraph{Data Alignment (DA).} 
DA focuses on aligning natural-language reasoning with structured financial data and producing correct outcomes for indicator queries, numerical computations, and rule-based judgments. It simulates the workflow of research analysts who extract financial indicators from databases, requiring an accurate understanding of data schemas and the ability to convert analytical intent into database operations.

In constructing the dataset, we design a rule framework based on a structured database of A-share market data from 2023 to 2025.
For verification tasks, we define 4 categories of numerical-consistency and text-perturbation rules to check whether natural-language statements match the values in the database.
For calculation tasks, we specify 35 indicator formulas that combine multiple fields and dates, supporting cross-period comparison and composite metric computation.
For rule-based reasoning tasks, we build 61 executable rules based on established financial analysis practices and commonly used industry heuristics, enabling professional-grade reasoning in trend assessment, ratio interpretation, and conditional judgments.
All rules are broken into atomic steps and executed through a unified calculator. 
During QA generation, we first sample real data and then apply a randomly selected rule to compute or infer the answer, ensuring that every output is grounded in verifiable data and explicit logic rather than the LLM's unreliable numerical guessing.

\paragraph{Deep Insight (DI).} 
This task assesses a model’s ability to generate research-grade analytical insights grounded in evidence, with logical coherence, causal reasoning, structured perspectives, and professional judgment. It mirrors the workflow of research analysts, who distill mechanisms, flag risks, and build clear analytical frameworks from large amounts of information when drafting reports.

In constructing the dataset, we select paragraphs from industry research reports that contain core events or key mechanisms as evidence, keeping limited redundant context to preserve a realistic writing environment. Based on these evidence segments, we craft open-ended analytical questions that prompt the model to reason about the multidimensional impacts of each event. Gold answers are generated by DeepSeek-V3 in an expert role-playing setup, following professional research-writing logic: a framing sentence that sets the structure, a three-step causal chain (event $\rightarrow$ intermediate mechanism $\rightarrow$ final impact), and at least one counterbalancing point addressing execution risks, external constraints, or alternative interpretations. All answers undergo automated and expert review to ensure every claim is evidence-grounded, and experts further verify the overall analytical quality.

\vspace{-5pt}
\begin{table}[htbp]
\centering
\scriptsize
\caption{Evaluation Metrics Overview. This table details the metrics and methodologies. For LLM-judge, we use a fused protocol combining DeepSeek-V3 and Qwen3-235B.}
\label{tab:metrics}
\scriptsize
\resizebox{\textwidth}{!}{
\setlength{\tabcolsep}{10pt}
\begin{tabular}{lll}
\toprule
Track & Evaluation Metrics & Method \\
\midrule
% 第一部分：Semantic Consistency
\multirow{3}{*}{Semantic Consistency} 
   & Error Location        & Sentence-F1 \\
%\cmidrule(lr){2-3}
   & {Error Explanation} & BERTScore, SimCSE, LLM-as-a-Judge \\
%\cmidrule(lr){2-3}
   & {Correct Content}  & BERTScore, SimCSE, LLM-as-a-Judge \\
\midrule
% 第二部分：Data Alignment
\multirow{3}{*}{Data Alignment} 
   & Success Rate          & Rate \\
%\cmidrule(lr){2-3}
   & Answer Accuracy       & Accuracy \\
%\cmidrule(lr){2-3}
   & Retrieval Accuracy     & F1-Score (Data ID, Field) \\
\midrule
% 第三部分：Deep Insight
Deep Insight  & Category-specific Scoring Criteria & LLM-as-a-Judge \\
\bottomrule
\end{tabular}}
\end{table}
\vspace{-5pt}

\subsection{Benchmark Evaluation}
\paragraph{Evaluation Method}
As shown in Table~\ref{tab:metrics}, FinReasoning adopts a hybrid evaluation framework that combines objective metrics with model-based judgments. To mitigate single-model bias, we employed DeepSeek-V3 and Qwen3-235B as judge models and fused their outputs via grid search over weighting parameters to maximize alignment with expert annotations. 
All metrics are normalized and then averaged to produce a composite score for each track. Detailed evaluation prompts and scoring rubrics are provided in Appendix~\ref{sec:evaluation_prompts}. 

In SC track, we use sentence-level Precision, Recall, and F1 to evaluate error localization, and employ BERTScore and SimCSE to measure the semantic alignment of error explanations and corrected text. To assess hallucination diagnosis and correction quality, we further introduce an LLM-as-a-Judge mechanism that assigns fine-grained scores to both the explanation and the revision along four dimensions: 
correctness, reasoning rigor, relevance, and completeness.

In DA track, we separate evaluation into outcome correctness and process correctness to prevent cases where a model reaches the right answer through an invalid reasoning path. Answer Accuracy measures whether the final judgment is correct, while Retrieval Accuracy, computed as the F1 score over retrieved fields and row identifiers, checks whether the model’s reasoning is grounded in the correct data source. Success Rate is also necessarily recorded because some LLMs may fail during the query phase or be unable to return the correct format.

In DI track, we employ an expert-designed rubric through an LLM-as-a-judge to score model-generated analyses across four dimensions comprising twelve indicators: Justification and causal depth, including completeness of the causal chain, necessity of causal explanation, and adherence to professional analytical norms; Factuality and contextualization, examining factual density, computational rigor, and contextual reasoning; Completeness and comparative analysis, evaluating critical thinking, comparative perspective, and balance of arguments; Architectural richness and strictness, assessing framework completeness, logical hierarchy, and expression granularity.

\paragraph{Evaluation Reliability}
We conduct several robustness checks beyond the main scoring protocol. 
Two financial experts independently reviewed 10\% of LLM-Judged samples for both DeepSeek-V3 and Qweb3-238B; the resulting inter-expert
agreement was $\kappa=0.79$ on SC and $\kappa=0.76$ on DI, which supports the use of the fused LLM-as-a-Judge protocol as a scalable evaluation mechanism. Full verification details and fusion strategy are provided in Appendix~\ref{sec:manual_judge_verification}.
Prompt perturbation experiments in Appendix~\ref{sec:prompt_sensitivity}  show that rephrasing the judge's instructions has limited impact on model rankings, with an average Spearman's $\rho$ of 0.95. Bootstrap confidence intervals provided in Appendix~\ref{sec:bootstrap_ci} further indicate that the major performance gaps are statistically stable. We also observe systematic differences in judge stringency and mild model-family preference in Appendix~\ref{sec:inter-judge_consisency}. The Pearson correlation coefficients for two LLM judges reach 0.94, so judge-based scores primarily serve ranking stability and diagnostic comparison rather than calibrated absolute quality estimates.

\section{Experiments}
\subsection{Experimental Setup}
We evaluate the zero-shot performance of 16 general LLMs and 3 financial LLMs. The closed-source general models include GPT series~(GPT-5, GPT-4o)~\cite{hurst2024gpt,singh2025openai}, Gemini-3-pro~\cite{team2025gemma}, Claude-sonnet-4.5~\cite{anthropic2025claude4}, GLM-4.6~\cite{zeng2025glm}, Qwen3-Max~\cite{yang2025qwen3}, Kimi-K2~\cite{kimiteam2026kimik2openagentic} and Doubao-Seed-1.8~\cite{seed2025seed1}. 
The open-source general models include  DeepSeek series~(DeepSeek-V3,  DeepSeek-R1)~\cite{deepseekai2025deepseekv3technicalreport, guo2025deepseek}, Qwen series~(Qwen3-8B, Qwen3-32B, Qwen3-235B)~\cite{yang2025qwen3}, Llama series~(Llama-3.1-8B, Llama-3.1-70B)~\cite{dubey2024llama} and Intern-S1~\cite{bai2025intern}. 
The open-source financial models include \textbf{Fin-R1}~\cite{liu2026finr1largelanguagemodel}, TouchstoneGPT~\cite{wu-etal-2025-golden} and DianJin-R1-7B~\cite{zhu2025dianjin}, all of which are fine-tuned on Qwen2.5~\cite{yang2024qwen2}.
For open-source LLMs, we directly utilize their released model weights on Nvidia A100 GPUs during the evaluation. All models are evaluated under a unified decoding temperature of 0.1. For those who wish to use our evaluation suite, we provide detailed evaluation cost in Appendix~\ref{sec:computational_cost} as reference.

\begin{table}[htbp]
    \centering
    \scriptsize
    \setlength{\tabcolsep}{2pt}
    \caption{Performance comparison of LLMs across different tracks (Best in bold, second best underlined; the full leaderboard is ranked by the Overall Normalized Score). Ver.=Verification; Cal.=Calculation; Reason.=Reasoning; ARS=Architectural Richness and Strictness; FC=Factuality and Contextualization; CA=Completeness and Comparative Analysis; JCD=Justification and Causal Depth; Avg.=Average Score; Norm.=Normalized Score.}
    \label{tab:llm_performance_norm}
    \resizebox{\textwidth}{!}{
    \begin{tabular}
    {
    l|
    ccc>{\columncolor[HTML]{EDECEC}}c>{\columncolor[HTML]{E4EAF7}}c|
    ccc>{\columncolor[HTML]{EDECEC}}c>{\columncolor[HTML]{ECF6E5} }c|
    cccc>{\columncolor[HTML]{EDECEC}}c>{\columncolor[HTML]{E5E5FA}}c|
    c}
        \toprule
        \multirow{2}{*}{Model} & \multicolumn{5}{c|}{Semantic Consistency} & \multicolumn{5}{c|}{Data Alignment} & \multicolumn{6}{c|}{Deep Insight} & Overall \\
        \cmidrule(lr){2-6} \cmidrule(lr){7-11} \cmidrule(lr){12-17} \cmidrule(lr){18-18}
         & Term. & Fact & Logic & Avg. & Norm. & Ver. & Cal. & Reason. & Avg. & Norm. & ARS & F\&C & C\&A & JCD & Avg. & Norm. & Norm. \\
        \midrule
        Doubao-Seed-1.8 & \underline{63.9} & \textbf{74.4} & \textbf{63.3} & \textbf{67.2} & \textbf{100.0} & \underline{93.5} & 66.6 & 55.6 & 71.9 & 92.1 & \textbf{93.5} & 87.2 & 86.2 & 95.7 & 90.6 & 98.3 & \textbf{96.8} \\
        GPT-5 & \textbf{64.3} & 71.4 & 57.6 & 64.4 & 94.9 & 88.3 & \textbf{72.8} & \textbf{68.5} & \textbf{76.5} & \textbf{100.0} & 92.4 & 84.6 & 86.7 & 81.8 & 86.4 & 91.1 & \underline{95.4} \\
        Kimi-K2 & 49.5 & 66.0 & 59.1 & 58.2 & 83.7 & 90.2 & 69.3 & 63.2 & 74.2 & 96.1 & \underline{93.4} & \underline{87.4} & \textbf{89.5} & \textbf{96.3} & \textbf{91.6} & \textbf{100.0} & 93.3 \\
        Claude-sonnet-4.5 & 54.9 & 69.0 & 51.8 & 58.6 & 84.4 & 85.3 & 64.6 & 62.7 & 70.9 & 90.4 & 92.2 & 87.1 & 84.5 & \underline{96.2} & 90.0 & 97.3 & 90.7 \\
        GLM-4.6 & 59.0 & \underline{73.4} & \underline{62.5} & \underline{65.0} & \underline{96.0} & 89.4 & 67.5 & 58.4 & 59.4 & 70.7 & 92.6 & \textbf{88.1} & \underline{87.3} & 96.1 & \underline{91.0} & \underline{99.0} & 88.6 \\
        Gemini-3-pro & 63.1 & 71.1 & 50.2 & 61.5 & 89.7 & \textbf{94.3} & \underline{69.9} & 56.8 & 73.7 & 95.2 & 82.1 & 79.0 & 76.7 & 80.7 & 79.6 & 79.5 & 88.1 \\
        Qwen3-Max & 51.2 & 70.0 & 57.1 & 59.4 & 85.9 & 92.3 & 67.1 & \underline{66.3} & \underline{75.2} & \underline{97.8} & 85.8 & 76.0 & 75.8 & 82.9 & 80.1 & 80.4 & 88.0 \\
        DeepSeek-V3 & 41.7 & 63.1 & 54.0 & 52.9 & 74.1 & 91.8 & 63.7 & 60.4 & 72.0 & 92.3 & 82.0 & 81.0 & 77.9 & 86.4 & 81.8 & 83.3 & 83.2 \\
        Intern-S1 & 39.5 & 57.1 & 35.4 & 44.0 & 58.0 & 88.8 & 62.9 & 53.9 & 68.5 & 86.3 & 84.4 & 82.1 & 81.6 & 88.0 & 84.0 & 87.0 & 77.1 \\
        GPT-4o & 35.7 & 53.8 & 37.6 & 42.4 & 55.2 & 90.4 & 61.0 & 63.1 & 71.5 & 91.4 & 80.0 & 77.6 & 74.8 & 76.9 & 77.3 & 75.6 & 74.1 \\
        Qwen3-235B & 40.3 & 57.7 & 33.0 & 43.7 & 57.5 & 90.1 & 63.1 & 54.2 & 69.1 & 87.3 & 75.4 & 63.8 & 63.6 & 67.5 & 67.6 & 59.0 & 68.0 \\
        DeepSeek-R1 & 51.6 & 66.9 & 46.7 & 55.0 & 77.9 & 25.7 & 68.9 & 22.5 & 39.1 & 36.0 & 88.0 & 84.6 & 80.0 & 89.8 & 85.6 & 89.8 & 67.9 \\
        Qwen3-32B & 38.1 & 54.9 & 34.2 & 42.4 & 55.2 & 91.6 & 55.7 & 65.0 & 70.8 & 90.2 & 73.6 & 61.7 & 60.3 & 65.7 & 65.3 & 55.1 & 66.8 \\
        Qwen3-8B & 32.4 & 46.1 & 37.6 & 38.7 & 48.5 & 83.5 & 51.8 & 54.2 & 63.2 & 77.2 & 69.3 & 51.1 & 53.9 & 57.4 & 57.9 & 42.5 & 56.1 \\
        Fin-R1 & 21.4 & 22.6 & 24.2 & 22.8 & 19.7 & 65.4 & 40.6 & 38.6 & 48.2 & 51.5 & 77.9 & 71.8 & 73.5 & 71.5 & 73.7 & 69.5 & 46.9 \\
        Llama-3.1-70B & 19.7 & 28.8 & 20.4 & 23.0 & 20.1 & 88.2 & 61.3 & 61.3 & 70.3 & 89.4 & 42.6 & 36.5 & 34.7 & 36.8 & 37.7 & 8.0 & 39.2 \\
        DianJin-R1-7B & 19.1 & 22.7 & 20.3 & 20.7 & 15.9 & 65.9 & 20.1 & 6.9 & 30.9 & 21.9 & 77.1 & 63.6 & 73.3 & 69.1 & 70.8 & 64.5 & 34.1 \\
        Llama-3.1-8B & 19.4 & 21.1 & 16.9 & 19.2 & 13.2 & 78.4 & 49.1 & 42.6 & 56.7 & 66.1 & 39.1 & 30.8 & 31.3 & 32.1 & 33.3 & 0.5 & 26.6 \\
        TouchstoneGPT & 12.0 & 12.0 & 11.6 & 11.8 & 0.0 & 23.8 & 17.9 & 12.6 & 18.1 & 0.0 & 38.3 & 28.7 & 30.8 & 34.2 & 33.0 & 0.0 & 0.0 \\
        \midrule
        Average & 40.9 & 52.7 & 40.7 & 44.8 & 59.5 & 79.8 & 57.6 & 50.9 & 62.1 & 75.4 & 76.8 & 69.6 & 69.6 & 73.9 & 72.5 & 67.4 & 67.4 \\
        \bottomrule
    \end{tabular}
    }
\end{table}

\subsection{Main Results}

In order to enable meaningful cross-track comparison despite heterogeneous scoring mechanisms, we apply a two-stage Z-Score \& Min-Max normalization to produce a unified leaderboard shown in Table~\ref{tab:llm_performance_norm}. The overall ranking reveals Doubao-Seed-1.8 (96.8), GPT-5 (95.4), and Kimi-K2 (93.3) as the top 3 models in these tasks, while all three financial domain models rank in the bottom. 
Beyond overall ranking, a key observation is the systematic divergence of model capabilities across different reasoning stages. % 强化“分层带来的新发现”
We present the detailed results of each evaluation rubric in Appendix~\ref{sec:detailed_model_results}, additional quantitative statistics and case analyses are provided in Appendix~\ref{sec:failure_studies}.

\subsubsection{Semantic Consistency}
Table~\ref{tab:SC_details_metrics_tasks}
reports the detailed scores in SC, where the Overall Score is computed as the mean of Error Location, Error Explanation, and Correct Content; the latter two are each averaged over three sub-metrics. Validation shows that BERTScore and SimCSE achieve Pearson correlations of 0.86 and
0.91 with LLM-Judge, confirming the reliability of LLM scoring.

\begin{table}[h]
    \centering
    \scriptsize
    \setlength{\tabcolsep}{5pt}
    \caption{Detailed results across tasks in Semantic Consistency.}
\label{tab:SC_details_metrics_tasks}
    \resizebox{\textwidth}{!}{
    \begin{tabular}{l|c|ccc|ccc|c}
        \toprule
        \multirow{2}{*}{Task} & Error Location & \multicolumn{3}{c|}{Error Explanation} & \multicolumn{3}{c|}{Correct Content} & \multirow{2}{*}{Overall Score} \\
        \cmidrule{2-8} & Sentence-F1 & BERTScore & SimCSE & LLM & BERTScore & SimCSE & LLM & \\
        \midrule
        Terminology & 20.6 & 66.0 & 67.5 & 49.9 & 36.7 & 35.7 & 50.4 & 40.9 \\
        Fact        & 40.0 & 64.8 & 69.3 & 57.7 & 52.8 & 51.5 & 58.5 & 52.7 \\
        Logic       & 21.5 & 61.9 & 62.7 & 62.8 & 26.4 & 25.8 & 62.2 & 40.7 \\
        \bottomrule
    \end{tabular}
    }
\end{table}
\vspace{-5pt}

\textbf{Finding 1: Models can explain errors but struggle to correct them.}  Across all three tasks, Error Explanation scores are consistently higher than Correct Content scores, with gaps of 20.2 points for Terminology, 9.7 for Fact, and 24.3 for Logic. These results indicate that models often rely on surface clues to describe what is wrong, yet struggle to produce revisions that align with financial-domain conventions or maintain contextual coherence. The difficulty is most pronounced in Logic, where correcting an error requires reconstructing an entire reasoning chain rather than adjusting a term or a number. This suggests that models lack the structured financial reasoning needed to transform error detection into valid, domain-grounded corrections.

\textbf{Finding 2: Fact-type errors are easier to detect than Terminology and Logic errors.} Fact errors (Avg. = 52.7) outperform Terminology (40.9) and Logic (40.7) by approximately 12 points. This is because factual errors, such as incorrect entities or misattributed relationships, create clearer textual inconsistencies that models can detect from surface patterns. Terminology errors depend on precise financial vocabulary knowledge, while Logic errors require multi-step causal reasoning.

\subsubsection{Data Alignment}
Table~\ref{tab:DA_details_metrics}
presents the detailed scores of selected LLMs, where the Overall Score is computed by multiplying the Success Rate by the average of other scoring metrics.

\textbf{Finding 1: Performance declines consistently with task difficulty, but the extent of degradation varies sharply across model categories.} As shown in Figure~\ref{fig:L_decay}, average performance degrades monotonically from Ver. (79.8) to Cal. (57.6) to Reason. (50.9), with a mean decay rate of 36.6\%. Success rates follow the same trend, showing that models not only answer less accurately but also fail more often to produce valid outputs as tasks require more steps and cross-field reasoning.
However, the magnitude of degradation differs substantially by model family.

Closed-source models maintain near-perfect format compliance and experience only moderate accuracy declines, reflecting stronger general reasoning and error-recovery capabilities. This stability likely stems from their larger model capacity and more robust instruction tuning, which help preserve structured outputs as task complexity increases. In contrast, financial domain models deteriorate severely as task complexity rises. For example, DianJin-R1-7B falls from 65.9 (Ver.) to 6.9 (Reason.), an 89.5\% decline, driven largely by its success rate dropping to 16.0\%. This indicates that small financial models lack the structural capacity to maintain coherent output formats when reasoning chains become long or interdependent.

\noindent
\begin{minipage}[c]{0.48\textwidth}
    \centering
    \scriptsize
    \setlength{\tabcolsep}{1pt}
    \captionof{table}{Detailed results of Reasoning tasks. Success R.=Success Rate; Ans. Acc.=Answer Accuracy; Over. S.=Overall Score.}
    \scriptsize
    \label{tab:DA_details_metrics}
    \begin{tabular}{lccccc}
        \hline
        Model & Success R. & Ans. Acc. & ID F1 & Fields F1 & Over. S. \\
        \hline
        GPT-5 & 100.0 & 63.0 & 64.7 & 77.8 & 68.5 \\
 Qwen3-Max& 99.5 & 60.8 & 63.0 & 76.1 &66.3 
\\
        % Gemini-3-pro & 100.0 & 60.8 & 38.6 & 71.1 & 56.8 \\
        DeepSeek-R1& 38.8& 60.6 & 49.0 & 64.6 & 22.5 
\\
         Fin-R1 & 59.8& 60.7 & 80.4 & 53.0 & 38.6 \\
     DianJin-R1-7B& 16.0& 51.6 & 44.0 & 34.0 &6.9 \\
        \bottomrule
    \end{tabular}
\end{minipage}
\hfill
\begin{minipage}[c]{0.5\textwidth}
    \centering
    \includegraphics[width=\linewidth]{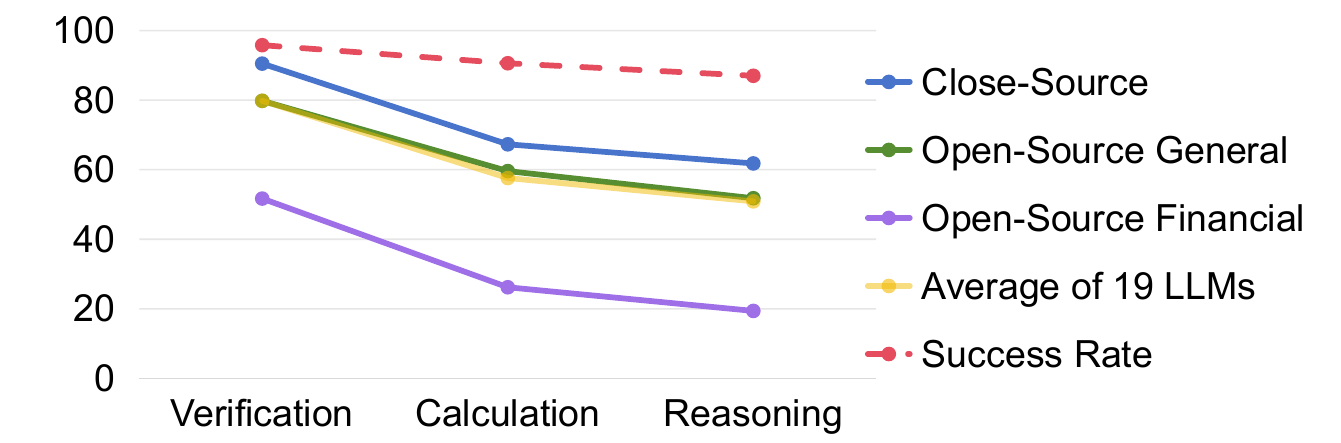}
    \captionof{figure}{Comparative performance across model categories over three levels of the DA track.}
    \label{fig:L_decay}
\end{minipage}
\vspace{5pt}

\textbf{Finding 2: Reasoning-specialized models fail unexpectedly on simple tasks.} Although DeepSeek-R1 performs competitively on Cal. (68.9), its Ver. score drops sharply to 25.7. Diagnostics show a 100\% Success Rate but near-random Answer Accuracy (23.5\%) and ID F1 (28.1\%), indicating that the model produces outputs in the correct format but retrieves the wrong content. Similar patterns appear in Reason.-style tasks, where DeepSeek-R1 reaches 60.6\% Answer Accuracy yet only 38.8\% Success Rate, with frequent format violations; DianJin-R1-7B and Fin-R1 display the same trend. 
These results suggest that models optimized for complex reasoning may over-reason on direct lookup tasks, inserting unnecessary inference steps that disrupt simple data retrieval. They also tend to lose format stability when the reasoning load increases. These behaviors collectively reduce their reliability on tasks that require precise and strictly structured responses.

\begin{figure}[h]
    \centering
    \includegraphics[width=0.7\linewidth]{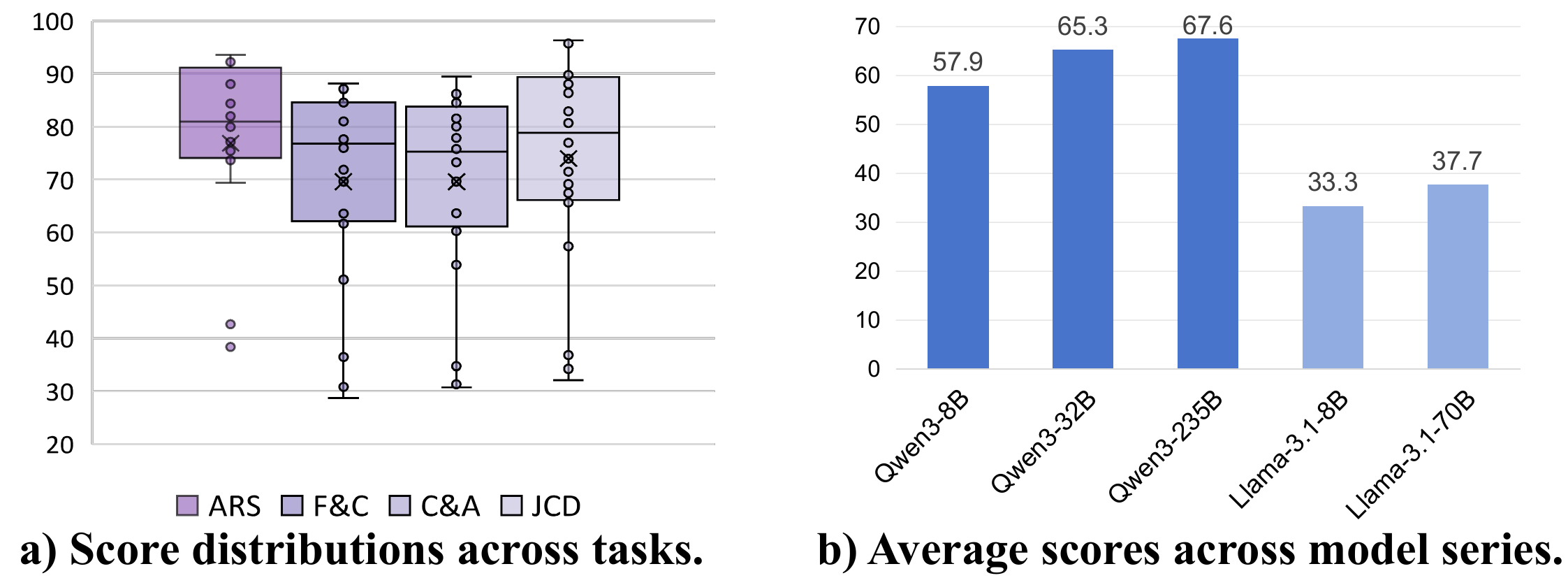}
    \caption{Visualized DI scores across tasks and model series.}
    \label{fig:DI_box}
\end{figure}
\vspace{-8pt}

\subsubsection{Deep Insight}
The DI track evaluates open-ended financial analysis through four dimensions via LLM-as-a-Judge.

\textbf{Finding 1: Causal reasoning is the primary capability that separates top-tier models from the rest.} Figure~\ref{fig:DI_box}.a shows per-dimension score distributions, revealing that most models handle structured exposition well. However, JCD displays a strong split: four models score above 95 (Kimi-K2, Claude-Sonnet-4.5, GLM-4.6, Doubao-Seed-1.8), while the remaining models cluster much lower. In contrast, F\&C and C\&A (both mean = 69.6) remain difficult, requiring accurate evidence grounding and multi-perspective analytical judgment rather than surface-level summarization. These results suggest that only a few frontier models can reliably form multi-step causal links, whereas most models lose coherence once the task moves beyond descriptive summarization. The ability to integrate events, mechanisms, and impacts appears to underlie the performance gap.

\textbf{Finding 2: Model scale is helpful but insufficient for deep financial analysis.} As shown in Figure~\ref{fig:DI_box}.b, across Qwen3 variants, scaling from 8B to 32B yields a clear 12.8\% gain, yet increasing to 235B adds only another 3.5\%. Even more notably, Llama-3.1-8B (33.3) and Llama-3.1-70B (37.7) both remain far below the DI track average (72.5), despite the 70B model performing competitively on DA (70.3). Their weak DI results are further affected by limited Chinese comprehension. These observations show that improvements in parametric scale and structured reasoning (as reflected in DA) do not automatically translate into strong open-ended financial analysis. Deep Insight tasks rely on knowledge, disciplined instruction-following, and multi-step analytical planning.

\subsection{Discussion}

% \paragraph{Failure Pattern Analysis.}
% We also construct failure studies and find that model errors are rarely isolated local mistakes; they often stem from a mismatch between the intended financial task and the reasoning procedure adopted by the model. In SC track, many outputs exhibit knowledge-absence-driven negative reasoning, models mark claims as erroneous merely because they are not explicitly supported by the external knowledge base. 
% In DA track, the dominant risk is pseudo-traceable reasoning. Models may retrieve plausible fields yet still execute a different rule from the one specified by the benchmark. This pattern shows auditability depends jointly on answer correctness, data alignment, and rule fidelity.
% In DI track, the main weakness is insufficient analytical conditionality. Many low-scoring responses imitate the surface form of research reports but fail to identify the core mechanism, or specify when a positive thesis would break down.
% These findings suggest that current LLMs can imitate financial-report style more easily than they can perform disciplined financial reasoning. More detailed quantitative statistics and case analyses are provided in Appendix~\ref{sec:failure_studies}.

\paragraph{Hierarchical Model Diagnosis.}
Performance varies sharply across stages. For error detection and correction in long-text, Doubao-Seed-1.8 and GLM-4.6 show stronger control over factual accuracy and logical consistency. Tasks requiring numerical verification and multi-step reasoning are better handled by GPT-5 and Qwen3-Max, reflecting stronger execution over tabular data and rule-based operations. For higher-level analytical writing abilities, Kimi-K2 and Claude-sonnet-4.5 produce more coherent and well-grounded insights.
These patterns indicate that models strong in high-level analysis do not necessarily maintain stable foundations, while those with strong semantic reliability may not excel in structured reasoning. Across model families, closed-source models are more balanced, open-source models show larger variance, and financial fine-tuned models lag in foundational stages despite moderate high-level performance. This suggests that model capabilities should be matched to stage-specific roles in downstream systems.

\paragraph{Directions for Improving Financial Fine-Tuned Models.}
Across all three tracks, financial-domain models lag behind closed-source and general open-source models of similar scale, with average gaps of 26.4 points on Semantic Consistency and 29.7 on Data Alignment. Even on Deep Insight, they remain 13.34 points behind. For example, the 7B financial models Fin-R1 (SC: 22.8) and DianJin-R1-7B (SC: 20.7) score far below Qwen3-8B (SC: 38.7). These patterns indicate that current financial fine-tuning improves terminology recognition and event-based summarization but does not develop the reasoning needed to validate, connect, and transform financial information. Because training signals emphasize surface-level associations rather than structured inference, models learn to imitate financial language but struggle with systematic, rule-consistent manipulation of data.

Future development should integrate objectives that enhance reliability and structured operations, such as hallucination correction and supervised practice with database- and API-grounded reasoning. These capabilities are increasingly necessary as financial institutions adopt agent-based workflows that require verifiable retrieval, multi-step execution, and consistent outputs under reasoning demands.

\paragraph{Future Work.}
Our results show that models perform substantially worse on Semantic Consistency and Data Alignment (average scores of 44.8 and 62.1) than on Deep Insight (72.5). This suggests that the fundamental bottlenecks for financial LLMs lie in hallucination detection and correction, as well as in executing structured, database-grounded operations. However, our current evaluation does not measure how hallucinations propagate into downstream tasks. For instance, whether undetected inconsistencies in SC ultimately distort higher-level analysis in DI. Designing end-to-end settings to quantify these effects remains an important direction.

Moreover, Data Alignment is currently evaluated through controlled interactions with a local database, which prevents destructive operations but also limits the scope of executable queries. Future work should develop a sandboxed environment that safely allows the full range of database operations, enabling models to perform realistic retrievals, updates, joins, and multi-step financial computations. Such environments would better simulate real-world financial workflows and provide a more comprehensive assessment of LLMs as autonomous financial agents.
% \subsubsection{Limitations}
% \textbf{(1) The evaluation methodology itself reveals important limitations.} The disparity between SC's rule-based evaluation (mean = 44.8) and DI's LLM-as-a-Judge evaluation (mean = 72.5) reflects known score inflation in generative evaluation. The elevated DI scores for DeepSeek series (evaluated by DeepSeek-V3) warrant caution. Our future work will incorporate more specific and sufficient cross-validation to enhance evaluation reliability for open-ended financial reasoning tasks.

\section{Conclusion}
We introduce FinReasoning, a Chinese benchmark that evaluates the core abilities required in financial report generation across three dimensions: semantic consistency, data alignment, and deep analytical insight, which are specifically constructed through hierarchical tasks. 
To support rigorous assessment, we design a framework with a unified metric suite combining objective scoring for error localization and database-grounded operations with LLM-based judging for analytical quality.
% Experiments on 19 LLMs reveal three trends: models struggle most with semantic reliability and structured data grounding, financial-domain models show substantially lower robustness than general models, and only a few frontier models can produce coherent causal reasoning. These findings expose persistent weaknesses in factual stability and multi-step reasoning.
Experiments on 19 LLMs expose persistent weaknesses in factual stability and multi-step reasoning, and reveal that models with similar overall performance often exhibit markedly different strengths across different capabilities. This indicates that financial reasoning is not a monolithic ability but a composition of unevenly developed competencies.
Thus, our hierarchical evaluation not only highlights a direction for future financial model development, but also serves as a foundation for designing more reliable and modular multi-agent financial systems in the future. %强化benchmark到system implication这一收束逻辑
% Thus, FinReasoning highlights a direction for future financial model development: fine-tuning must strengthen hallucination correction, structured data interaction, and multi-step reasoning rather than merely adding domain knowledge.

\small{
\bibliographystyle{nips}
\bibliography{cite}
}

%%%%%%%%%%%%%%%%%%%%%%%%%%%%%%%%%%%%%%%%%%%%%%%%%%%%%%%%%%%%
\newpage
\appendix

% \section{Technical appendices and supplementary material}
% Technical appendices with additional results, figures, graphs, and proofs may be submitted with the paper submission before the full submission deadline (see above). You can upload a ZIP file for videos or code, but do not upload a separate PDF file for the appendix. There is no page limit for the technical appendices. 

% Note: Think of the appendix as ``optional reading'' for reviewers. The paper must be able to stand alone without the appendix; for example, adding critical experiments that support the main claims to an appendix is inappropriate. 

%%%%%%%%%%%%%%%%%%%%%%%%%%%%%%%%%%%%%%%%%%%%%%%%%%%%%%%%%%%%

\section{Limitations}
The substantive contribution of this benchmark lies in its task design, which closely mirrors real-world business logic. Rather than confining itself to conventional financial question answering or information extraction, the benchmark deconstructs research report generation into reading, validation, reasoning, and writing stages that align with the cognitive processes of human analysts, offering strong diagnostic value for industrial applications.
However, FinReasoning still has few limitations:
\begin{itemize}[left=15pt, labelsep=3pt]
    \item LLM-as-a-Judge evaluation still carries the risk of systematic scoring bias. As shown in Section~\ref{sec:inter-judge_consisency}, DeepSeek-V3 and Qwen3-235B preserve highly consistent model rankings, but their absolute scores are not identical. This suggests that the robustness of our judge-based evaluation should be interpreted primarily at the ranking and diagnostic-pattern level rather than as perfectly calibrated absolute scores. In addition, since part of the benchmark construction relies on LLM-generated reference answers, the gold responses may retain certain stylistic or structural preferences. We mitigate this risk through expert validation, fused judging, prompt perturbation tests, and correlation checks with objective metrics, but future extensions should further incorporate independent human-written references or judge calibration sets that exclude model families involved in data generation or evaluation.
    \item Although the benchmark is designed around the workflow of financial research reporting, the three tracks are evaluated as staged capability modules rather than as a fully end-to-end report-generation pipeline; therefore, the current setting does not directly measure how an error in semantic consistency or data alignment propagates into downstream investment analysis.
    \item Part of the dataset construction relies on LLM-assisted generation followed by expert verification; while this enables scalable and fine-grained benchmark construction, it may still introduce residual artifacts in question style, answer structure, or reasoning templates.
\end{itemize}
Our future work will try to extend FinReasoning toward cross-market and multilingual settings, sandboxed executable data environments, and end-to-end evaluations that quantify how local factual or numerical failures affect final research conclusions.

\section{Detailed Corpus and Task Statistics}
\label{sec:detailed_statistics}
In this section, we provide a comprehensive statistical overview of the FinReasoning benchmark to ensure transparency and facilitate reproducibility. In section~\ref{sec:data_collection}, we detail the composition of the textual corpus, which aggregates professional literature, financial news, and research reports to simulate authentic analyst work environments. We also present the specifics of the structured numerical database, encompassing extensive market indicators and historical trading records across the A-share market. In section~\ref{sec:hierarchical_task_distribution}, we provide a granular breakdown of the task distribution across the Semantic Consistency, Data Alignment, and Deep Insight tracks. These statistics illustrate the structural diversity and logical complexity of our evaluation scenarios, highlighting the benchmark's capacity to assess multi-level financial reasoning.

\subsection{Data Provenance and Collection Pipelines}~\label{sec:data_collection}
FinReasoning Evaluation Resource Construction aims to comprehensively simulate real-world financial research and analysis workflows by integrating high-quality, multi-dimensional domain-specific corpora. All source materials are obtained from leading domestic professional financial data service platforms and are accompanied by public authorization for academic research use.
\begin{itemize}[left=15pt, labelsep=3pt]
    \item \textbf{Textual Resources:} We collected 677 professional documents covering financial industry standards and term definitions, 901 in-depth Chinese research reports authored by licensed analysts, and over 10,000 publicly available financial news articles. Spanning market dynamics and industry trends from 2023 to 2025, these materials provide a rich contextual foundation for assessing model performance in terms of logical consistency and analytical depth over long texts. From the above documents, we further extracted authoritative industry rules and term definitions, and combined these with key paragraphs from research reports that describe core events or economic mechanisms, thereby constructing a set of text segments grounded in authentic analytical contexts.
    \item \textbf{Structured Numerical Data:} We integrated comprehensive market data and company fundamental indicators for the A-share market from 2023 to 2025. This database comprises over 3.5 million records, including but not limited to daily trading prices, financial statement line items, and various derived financial indicators, forming the underlying environment for SQL execution and numerical logic validation in the Data Alignment track.
    \item \textbf{Data Preprocessing:} We implemented a rigorous preprocessing pipeline. First, through a combination of automated scripts and manual review, we removed personally identifiable information (PII) from the corpora, such as personal contact details and internal employee IDs. Second, we filtered redundant content from the research reports, eliminating advertisements and duplicate information, and retaining only narratively valuable segments with evidential value. This approach enhances the focus of the evaluation tasks while simulating authentic, complex real-world environments.
\end{itemize}

\subsection{Hierarchical Task Distribution}~\label{sec:hierarchical_task_distribution}
In this subsection, we provide a detailed breakdown of FinReasoning's taxonomy, which is structured into a three-track hierarchy to evaluate distinct cognitive stages in financial research reporting. As summarized in Table \ref{tab:task_distribution}, the benchmark comprises 4,800 high-quality evaluation samples balanced across \textbf{Semantic Consistency (SC)}, \textbf{Data Alignment (DA)}, and \textbf{Deep Insight (DI)}. The SC track encompasses 1,800 samples focusing on terminology precision, factual accuracy, and logical continuity. The DA track includes 1,800 samples assessing a model's ability to perform precise numerical calculations and rule-driven reasoning over structured databases. The DI track consists of 1,200 samples evaluating research-grade analytical depth across 12 expert indicators. This hierarchical distribution ensures a granular and systematic assessment of model reliability within professional financial workflows.
\vspace{-10pt}
\begin{table}[h]
\centering
\caption{Hierarchical Task Distribution of FinReasoning. This table provides a detailed breakdown of the tracks, categories, and dimensions, including the specific logic and sample size for each task.}
\label{tab:task_distribution}
%\small
\resizebox{\textwidth}{!}{
\setlength\tabcolsep{2pt}
\begin{tabular}{llp{8cm}c}
\toprule
Category & Dimension & Explanation & Number \\
\midrule

% --- Track 1: Semantic Consistency ---
\rowcolor{TableBlue}
\multicolumn{4}{l}{\textbf{Semantic Consistency (SC)}} \\\midrule
\multirow{3}{*}{\shortstack[l]{Terminology}} & Improper Usage & Inconsistent terminology, reducing rigor & 200 \\

 & Terminology Inconsistency& Use of colloquial or non-domain standard vocabulary & 200 \\
 & Terminology Confusion & Words with similar semantics but different definitions & 200 \\
\cmidrule(l){1-4}
\multirow{3}{*}{\shortstack[l]{Fact}} & Relation Error & Errors in predicates, actions, or relationship descriptions & 200 \\
 & Entity Error & Wrong key entities or subject-object reversal & 200 \\
 & Context Error & Errors in modifiers such as time, location, quantity, etc. & 200 \\
\cmidrule(l){1-4}
\multirow{3}{*}{\shortstack[l]{Logic}} & Reasoning Chain Error & Logical gaps, skipped steps, or contradictory reasoning & 200 \\
 & Discourse Relation & Errors in chronological order or causal relationships & 200 \\
 & Context Inconsistency & Conclusion conflicts with previous settings & 200 \\

\midrule
% --- Track 2: Data Alignment ---
\rowcolor{TableGreen}
\multicolumn{4}{l}{\textbf{Data Alignment (DA)}} \\\midrule
\multirow{4}{*}{\shortstack[l]{Simple Alignment \& Verification}} & Value Modification & Random arithmetic operations on original values & 150 \\
 & Unit Modification & Inconsistent numerical scales in financial values & 150 \\
 & Comparison Relation & Tampering with logical words (e.g., "increase/decrease") & 150 \\
 & Synonym Replacement & Using financial synonym dictionary to test grounding & 150 \\
\cmidrule(l){1-4}
\multirow{4}{*}{\shortstack[l]{Complex Numerical Calculation}} & Single Date Multi-Indicator & Composite calculation of multiple indicators for one date & 200 \\
 & Multi-Date Single Indicator & Cross-date comparison and calculation for one indicator & 200 \\
 & Multi-Date Multi-Indicator & Cross-date calculation for multiple indicators & 200 \\
 & Cross-Company/Indicator & Ranking, Max/Min, Group comparison, etc. & 200 \\
\cmidrule(l){1-4}
Rule-Driven Reasoning & / & Quantitative rule evaluation over indicators & 400 \\

\midrule
% --- Track 3: Deep Insight ---
\rowcolor{TablePurple}
\multicolumn{4}{l}{\textbf{Deep Insight (DI)}} \\\midrule
Justification \& Causal Depth & / & Conformity to business logic and clarity of causal chains & 300 \\
\cmidrule(l){1-4}
Factuality \& Context & / & Accuracy of evidence-based factual support & 300 \\
\cmidrule(l){1-4}
Completeness \& Comparative Analysis & / & Identification of risks, limitations, and counterarguments & 300 \\
\cmidrule(l){1-4}
Architecture Richness \& Strictness & / & Professional logical framework and conceptual depth & 300 \\
\bottomrule
\end{tabular}
}
\end{table}

\section{Instructions for Manual Examination}

In this section, we delineate the manual examination protocols established to safeguard the quality and evaluative credibility of \textbf{FinReasoning}. To mitigate potential biases inherent in automated pipelines and establish a definitive ``Gold Standard,'' we implemented a quality control system led by two financial experts, each with 5–10 years of professional experience spanning investment analysis and equity research report writing. In Section~\ref{sec:manual_data_construction}, we detail the expert verification process for data construction, encompassing the evidentiary grounding, logical cross-checking, and manual refinement of the 4,800 evaluation samples. In Section~\ref{sec:manual_judge_verification}, we present the expert auditing procedure for the \textbf{LLM-as-a-Judge} component. Through expert blind reviews and quantitative agreement analysis, we validate the effectiveness and robustness of the automated scoring system. These manual examination guidelines, executed by domain practitioners with substantial frontline experience, provide the foundational assurance for the benchmark's professional rigor.

\subsection{Expert Verification Process For Data Construction}
\label{sec:manual_data_construction}
To ensure the reliability of FinReasoning in terms of task definition, answer quality, and financial professionalism, we conducted expert manual verification over all samples constructed in the QA construction stage. The \textbf{Semantic Consistency (SC)} track contains 1,800 samples, the \textbf{Data Alignment (DA)} track contains 1,800 samples, and the \textbf{Deep Insight (DI)} track contains 1,200 samples, for a total of 4,800 QA instances. We invited two experts with experience in financial analysis and data annotation to independently review and cross-revise all samples under a unified verification protocol, with the goal of ensuring professional rigor, task adjudicability, and the stability of the gold-standard answers.

In practice, we adopted a five-stage verification pipeline: \textit{Review -- Judgment -- Consensus -- Recheck -- Revision}.

\textbf{1. Review.} The two experts independently examined all samples. For the SC track, they mainly checked whether the injected errors had clear financial diagnostic significance, whether the overall coherence of the original text was preserved, and whether the error type labels, error locations, and correction suggestions were accurate. For the DA track, they focused on whether the fields, date ranges, and sample IDs required by each question were consistent with the underlying database records, and whether the numerical answers, retrieval paths, and structured formats conformed to the predefined rules. For the DI track, they examined whether the evidence fully anchored an analyzable event, whether the question genuinely required open-ended analysis, and whether the ground-truth answer demonstrated factual grounding, causal reasoning, and critical analysis. The detailed verification guidelines are provided in the colorbox below.

\begin{tcolorbox}[auto outer arc, boxrule=0.5pt, colback=white!90!yellow, left=1mm, right=1mm, top=1mm, bottom=1mm]
\fontsize{8pt}{9pt}\selectfont
\textbf{I. Semantic Consistency:}
\begin{itemize}[left=15pt, labelsep=3pt]
    \item Diagnostic Perturbation Verification: Verify whether the 1–3 injected errors carry clear financial diagnostic significance while preserving the overall coherence of the text. Perturbations that are overly obvious (e.g., trivial spelling mistakes) or overly obscure (i.e., even experts cannot reach consensus) should be flagged and removed.\
    \item Annotation Precision Check: Examine whether the classification of error types is accurate, and whether the location of the erroneous sentence within the long text is identified with character-level precision.
    \item Validity of Correction Suggestions: Confirm that the ``corrected text" removes the error while preserving all remaining original content word for word, and that its accompanying ``Reason" conforms to financial auditing standards and is logically persuasive.
\end{itemize}
\textbf{II. Data Alignment:}
\begin{itemize}[left=15pt, labelsep=3pt]
    \item Consistency with Numerical Gold Standards: Based on predefined indicator formulas and financial execution rules, experts must cross-check the gold-standard answer against the raw records in the underlying A-share database to verify its uniqueness and precision on an item-by-item basis.
    \item Logical Verification of Retrieval Path: Verify whether the fields, date ranges, and sample identifiers (IDs) selected by the model in complex calculations fully match the logic of the question, thereby preventing the model from arriving at the correct number by chance through an incorrect retrieval path.
    \item Format Compliance Review: Ensure that all generated data pairs strictly follow the predefined structured format (e.g., SQL statements or JSON objects) so as to support automated statistics of execution success rate.
\end{itemize}

\textbf{III. Deep Insight:}
\begin{itemize}[left=15pt, labelsep=3pt]
    \item Factual Grounding in Evidence: Rigorously verify that every argument generated by the model can be traced to explicit evidence in the provided research report or literature. The model must not introduce external hallucinations or unsupported speculation.
    \item Evaluation of Causal Chain Completeness: Using the three-step framework of ``event → mechanism → impact", assess whether the model reveals the underlying economic logic. Responses that fail to establish logical connections or merely provide surface-level summaries should be marked as low-quality.
    \item Multi-perspective and Hedging Check: Pay particular attention to whether the model identifies potential execution risks or countervailing constraints. Samples that exhibit excessive optimism or pessimism without critical reasoning should be revised or removed.
\end{itemize}

\normalsize
\end{tcolorbox}

\textbf{2. Judgment.} Based on the same standardized criteria, the two experts independently assigned each sample a \textit{Valid} or \textit{Invalid} label and recorded revision suggestions. A \textit{Valid} label indicates that the sample satisfies the release requirements in terms of task definition, evidential support, gold-standard answer quality, and format specification. An \textit{Invalid} label indicates that the sample contains critical defects, such as unnatural error injection, a non-unique numerical reasoning chain, an overly superficial analytical question, insufficient factual grounding in the gold answer, or an incomplete reasoning loop.

\textbf{3. Consensus.} For samples on which the two experts disagreed, the research team organized item-by-item discussion to align the final decision. The discussion focused on issues such as whether an injected error was diagnostically meaningful without being overly explicit, whether the data retrieval path was unique and reproducible, and whether the gold answer truly reflected deep analysis rather than surface-level summarization. This process not only resolved final labels, but also refined the boundary conditions of the verification criteria, which in turn constrained subsequent data revision.

\textbf{4. Recheck.} All revised samples underwent second-round verification by at least one expert to confirm that the original issues had been properly resolved without introducing new ambiguities or formatting errors. For the open-ended analytical questions in the DI track, the re-check stage additionally required verification that the gold-standard answer simultaneously satisfied causal depth, factual accuracy, critical thinking, and structural completeness, to avoid including samples that were formally correct but not substantively judgeable in final set.

\textbf{5. Revision.} Based on the experts' feedback, we systematically revised the samples, including rewriting question statements, replacing or expanding evidence, correcting gold-standard answers, updating error labels, adjusting data retrieval ranges, and unifying output formats.

Under this standardized protocol, the two experts independently assigned \textit{Valid}/\textit{Invalid} labels to all samples, and we quantified their agreement using Cohen's Kappa:
\[
\kappa=\frac{p_o-p_e}{1-p_e},
\]
where \(p_o\) is observed agreement and \(p_e\) is chance agreement from marginal label distributions.

In our setting, \(N=4800\), and the two experts agreed on \(4624\) samples, yielding
\[
p_o=\frac{4624}{4800}=0.9633.
\]
The marginal label counts were:
\[
\text{Expert A: Valid}=4479,\ \text{Invalid}=321,
\]
\[
\text{Expert B: Valid}=4381,\ \text{Invalid}=419.
\]
Accordingly, the expected agreement is
\[
p_e
=
p_{A,V}p_{B,V}+p_{A,I}p_{B,I}
=
\frac{4479}{4800}\cdot\frac{4381}{4800}
+
\frac{321}{4800}\cdot\frac{419}{4800}
\approx 0.8561.
\]
Therefore,
\[
\kappa
=
\frac{0.9633-0.8561}{1-0.8561}
\approx 0.7454.
\]

The high inter-expert agreement supports the reliability of the manual verification protocol, while the consensus and re-check stages further mitigate remaining discrepancies.

\subsection{Expert Verification Process For LLM-as-a-Judge}\label{sec:manual_judge_verification}

The purpose of this stage is to assess whether the \textit{LLM-as-a-Judge} outputs provide reliable evaluations of model responses. This differs from expert verification during QA construction, which focuses on the quality of the benchmark data itself. We randomly sampled \(20\%\) of the \textit{LLM-as-a-Judge} results for each model. According to the benchmark design, only the Semantic Consistency (SC) and Deep Insight (DI) tracks rely on \textit{LLM-as-a-Judge} evaluation. Therefore, expert review at this stage was restricted to judge outputs from these two tracks. With full model coverage over all SC and DI questions, this yielded 11,400 reviewed model--judgment cases in total.

The overall workflow followed the same five-stage protocol as in Section~\ref{sec:manual_data_construction}, which is \textit{Review -- Judgment -- Consensus -- Recheck -- Revision}. The verification criteria were redefined for the \textit{LLM-as-a-Judge} setting. Experts were instructed to examine whether the judge outputs were rubric-faithful, evidence-grounded, and discriminative with respect to answer quality.

\paragraph{Verification Guidelines for Experts.}
For each sampled case, the experts were provided with the question, the gold answer, the model answer, the scoring rubric, and the corresponding \textit{LLM-as-a-Judge} output. They determined whether the judge output should be labeled as \textit{Valid} or \textit{Invalid} according to the following guidelines.
\begin{tcolorbox}[auto outer arc, boxrule=0.5pt, colback=white!90!yellow, left=1mm, right=1mm, top=1mm, bottom=1mm]

\fontsize{8pt}{9pt}\selectfont
\textbf{I. Rubric Fidelity.}
Experts should examine whether the judge's assigned scores and explanations are consistent with the predefined rubric, including the intended scoring dimensions, severity boundaries, and decision logic. A judge output should be marked as invalid if it assigns obviously inappropriate scores, ignores key rubric requirements, or bases its judgment on criteria not specified in the rubric.

\textbf{II. Evidence Grounding.}
Experts should verify whether the judge's reasoning is grounded in the actual content of the gold answer and the model answer. The judge should not introduce unsupported assumptions, hallucinated deficiencies, or irrelevant evaluation criteria. Its explanation should be traceable to concrete qualities or flaws in the evaluated response, rather than generic evaluative language.

\textbf{III. Discriminative Validity.}
Experts should assess whether the judge can meaningfully distinguish high-quality answers from low-quality ones. In particular, the judge should reward substantive correctness, grounded reasoning, and rubric-relevant strengths, while penalizing key omissions, unsupported claims, broken logic, or superficial responses. Outputs that are excessively lenient or harsh due only to fluency, verbosity, or superficial formatting should be treated as invalid.

\textbf{IV. Track-Specific Verification for SC.}
For the SC track, experts should focus on whether the judge correctly evaluates semantic error diagnosis and correction quality. This includes checking whether the judge appropriately scores the identification of injected errors, the validity of the error explanation, and the fidelity of the corrected text to the gold-standard correction. Special attention should be paid to whether the judge fails to penalize missed key errors, false-positive error claims, incorrect error typing, or unnecessary over-correction.

\textbf{5. Track-Specific Verification for DI.}
For the DI track, experts should focus on whether the judge accurately assesses deep analytical quality according to the corresponding rubric dimensions, such as factual grounding, causal depth, comparative or critical reasoning, and structural rigor. High scores should only be assigned when the model answer demonstrates evidence-based financial analysis, coherent causal linkage, and balanced reasoning. Superficially fluent but shallow, ungrounded, or overly generic responses should not receive inflated scores.

\textbf{6. Explanation Quality.}
Experts should verify whether the judge's explanation is specific, decision-relevant, and aligned with the assigned score. A valid explanation should identify the key strengths or weaknesses that actually justify the score under the rubric. Explanations that are vague, templated, or disconnected from the score should lower confidence in the judge output and may lead to an invalid label.

\textbf{7. Validity Decision Rule.}
A judge output is labeled as \textit{Valid} if its score profile and justification are broadly consistent with the rubric and with expert judgment on the same case. A judge output is labeled as \textit{Invalid} if it contains major scoring errors, unsupported reasoning, weak grounding, or systematic bias toward superficial response characteristics.

\normalsize
\end{tcolorbox}

Following the same protocol as in Section~\ref{sec:manual_data_construction}, the two experts independently assigned \textit{Valid}/\textit{Invalid} labels to the sampled \textit{LLM-as-a-Judge} outputs, and we quantified their agreement using Cohen's Kappa:
\[
\kappa=\frac{p_o-p_e}{1-p_e},
\]
where \(p_o\) is observed agreement and \(p_e\) is chance agreement from marginal label distributions.

Since only the Semantic Consistency (SC) and Deep Insight (DI) tracks require \textit{LLM-as-a-Judge} evaluation, we report the agreement statistics separately for these two tracks.

\paragraph{Semantic Consistency.}
For the SC track, the total number of reviewed judge outputs was
\[
N_{SC}=19\times 1800\times 10\% \times 2=6840.
\]
In this setting, the two experts agreed on \(6552\) sampled cases, yielding
\[
p_o=\frac{6552}{6840}=0.9725.
\]
The marginal label counts were:
\[
\text{Expert A: Valid}=6345,\ \text{Invalid}=495,
\]
\[
\text{Expert B: Valid}=6376,\ \text{Invalid}=464.
\]
Accordingly, the expected agreement is
\[
p_e
=
p_{A,V}p_{B,V}+p_{A,I}p_{B,I}
=
\frac{6345}{6840}\cdot\frac{6376}{6840}
+
\frac{495}{6840}\cdot\frac{464}{6840}
\approx 0.8696.
\]
Therefore,
\[
\kappa_{SC}
=
\frac{0.9725-0.8696}{1-0.8696}
\approx 0.7891.
\]

\paragraph{Deep Insight.}
For the DI track, the total number of reviewed judge outputs was
\[
N_{DI}=19\times 1200\times 10\% \times 2=4560.
\]
In this setting, the two experts agreed on \(4302\) sampled cases, yielding
\[
p_o=\frac{4302}{4560}=0.9434.
\]
The marginal label counts were:
\[
\text{Expert A: Valid}=3964,\ \text{Invalid}=596,
\]
\[
\text{Expert B: Valid}=3883,\ \text{Invalid}=677.
\]
Accordingly, the expected agreement is
\[
p_e
=
p_{A,V}p_{B,V}+p_{A,I}p_{B,I}
=
\frac{3964}{4560}\cdot\frac{3883}{4560}
+
\frac{596}{4560}\cdot\frac{677}{4560}
\approx 0.7596.
\]
Therefore,
\[
\kappa_{DI}
=
\frac{0.9434-0.7596}{1-0.7596}
\approx 0.7646.
\]

These results still indicate stronger inter-expert agreement on SC than on DI. This difference is expected given the distinct nature of the two tasks. SC primarily evaluates whether errors are correctly detected and corrected, making its judgment criteria relatively objective and well-defined. In contrast, DI assesses the depth and quality of financial analysis, where experts must further evaluate causal reasoning, critical thinking, and structural rigor, leading to greater room for subjective variation.

\paragraph{Fusion Strategy.}
To determine the optimal fusion weights for the two judge models, we conducted a grid search over the weighting parameter $\alpha \in [0, 1]$ with a step size of 0.02, where the fused score for each sample is computed as $\alpha \cdot s_{\text{DeepSeek}} + (1 - \alpha) \cdot s_{\text{Qwen}}$. The optimal $\alpha$ for each track was selected to maximize agreement with expert annotations on the 20\% reviewed samples. Table~\ref{tab:fusion_weights} reports the resulting weights and the corresponding expert agreement rates.

\begin{table}[ht]
\centering
\scriptsize
\setlength{\tabcolsep}{2pt}
\caption{Optimal fusion weights and expert agreement by track.}
\label{tab:fusion_weights}
\begin{tabular}{lccc}
\toprule
\textbf{Track} & \textbf{DeepSeek-V3 Weight}& \textbf{Qwen3-235B Weight}\\
\midrule
Semantic Consistency (SC) & 0.58 & 0.42 \\
Deep Insight (DI)         & 0.60 & 0.40 \\
\bottomrule
\end{tabular}
\end{table}

\section{Extended Experimental Results and Sensitivity Analysis}\label{sec:extended_experiment}

\subsection{Computational Cost and Scalability Analysis}\label{sec:computational_cost}

Table~\ref{tab:computational_cost_by_track} presents supplementary computational cost statistics for each evaluation track across different task categories, offering an overview of the benchmarking runtime characteristics. It enables readers to understand the distribution of input/output sizes and inference latency in the test procedures. Notably, this table is not intended for model capability comparison or as a basis for the main conclusions; rather, it enhances the completeness and reproducibility of the experimental report, serving as a reference for experiment replication, evaluation cost estimation, and system load analysis.

\begin{table}
\centering
\scriptsize
\setlength{\tabcolsep}{3pt}
\caption{Computational Cost and Scalability Statistics Across Tracks (average per sample). We report input/output/total tokens and latency for each category.}
\label{tab:computational_cost_by_track}
% \small
\resizebox{\textwidth}{!}{
\begin{tabular}{lcccc}
\toprule
Category / Dimension & Input Tokens (Avg) & Output Tokens (Avg) & Total Tokens (Avg) & Latency (ms, Avg) \\
\midrule

% --- Track 1: Semantic Consistency ---
\rowcolor{TableBlue}
\multicolumn{5}{l}{\textbf{Semantic Consistency }} \\\midrule
Terminology Confusion & 5078.2 & 3385.0 & 8463.3 & 105947.4 \\
Terminology Inconsistency & 5322.2 & 4286.5 & 9608.8 & 96807.5 \\
Improper Usage & 5049.1 & 3540.5 & 8589.6 & 88156.3 \\
Entity Error & 5024.5 & 3462.3 & 8486.7 & 86872.0 \\
Relation Error & 5045.9 & 3573.3 & 8619.2 & 87682.7 \\
Context Error & 5218.9 & 3940.3 & 9159.2 & 119598.2 \\
Reasoning Chain Error & 5476.4 & 3894.4 & 9370.8 & 87639.3 \\
Discourse Relation & 5454.1 & 4592.5 & 10046.6 & 100362.0 \\
Context Inconsistency & 5246.8 & 4072.8 & 9319.7 & 95265.3 \\
\textbf{Average} & \textbf{5212.9} & \textbf{3860.8} & \textbf{9073.8} & \textbf{96481.2} \\

\midrule
% --- Track 2: Data Alignment ---
\rowcolor{TableGreen}
\multicolumn{5}{l}{\textbf{Data Alignment}} \\\midrule
Verification\_stage1 & 2096.3 & 1156.7 & 3253.0 & 24761.2 \\
Verification\_stage2 & 509.2 & 1601.3 & 2110.4 & 29014.0 \\
Calculation\_stage1 & 2098.7 & 1622.0 & 3720.7 & 49554.6 \\
Calculation\_stage2 & 2585.3 & 4913.2 & 7498.5 & 121589.2 \\
Reasoning\_stage1 & 2110.7 & 2235.1 & 4345.9 & 41187.1 \\
Reasoning\_stage2 & 11271.7 & 10658.3 & 21930.0 & 184907.1 \\
\textbf{Average} & \textbf{3445.3} & \textbf{3697.8} & \textbf{7143.1} & \textbf{75168.9} \\

\midrule
% --- Track 3: Deep Insight ---
\rowcolor{TablePurple}
\multicolumn{5}{l}{\textbf{Deep Insight}} \\\midrule
Justification \& Causal Depth & 2898.2 & 367.9 & 3266.0 & 18014.0 \\
Factuality \& Contextualization & 2700.6 & 375.7 & 3076.3 & 16779.5 \\
Completeness \& Comparative Analysis & 2486.0 & 387.5 & 2873.6 & 16413.8 \\
Architectural Richness \& Strictness & 2888.7 & 325.2 & 3213.9 & 14143.2 \\
\textbf{Average} & \textbf{2743.4} & \textbf{364.1} & \textbf{3107.5} & \textbf{16337.6} \\

\bottomrule
\end{tabular}
}
\end{table}

\subsection{Prompt Sensitivity and Evaluation Robustness}
\label{sec:prompt_sensitivity}
To assess the robustness of the \textit{LLM-as-a-Judge} evaluation, we rephrase the evaluation prompt while preserving the original task definition, scoring criteria, and rubric semantics. We introduce controlled prompt perturbations by rewriting the instruction wording and reordering the rubric items without changing the scoring dimensions or evaluation targets. Table~\ref{tab:sc_prompt_sensitivity}, \ref{tab:sc_prompt_fact}, \ref{tab:sc_prompt_logic}, \ref{tab:sc_prompt_di} reports the model scores under the original and rephrased judge prompts.

We use rank correlation to test whether rephrasing the judge prompt changes the model ranking. We first compute Spearman’s $\rho$ between the rankings produced by the original and rephrased prompts:

$$
\rho = 1 - \frac{6\sum_{i=1}^{n} d_i^2}{n(n^2-1)}
$$

where $d_i$ is the rank difference of model $i$ between the two prompt settings. Spearman’s $\rho$ measures whether the overall ranking order is preserved.

We then compute Kendall’s $\tau_b$:

$$
\tau_b = \frac{C-D}{\sqrt{(C+D+T_x)(C+D+T_y)}}
$$

where $C$ and $D$ denote the numbers of concordant and discordant model pairs, and $T_x$ and $T_y$ denote tied pairs in the two rankings. Kendall’s $\tau_b$ measures the consistency of pairwise ordering between models.

The average Spearman’s $\rho$ is 0.9478 and the average Kendall’s $\tau_b$ is 0.8651, showing that prompt rephrasing has little effect on the final model ranking.

\begin{table}
\centering
\setlength\tabcolsep{25pt}
\scriptsize
\caption{Prompt sensitivity of LLM-as-a-Judge evaluation under original and rephrased prompts for the \textbf{Terminology} category in the Semantic Consistency track.}
\label{tab:sc_prompt_sensitivity}
\resizebox{\textwidth}{!}{
\begin{tabular}{lcccc}
\toprule
\multicolumn{1}{c}{\multirow{2}{*}{Models}} & \multicolumn{2}{c}{Error Explanation} & \multicolumn{2}{c}{Correct Content} \\
\cmidrule(lr){2-3} \cmidrule(lr){4-5}
\multicolumn{1}{c}{}                        & Origin           & Rephrase           & Origin          & Rephrase          \\
\midrule
\rowcolor{TableYellow}\multicolumn{5}{l}{\textbf{Close-Source}}                                                                                \\
\midrule
GPT-5                                       & 85.09            & 83.31              & 88.12           & 87.43             \\
GPT-4o                                      & 48.24            & 49.78              & 50.34           & 49.94             \\
Gemini-3-pro                                & 79.00            & 81.45              & 82.17           & 81.39             \\
Claude-sonnet-4.5                           & 80.65            & 79.69              & 83.82           & 80.92             \\
GLM-4.6                                     & 67.91            & 64.22              & 69.45           & 65.33             \\
Qwen3-Max                                   & 62.20            & 66.45              & 64.78           & 62.82             \\
Kimi-K2                                     & 63.16            & 65.97              & 65.46           & 69.47             \\
Doubao-Seed-1.8                             & 71.53            & 69.21              & 73.69           & 75.27             \\
\midrule
\rowcolor{TableOrange}\multicolumn{5}{l}{\textbf{Open-Source General}}           
\\\midrule
DeepSeek-R1                                 & 58.22            & 60.33              & 59.48           & 57.41             \\
DeepSeek-V3                                 & 44.87            & 41.42              & 47.97           & 50.52             \\
Intern-S1                                   & 50.81            & 51.37              & 50.87           & 54.13             \\
Qwen3-8B                                    & 48.81            & 49.95              & 49.66           & 50.31             \\
Qwen3-32B                                   & 51.53            & 52.63              & 50.66           & 53.92             \\
Qwen3-235B                                  & 51.28            & 54.89              & 52.87           & 51.09             \\
Llama-3.1-8B                                & 13.49            & 14.93              & 11.66           & 14.86             \\
Llama-3.1-70B                               & 16.39            & 17.51              & 10.97           & 14.35             \\\midrule
\rowcolor{TableRed}\multicolumn{5}{l}{\textbf{Open-Source Financial}}                                                                        \\\midrule
Fin-R1                                      & 26.73            & 28.53              & 26.04           & 30.17             \\
TouchstoneGPT                               & 2.88             & 2.32               & 2.36            & 1.95              \\
DianJin-R1-7B                               & 25.59            & 24.18              & 17.34           & 15.23    \\     
\bottomrule
\end{tabular}}

\end{table}

\begin{table}
\centering
\setlength\tabcolsep{25pt}
\scriptsize
\caption{Prompt sensitivity of LLM-as-a-Judge evaluation under original and rephrased prompts for the \textbf{Fact} category in the Semantic Consistency track.}
\label{tab:sc_prompt_fact}
\resizebox{\textwidth}{!}{
\begin{tabular}{lcccc}
\toprule
\multicolumn{1}{c}{\multirow{2}{*}{Models}} & \multicolumn{2}{c}{Error Explanation} & \multicolumn{2}{c}{Correct Content} \\
\cmidrule(lr){2-3} \cmidrule(lr){4-5}
\multicolumn{1}{c}{}                        & Origin           & Rephrase           & Origin          & Rephrase          \\\midrule
\rowcolor{TableYellow}\multicolumn{5}{l}{\textbf{Close-Source}}                                                                                 \\\midrule
GPT-5                                       & 80.14            & 79.76              & 82.38           & 85.43             \\
GPT-4o                                      & 60.08            & 62.63              & 62.74           & 63.79             \\
Gemini-3-pro                                & 81.43            & 78.12              & 83.35           & 87.07             \\
Claude-sonnet-4.5                           & 82.12            & 83.45              & 85.22           & 87.80             \\
GLM-4.6                                     & 78.74            & 81.27              & 80.78           & 78.90             \\
Qwen3-Max                                   & 76.00            & 74.05              & 81.62           & 81.04             \\
Kimi-K2                                     & 72.43            & 73.16              & 74.68           & 75.12             \\
Doubao-Seed-1.8                             & 80.19            & 77.49              & 82.36           & 80.36             \\\midrule
\rowcolor{TableOrange}\multicolumn{5}{l}{\textbf{Open-Source General}}                                                                          \\\midrule
DeepSeek-R1                                 & 70.46            & 68.90              & 71.61           & 71.32             \\
DeepSeek-V3                                 & 67.11            & 66.51              & 71.19           & 68.53             \\
Intern-S1                                   & 64.58            & 63.63              & 65.88           & 66.76             \\
Qwen3-8B                                    & 60.70            & 59.77              & 61.55           & 63.79             \\
Qwen3-32B                                   & 68.10            & 70.99              & 68.09           & 72.01             \\
Qwen3-235B                                  & 61.48            & 64.82              & 67.17           & 66.99             \\
Llama-3.1-8B                                & 11.94            & 8.11               & 11.30           & 9.45              \\
Llama-3.1-70B                               & 22.70            & 25.94              & 15.02           & 10.96             \\\midrule
\rowcolor{TableRed}\multicolumn{5}{l}{\textbf{Open-Source Financial}}                                                                        \\\midrule
Fin-R1                                      & 23.44            & 26.81              & 22.31           & 26.44             \\
TouchstoneGPT                               & 2.71             & 4.90               & 2.05            & 4.01              \\
DianJin-R1-7B                               & 31.69            & 33.87              & 22.39           & 18.44      \\\bottomrule      
\end{tabular}}
\end{table}
\vspace{-5pt}

\begin{table}
\centering
\setlength\tabcolsep{25pt}
\scriptsize
\caption{Prompt sensitivity of LLM-as-a-Judge evaluation under original and rephrased prompts for the \textbf{Logic} category in the Semantic Consistency track.}
\label{tab:sc_prompt_logic}
\resizebox{\textwidth}{!}{
\begin{tabular}{lcccc}
\toprule
\multicolumn{1}{c}{\multirow{2}{*}{Models}} & \multicolumn{2}{c}{Error Explanation} & \multicolumn{2}{c}{Correct Content} \\
\cmidrule(lr){2-3} \cmidrule(lr){4-5}
\multicolumn{1}{c}{}                        & Origin           & Rephrase           & Origin          & Rephrase          \\\midrule
\rowcolor{TableYellow}\multicolumn{5}{l}{\textbf{Close-Source}}                                                                                 \\\midrule
GPT-5                                       & 87.89            & 84.74              & 90.47           & 87.67             \\
GPT-4o                                      & 67.15            & 68.27              & 68.71           & 72.19             \\
Gemini-3-pro                                & 87.15            & 89.34              & 88.32           & 91.65             \\
Claude-sonnet-4.5                           & 89.76            & 88.80              & 91.82           & 89.14             \\
GLM-4.6                                     & 83.23            & 82.98              & 84.88           & 82.04             \\
Qwen3-Max                                   & 84.29            & 84.75              & 86.74           & 82.52             \\
Kimi-K2                                     & 82.63            & 78.92              & 83.26           & 79.54             \\
Doubao-Seed-1.8                             & 84.33            & 83.81              & 86.56           & 88.03             \\\midrule
\rowcolor{TableOrange}\multicolumn{5}{l}{\textbf{Open-Source General}}                                                                          \\\midrule
DeepSeek-R1                                 & 67.54            & 64.23              & 68.25           & 72.69             \\
DeepSeek-V3                                 & 69.79            & 71.31              & 72.05           & 71.10             \\
Intern-S1                                   & 66.39            & 62.97              & 68.17           & 66.95             \\
Qwen3-8B                                    & 67.46            & 64.68              & 69.97           & 65.91             \\
Qwen3-32B                                   & 70.33            & 66.10              & 68.34           & 66.37             \\
Qwen3-235B                                  & 67.10            & 65.06              & 66.63           & 63.54             \\
Llama-3.1-8B                                & 17.14            & 16.80              & 10.48           & 7.76              \\
Llama-3.1-70B                               & 25.10            & 28.36              & 16.65           & 14.16             \\\midrule
\rowcolor{TableRed}\multicolumn{5}{l}{\textbf{Open-Source Financial}}                                                                        \\\midrule
Fin-R1                                      & 36.39            & 36.81              & 30.97           & 33.57             \\
TouchstoneGPT                               & 1.93             & 2.09               & 1.95            & 1.88              \\
DianJin-R1-7B                               & 37.66            & 35.12              & 28.15           & 29.32            \\\bottomrule
\end{tabular}}
\end{table}
\vspace{-5pt}

\begin{table}
\centering
\setlength\tabcolsep{9pt}
\scriptsize
\caption{Prompt sensitivity of LLM-as-a-Judge evaluation under original and rephrased prompts for the Deep Insight track.}
\label{tab:sc_prompt_di}
\begin{tabular}{lcccccccc}
\toprule
\multicolumn{1}{c}{\multirow{2}{*}{Models}} & \multicolumn{2}{c}{ARS} & \multicolumn{2}{c}{F\&C} & \multicolumn{2}{c}{C\&A} & \multicolumn{2}{c}{JCD} \\\cmidrule(lr){2-3} \cmidrule(lr){4-5} \cmidrule(lr){6-7}\cmidrule(lr){8-9}
\multicolumn{1}{c}{}                        & Origin    & Rephrase    & Origin     & Rephrase    & Origin     & Rephrase    & Origin    & Rephrase    \\\midrule
\rowcolor{TableYellow}\multicolumn{9}{l}{\textbf{Close-Source}}                                                                                                             \\\midrule
GPT-5                                       & 92.39     & 90.29       & 84.58      & 87.63       & 86.71      & 83.48       & 81.80     & 82.97       \\
GPT-4o                                      & 79.96     & 84.19       & 77.64      & 75.83       & 74.78      & 75.33       & 76.93     & 72.87       \\
Gemini-3-pro                                & 82.14     & 79.78       & 79.01      & 80.95       & 76.65      & 75.88       & 80.67     & 79.56       \\
Claude-sonnet-4.5                           & 92.19     & 96.52       & 87.07      & 86.48       & 84.51      & 80.57       & 96.17     & 96.83       \\
GLM-4.6                                     & 92.58     & 91.03       & 88.10      & 85.63       & 87.33      & 87.12       & 96.06     & 97.43       \\
Qwen3-Max                                   & 85.80     & 88.79       & 75.95      & 75.96       & 75.78      & 73.56       & 82.89     & 83.66       \\
Kimi-K2                                     & 93.38     & 93.92       & 87.42      & 90.51       & 89.47      & 85.26       & 96.27     & 96.16       \\
Doubao-Seed-1.8                             & 93.52     & 95.83       & 87.15      & 85.37       & 86.17      & 87.86       & 95.73     & 99.33       \\\midrule
\rowcolor{TableOrange}\multicolumn{9}{l}{\textbf{Open-Source General}}                                                                                                      \\\midrule
DeepSeek-R1                                 & 88.03     & 90.13       & 84.63      & 87.00       & 80.03      & 80.63       & 89.76     & 91.96       \\
DeepSeek-V3                                 & 81.99     & 85.49       & 81.03      & 78.65       & 77.89      & 78.39       & 86.36     & 83.92       \\
Intern-S1                                   & 84.36     & 86.40       & 82.08      & 79.18       & 81.59      & 77.85       & 88.00     & 83.12       \\
Qwen3-8B                                    & 69.33     & 71.25       & 51.09      & 48.56       & 53.85      & 49.78       & 57.37     & 56.64       \\
Qwen3-32B                                   & 73.61     & 71.91       & 61.67      & 63.73       & 60.29      & 61.31       & 65.70     & 67.44       \\
Qwen3-235B                                  & 75.38     & 78.41       & 63.79      & 66.52       & 63.65      & 59.86       & 67.48     & 71.86       \\
Llama-3.1-8B                                & 39.09     & 39.17       & 30.83      & 29.72       & 31.33      & 29.62       & 32.13     & 28.73       \\
Llama-3.1-70B                               & 42.64     & 41.50       & 36.47      & 34.09       & 34.73      & 37.71       & 36.84     & 37.13       \\\midrule
\rowcolor{TableRed}\multicolumn{9}{l}{\textbf{Open-Source Financial}}                                                                                                    \\\midrule
Fin-R1                                      & 77.90     & 75.99       & 71.83      & 69.43       & 73.53      & 75.65       & 71.47     & 71.82       \\
TouchstoneGPT                               & 38.33     & 37.39       & 28.70      & 30.75       & 30.78      & 34.67       & 34.22     & 30.77       \\
DianJin-R1-7B                               & 77.14     & 81.65       & 63.61      & 63.64       & 73.27      & 77.10       & 69.14     & 67.43 \\\bottomrule     
\end{tabular}
\end{table}
\vspace{-10pt}

\subsection{Inter-Judge Consistency and Bias Analysis}
\label{sec:inter-judge_consisency}
\begin{figure}[htbp]
    \centering
    \includegraphics[width=1\textwidth]{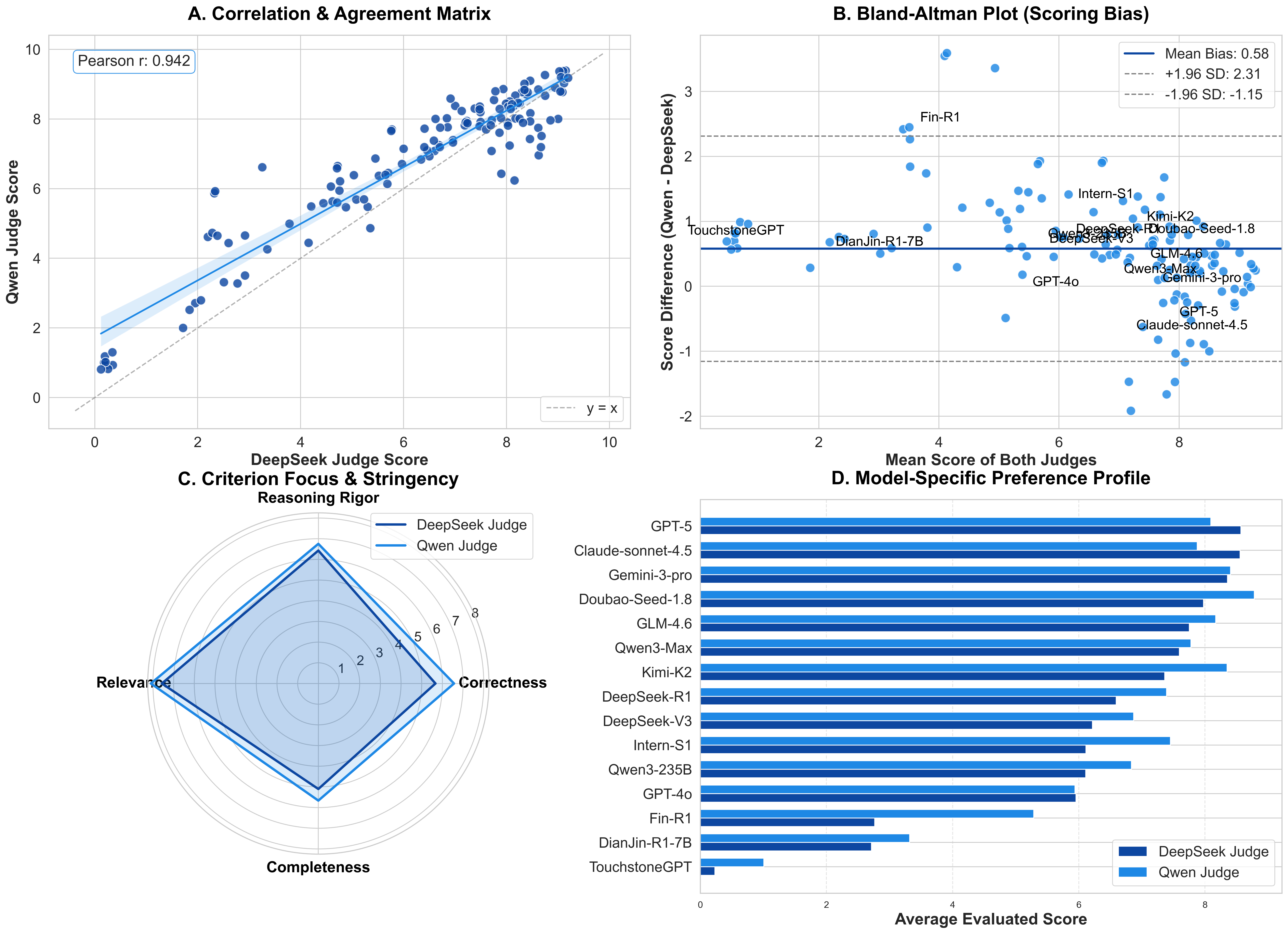}
    \caption{Comparative Evaluation of DeepSeek-V3 and Qwen3-235B in Semantic Consistency track.}
    \label{fig:llm_judge_analysis_sc}
%\vspace{-10pt}
\end{figure}

\begin{figure}[htbp]
    \centering
    \includegraphics[width=1\textwidth]{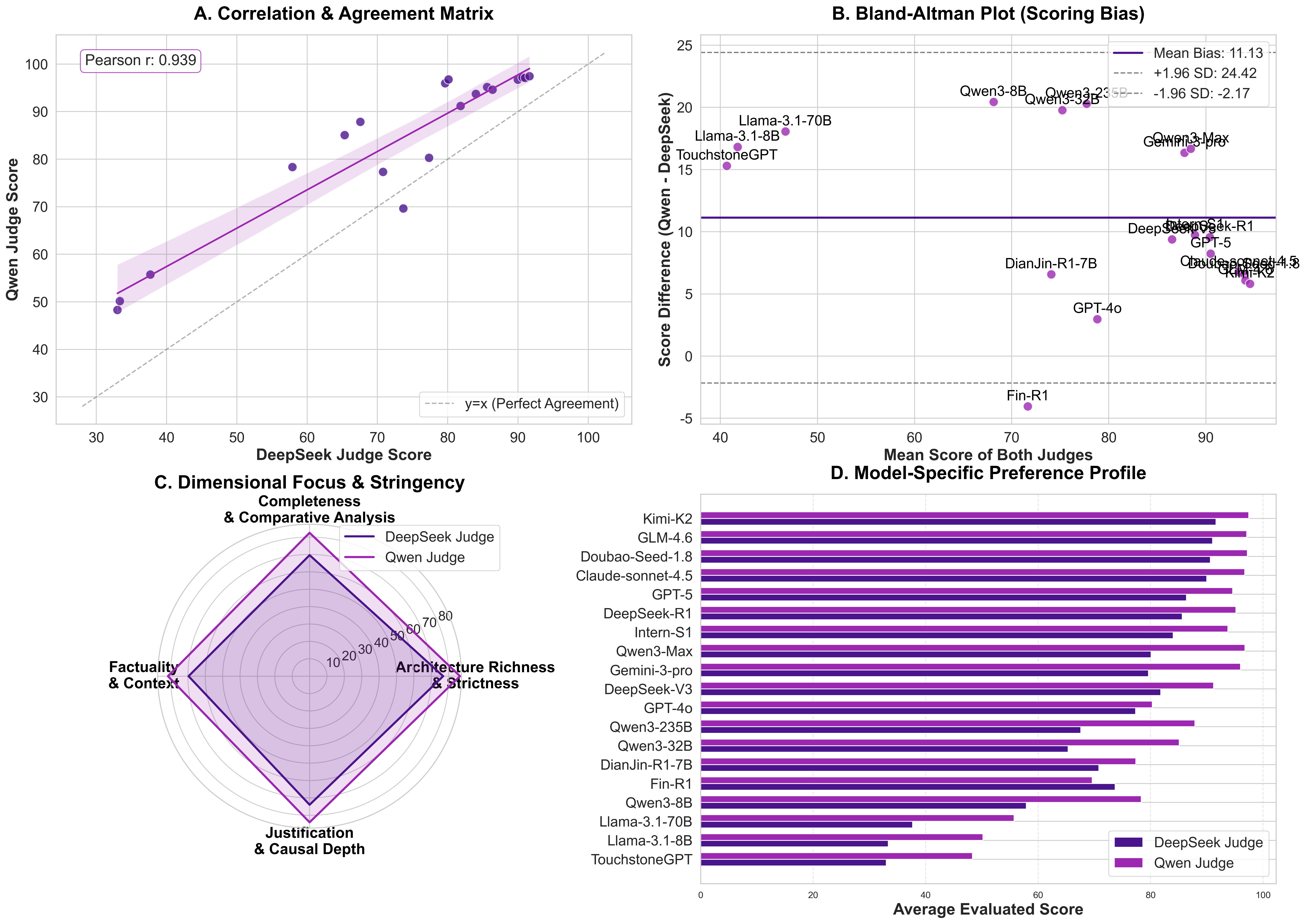}
    \caption{Comparative Evaluation of DeepSeek-V3 and Qwen3-235B in Deep Insight track.}
    \label{fig:llm_judge_analysis_di}
%\vspace{-10pt}
\end{figure}

Figure~\ref{fig:llm_judge_analysis_sc},\ref{fig:llm_judge_analysis_di} presents a multi-dimensional analysis of systematic differences between two LLM-based judges (DeepSeek-V3 and Qwen3-235B) in the Semantic Consistency evaluation track and Deep Insight. (A) The correlation scatter plot with Pearson correlation coefficient reveals the agreement level between the two judges' overall scoring patterns. (B) The Bland-Altman plot visualizes scoring bias distribution, highlighting which specific models exhibit the largest discrepancies between the two judges and detecting potential systematic favoritism. (C) The radar chart illustrates the dimensional focus and stringency profile of both judges. (D) The paired horizontal bar chart directly compares the absolute scores assigned by each judge to all 19 evaluated models, revealing model-specific preference profiles and potential self-bias patterns (e.g., whether DeepSeek-V3 as judge exhibits favoritism toward DeepSeek models). Together, these analyses assess the robustness and identify potential sources of bias in using large language models as evaluation judges.

First, when using DeepSeek-V3 and Qwen3-235B as judge models, both maintain high consistency in overall model ranking, although their absolute scores are not equivalent. The Pearson correlation coefficients for the two LLM judges in the SC and DI tracks reach 0.942 and 0.939, respectively. The scatter plot (Figure A) and the Bland-Altman plot (Figure B) jointly indicate a stable systematic bias between the two judges rather than random fluctuations. Qwen3-235B exhibits more lenient scoring across both tracks, with average scores consistently higher than those assigned by DeepSeek-V3.

In the SC track, this discrepancy is particularly evident at the dimensional level. Figure~\ref{fig:llm_judge_analysis_sc}.C shows that Qwen3-235B assigns higher scores across all four dimensions, with the largest difference observed in \emph{Correctness}, followed by \emph{Completeness}. This suggests that Qwen imposes relatively weaker constraints on factual precision and coverage completeness, whereas DeepSeek-V3 is more conservative and more sensitive to fine-grained inconsistencies. Further model-level comparisons (Figure ~\ref{fig:llm_judge_analysis_sc}.D) reveal that both judges exhibit a degree of in-group preference, but DeepSeek-V3 demonstrates a stronger self-model bias: DeepSeek family models receive notably higher scores when evaluated by DeepSeek-V3, while Qwen's preference for Qwen family models, though present, is weaker. The largest discrepancies between the two judges tend to occur for low-to-medium quality or borderline models, whereas discrepancies are smaller for strong general-purpose models such as Gemini and GPT-4o, indicating that bias primarily concentrates on critical samples.

The DI track exhibits a similar but not identical pattern. The two judges maintain high consistency in overall ranking, yet a stable absolute score difference persists, suggesting that the discrepancy primarily stems from judge ``stringency" rather than ranking reversal. Figure~\ref{fig:llm_judge_analysis_di}.D shows that Qwen3-235B is more lenient on most models, while DeepSeek-V3 assigns lower absolute scores. Synthesizing the results from both SC and DI tracks, the robustness of LLM judges should be understood as stability at the ranking level rather than consistency in raw scores. Therefore, for scenarios requiring comparability of absolute scores, judge calibration or ensemble evaluation should be introduced to mitigate systematic bias and enhance scoring reliability.

\subsection{Statistical Robustness Analysis}\label{sec:bootstrap_ci}

To ascertain whether the performance disparities observed among models are statistically stable, we conducted a robust uncertainty estimation using 95\% bootstrap confidence intervals (CIs). Specifically, for each model across all tracks and sub-dimensions, we performed 1,000 iterations of resampling with replacement. This methodology quantifies the variability of performance estimates under test-sample perturbation.

\begin{figure}[htbp]
    \centering
    \includegraphics[width=\linewidth]{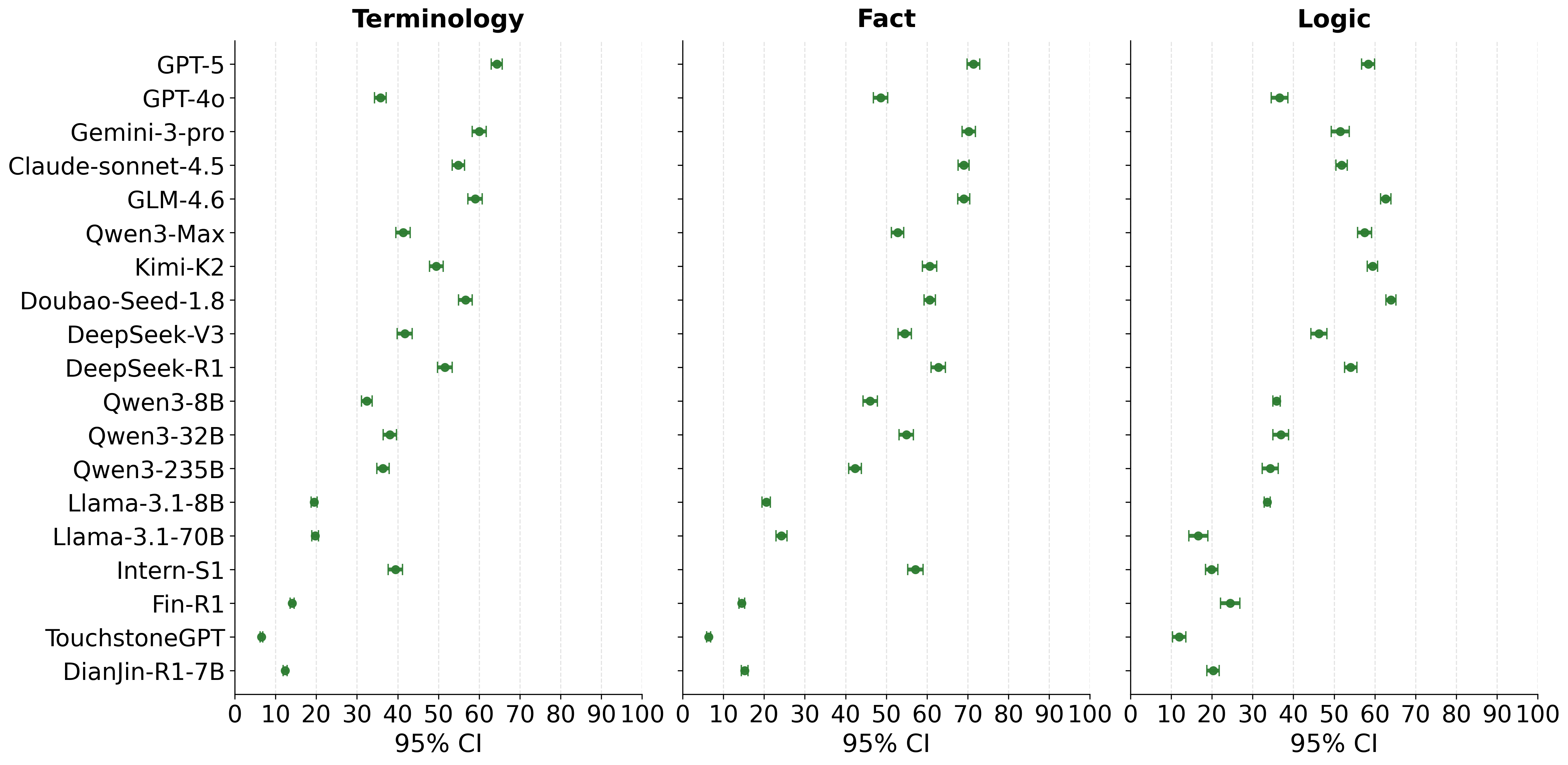}
    \caption{Bootstrap 95\% confidence intervals across Semantic Consistency.}
    \label{fig:sc_bootstrap_ci_intervals}
\end{figure}

\paragraph{Semantic Consistency Track (Figure~\ref{fig:sc_bootstrap_ci_intervals})}
Claude-sonnet-4.5 achieved the most superior and stable performance in the SC track (95\% CI: [57.14, 60.00]). Crucially, the non-overlapping nature of its CI with that of GPT-5 [50.82, 52.86] furnishes robust statistical evidence that its lead is not a byproduct of sample noise but a consistent capability advantage. A similar statistical separation was observed between GPT-5 and GLM-4.6. Conversely, the slight overlap between GPT-5 and Gemini-3-pro suggests their performance delta should be interpreted with caution, as it may be sensitive to specific sample distributions.

\vspace{-5pt}

\begin{figure}[htbp]
    \centering
    \includegraphics[width=\linewidth]{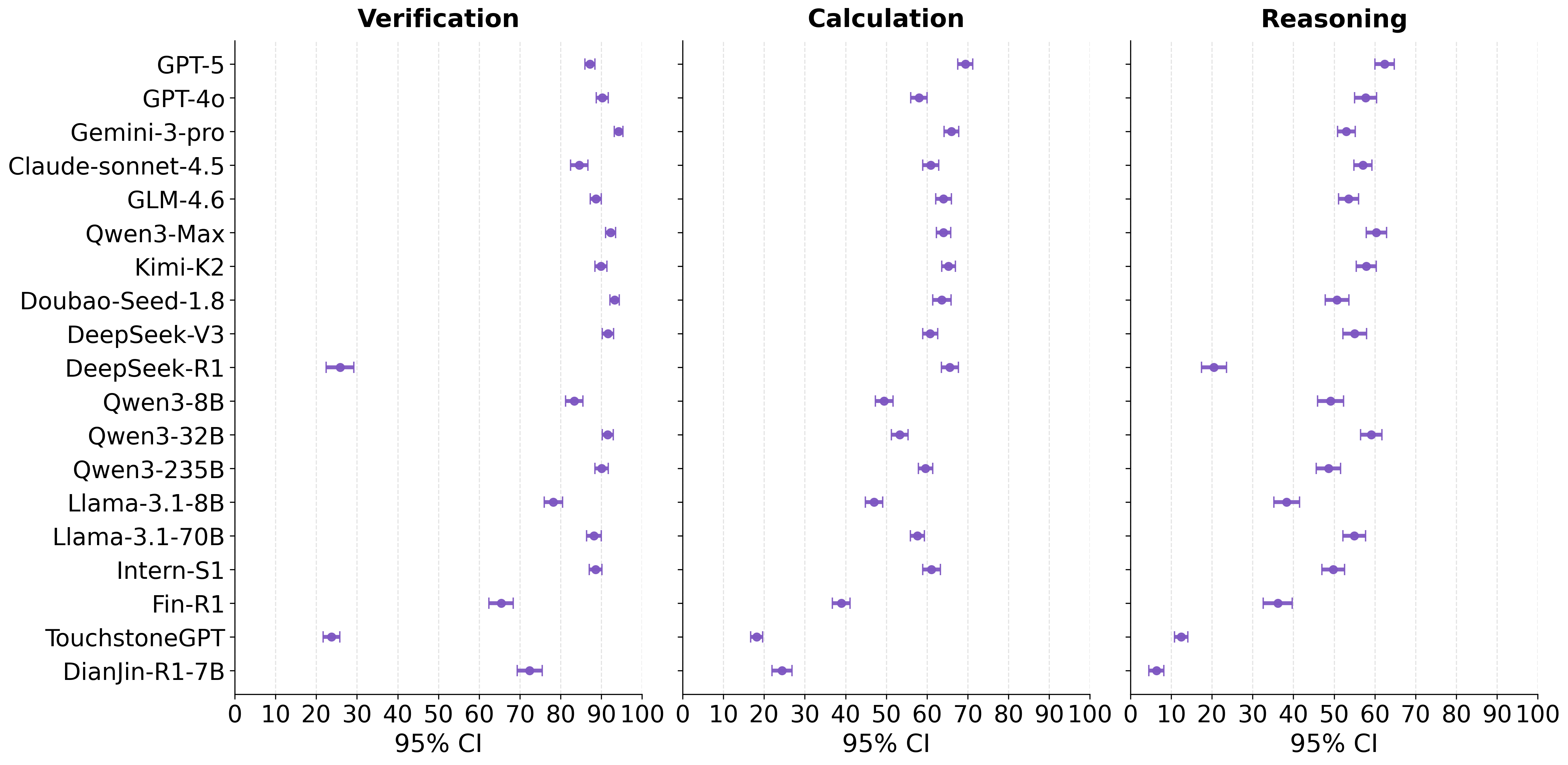}
    \caption{Bootstrap 95\% confidence intervals across Data Alignment.}
    \label{fig:da_bootstrap_ci_intervals}
\end{figure}
\vspace{-5pt}

\paragraph{Data Alignment Track (Figure~\ref{fig:da_bootstrap_ci_intervals})}
GPT-5 exhibited the most resilient overall performance with a 95\% CI of [71.19, 74.80], closely followed by clustered intervals for Qwen3-Max, Gemini-3-pro, and Kimi-K2. Notably, capabilities vary across sub-tasks: Gemini-3-pro secured the highest interval in Verification [93.14, 95.29], whereas GPT-5 maintained a statistically significant lead in Calculation [67.58, 71.22] and Reasoning [59.99, 64.77]. This divergence suggests that frontier models possess distinct specialized strengths in factual grounding versus numerical logic.

\begin{figure}[htbp]
    \centering
    \includegraphics[width=\linewidth]{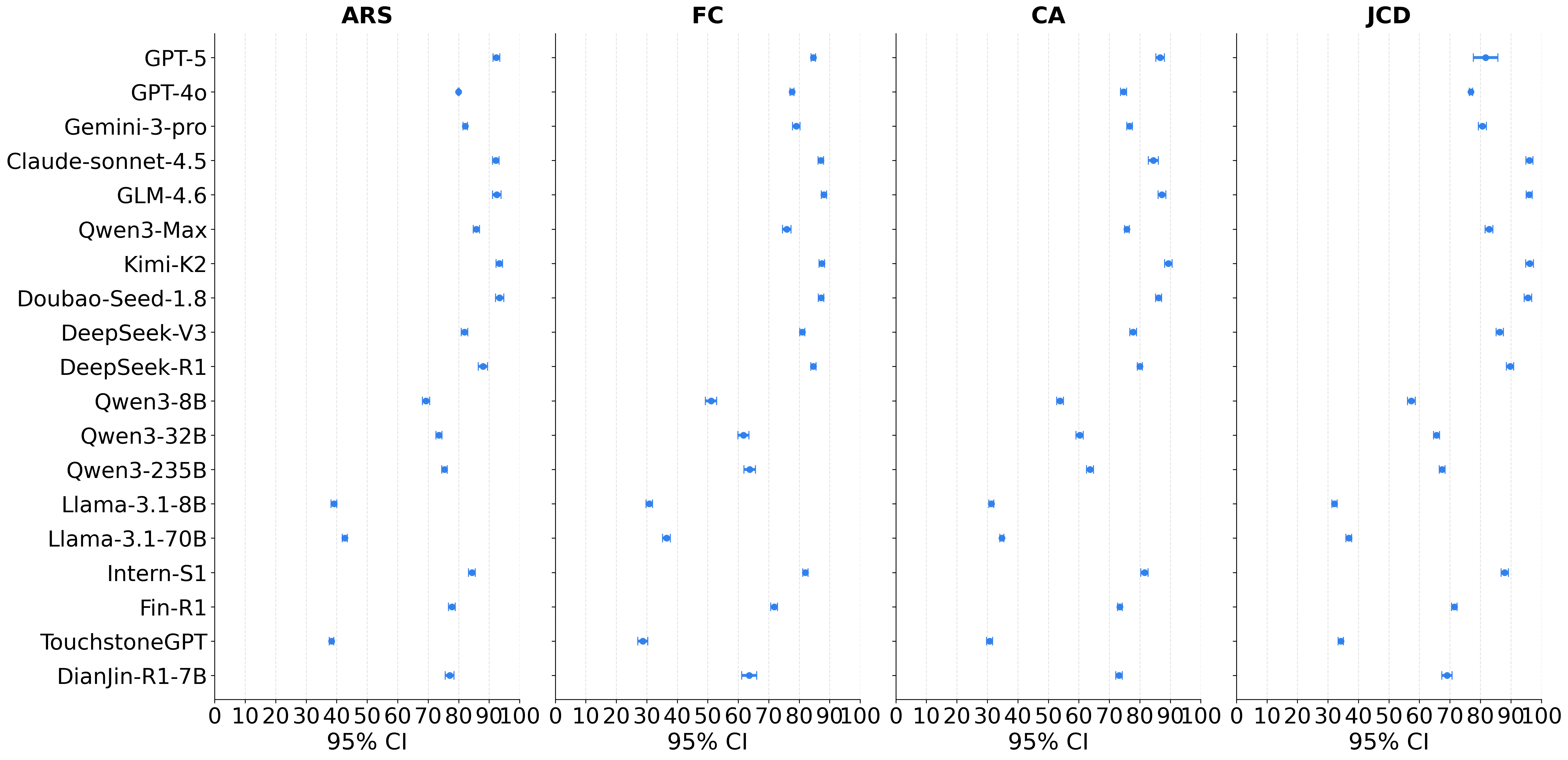}
    \caption{Bootstrap 95\% confidence intervals across Deep Insight.}
    \label{fig:di_bootstrap_ci_intervals}
\end{figure}

\paragraph{Deep Insight Track (Figure~\ref{fig:di_bootstrap_ci_intervals})}
Kimi-K2 led this track with an average CI of [90.48, 92.69], with GLM-4.6, Doubao-Seed-1.8, and Claude-sonnet-4.5 following closely. The high concentration and narrow width of these intervals indicate that their analytical depth is robust against sample fluctuations. Within the \textit{Justification \& Causal Depth} category, the CI lower bounds for these top-tier models consistently approached or exceeded 94, underscoring their steady superiority in constructing complex causal chains.

Overall, the bootstrap analysis underscores that model superiority is highly track-dependent. The complete separation of confidence intervals provides strong empirical support for the performance gaps between frontier models, while overlapping intervals highlight where competitive gradients remain marginal. 
% This granular diagnostic reinforces the necessity of our benchmark for capturing the subtle nuances of financial reasoning capabilities.

\vspace{-5pt}

\section{Failure studies}\label{sec:failure_studies}

In this section, we provide an in-depth exploration of failure patterns across models within the \textbf{FinReasoning} benchmark to identify core capability bottlenecks in current Large Language Models (LLMs) when executing financial research tasks. In Section~\ref{sec:failure_cases}, we present representative failure cases for each evaluation track---Semantic Consistency, Data Alignment, and Deep Insight---to qualitatively illustrate model limitations in semantic reliability, numerical grounding, and analytical depth. Following this, Section~\ref{sec:failure_taxonomy} establishes a rigorous Failure Mode Taxonomy. By synthesizing quantitative statistics and qualitative analysis, we systematically dissect performance deficiencies in key dimensions, including failures in data query and error explanation, and lack of causal reasoning depth. These diagnostic insights provide a clear strategic path for the future targeted enhancement of model capabilities in the financial domain.

\subsection{Failure Cases}\label{sec:failure_cases}

\subsubsection{SC Failure Case: Factual Relation Error}\label{case:SC_Failure_Case}
\begin{tcolorbox}[
  enhanced,
  breakable,
  colback=white,
  colframe=TableDarkBlue,
  colbacktitle=TableBlue,
  coltitle=black,
  title=\textbf{SC Track -- Factual Relation Error},
  fonttitle=\small,
  boxrule=0.6pt,
  arc=1.5mm,
  left=1.5mm,right=1.5mm,top=1mm,bottom=1mm
]

{\footnotesize

\textbf{Question / Input}

\begin{innerblock}
\fontsize{7pt}{6pt}\selectfont
Reference Chunks + Modified Long Text + Question.
\end{innerblock}

\textbf{Gold Answer}

\begin{innerblock}
\fontsize{7pt}{6pt}\selectfont
\textbf{Error Exists:} Yes

\textbf{Identified factual relation errors (3):}
% \begin{itemize}[leftmargin=4mm,itemsep=0.4mm,topsep=0.5mm]
%     \item \textbf{已完成再融资流程} $\rightarrow$ \textbf{已启动再融资流程}
%     \item \textbf{同比下降约50\%} $\rightarrow$ \textbf{同比增长约50\%}
%     \item \textbf{并立即重仓买入} $\rightarrow$ \textbf{deleted}
% \end{itemize}
\begin{itemize}[leftmargin=4mm,itemsep=0.4mm,topsep=0.5mm]
    \item \textbf{Refinancing process \color{red}{completed}} $\rightarrow$ \textbf{Refinancing process \color{red}{initiated}}
    \item \textbf{{\color{red}Decreased} by approximately 50\% year-on-year} $\rightarrow$ \textbf{{\color{red}Increased} by approximately 50\% year-on-year}
    \item \textbf{Immediately bought a heavy position} $\rightarrow$ \textbf{\color{red}{deleted}}
\end{itemize}

\textbf{Reference explanation}
The gold answer identifies three factual relation errors, all of which arise from altered event states, reversed data trends, or intensified recommendation strength. These are not stylistic issues; rather, they change the semantic relation between facts and therefore directly distort the financial meaning of the passage.
\end{innerblock}

\textbf{Model Output}
\begin{innerblock}
\fontsize{7pt}{6pt}\selectfont
\textbf{Error Exists:} Yes

\textbf{Predicted factual relation errors (4):}
% \begin{itemize}[leftmargin=4mm,itemsep=0.4mm,topsep=0.5mm]
%     \item \textbf{同比下降约50\%} $\rightarrow$ \textbf{具体数据未在外部知识库中提供，无法确认}
%     \item \textbf{美联储潜在调整补充杠杆率规则} $\rightarrow$ \textbf{美联储已调整补充杠杆率规则}
%     \item \textbf{再融资审核进程符合预期} $\rightarrow$ \textbf{再融资审核仍在进行中}
%     \item \textbf{AI硬件对散热、电磁屏蔽等功能性材料提出新需求} $\rightarrow$ \textbf{AI技术应用对公司的影响未明确提及具体需求}
% \end{itemize}
\begin{itemize}[leftmargin=4mm,itemsep=0.4mm,topsep=0.5mm]
    \item \textbf{Decreased by approximately 50\% year-on-year} $\rightarrow$ \textbf{{\color{red}Specific data not provided} in external knowledge base; cannot confirm}
    \item \textbf{{\color{red}Potential adjustment} by the Federal Reserve to the Supplementary Leverage Ratio (SLR) rules} $\rightarrow$ \textbf{The Federal Reserve {\color{red}has adjusted} the Supplementary Leverage Ratio (SLR) rules}
    \item \textbf{Refinancing review process {\color{red}proceeding as expected}} $\rightarrow$ \textbf{Refinancing review is {\color{red}still ongoing}}
    \item \textbf{AI hardware {\color{red}creates new demands} for functional materials such as thermal management and electromagnetic shielding} $\rightarrow$ \textbf{The impact of AI technology application on the company {\color{red}does not explicitly mention specific demands}}
\end{itemize}

\textbf{Model explanation}
The model attributes most of its error judgments to whether the claimed relation is explicitly supported by an external knowledge base. As a result, it shifts the task from \emph{text-grounded factual relation checking} to \emph{external factual verification}, which leads to both false positives and major omissions relative to the gold standard.
\end{innerblock}

\textbf{Failure Pattern Summary}

\begin{innerblock}
\fontsize{7pt}{6pt}\selectfont
This case illustrates a representative \textbf{task-misalignment} failure in the SC track. Instead of checking whether the modified text changes the factual relation expressed in the source passage, the model repeatedly relies on the absence or presence of support in an external knowledge base. This strategy causes it to miss two core gold errors (the event-state shift from \emph{started} to \emph{completed}, and the added recommendation \emph{buy heavily immediately}) while introducing three unsupported false positives. The result is a fluent but fundamentally misaligned prediction that departs from the intended benchmark objective.
\end{innerblock}

\textbf{Objective Metrics}

\begin{center}
\tiny
\setlength{\tabcolsep}{7pt} 
\begin{tabularx}{\linewidth}{p{3.0cm}p{0.5cm}Y}
\toprule
\textbf{Metric Group} & \textbf{Value} & \textbf{Observation} \\
\midrule
Ground-truth error count & 3 & The gold answer contains three factual relation errors. \\
Predicted error count & 4 & The model over-predicts the number of errors. \\
Sentence-level F1 & 0.286 & Very low overlap between predicted and gold error sentences. \\
Sentence-level Precision & 0.250 & Most predicted errors are false positives. \\
Sentence-level Recall & 0.333 & The model captures only one out of three gold errors. \\
Explanation BERTScore-F1 & 0.579 & The explanation is moderately similar at the surface level, but not well aligned in substance. \\
Corrected-text BERTScore-F1 & 0.538 & The revised text diverges substantially from the gold correction target. \\
Corrected-text SimCSE & 0.314 & Semantic similarity is low, indicating correction drift. \\
\bottomrule
\end{tabularx}
\end{center}

\textbf{LLM-as-a-Judge Evaluation}

\begin{center}
\tiny
\setlength{\tabcolsep}{7pt} 
\begin{tabularx}{\linewidth}{p{1.2cm}p{0.5cm}Yp{0.5cm}Y}
\toprule
\multirow{2}{*}{\textbf{Dimension}} & \multicolumn{2}{c}{\textbf{Error Reasoning}} & \multicolumn{2}{c}{\textbf{Corrected Text}} \\
\cmidrule(lr){2-3}\cmidrule(lr){4-5}
& \textbf{Score} & \textbf{Explanation} & \textbf{Score} & \textbf{Explanation} \\
\midrule
Correctness & 4.0 & The model's identified errors and justifications deviate substantially from the gold answer. It treats knowledge-base absence as the main basis for judging factual relation errors, which leads to incorrect decisions on several key cases. & 5.0 & The revised text partially fixes one relation error, but several corrections either contradict the gold answer or replace precise facts with vague fallback statements. \\
Reasoning Rigor & 3.0 & The reasoning chain is weak because it is built on a flawed premise: ``not found in the external knowledge base'' is treated as evidence of error. This does not match the intended relation-checking task. & 4.0 & The revised text is consistent with the model's own faulty reasoning, but that reasoning itself is not rigorous under the benchmark definition. \\
Relevance & 6.0 & The explanations are mostly related to the model's predicted error points, but several of those points are themselves off-target with respect to the true benchmark objective. & 7.0 & The revisions remain focused on the model's own predicted errors, with little irrelevant editing, but they are only partially relevant to the gold corrections. \\
Completeness & 2.0 & The model misses two central gold errors while adding multiple unsupported ones, resulting in very poor coverage of the true factual relation changes. & 3.0 & The revised text leaves key gold errors unresolved and therefore remains incomplete from the perspective of the reference answer. \\
\bottomrule
\end{tabularx}
\end{center}

}
\end{tcolorbox}

\subsubsection{DA Failure Case: Reasoning Error}\label{case:DA_Failure_Case}
\begin{tcolorbox}[
  enhanced,
  breakable,
  colback=white,
  colframe=TableDarkGreen,
  colbacktitle=TableGreen,
  coltitle=black,
  title=\textbf{DA Track -- KDJ Rule Misexecution},
  fonttitle=\small,
  boxrule=0.6pt,
  arc=1.5mm,
  left=1.5mm,right=1.5mm,top=1mm,bottom=1mm
]

{\footnotesize

\textbf{Question / Input}

\begin{innerblock}
\fontsize{7pt}{6pt}\selectfont
Structured financial reasoning over database records: given the queried OHLC data and the benchmark rule \texttt{KDJ003}, determine whether a bullish KDJ golden-cross signal exists for Shenzhen Guohua Network Security Technology Co., Ltd. from 2023-06-26 to 2023-08-03, using 20 days of historical context.
\end{innerblock}

\textbf{Gold Answer}

\begin{innerblock}
\fontsize{7pt}{6pt}\selectfont
\textbf{Expected Answer:} No

\textbf{Gold rule condition:}
\begin{itemize}[leftmargin=4mm,itemsep=0.4mm,topsep=0.5mm]
    \item \textbf{Current $K < 30$}
    \item \textbf{Current $K > D$}
    \item \textbf{Previous-day $K < D$}
\end{itemize}

\textbf{Gold explanation}
The benchmark computes the full KDJ sequence from the required historical window and evaluates the rule only at the final decision point. The final condition is \textbf{False}, so the correct answer is \textbf{No}. Although the series fluctuates during the interval, the last point does not satisfy the complete bullish-crossover rule.
\end{innerblock}

\textbf{Model Output}

\begin{innerblock}
\fontsize{7pt}{6pt}\selectfont
\textbf{Model Answer:} Yes

\textbf{Model's claimed reasoning:}
\begin{itemize}[leftmargin=4mm,itemsep=0.4mm,topsep=0.5mm]
    \item Uses standard KDJ formulas and states that it searches for a golden-cross signal within the interval.
    \item Claims that on \textbf{2023-06-29}, $K$ crossed above $D$ and therefore a bullish opportunity exists.
    \item Final cited data only contains \textbf{2 IDs} (\texttt{569572}, \texttt{574591}), far fewer than the full sequence required for rule execution.
\end{itemize}

\textbf{Model explanation}
The model presents a detailed-looking technical-analysis derivation, but its reasoning is not aligned with the benchmark objective. Instead of performing final-point rule judgment over the full required sequence, it rewrites the task into searching for whether any crossover occurred inside the interval, and supports this conclusion with severely insufficient data grounding.
\end{innerblock}

\textbf{Failure Pattern Summary}

\begin{innerblock}
\fontsize{6pt}{5pt}\selectfont
This case illustrates a representative \textbf{task-misalignment} failure in the DA track. The model does not simply make a numerical mistake; rather, it changes the task from \emph{final-state rule verification} to \emph{interval-wide event search}. At the same time, its query-planning stage is broadly reasonable, but its final answer uses only two data rows, creating a sharp disconnect between retrieval and reasoning. The result is a \textbf{pseudo-precise} explanation: the output looks mathematically rigorous, yet it is unsupported by the data actually used and does not execute the gold rule faithfully.
\end{innerblock}

\textbf{Objective Metrics}

\begin{center}
\tiny
\setlength{\tabcolsep}{3pt}
\begin{tabularx}{\linewidth}{p{2.0cm}p{0.7cm}Y}
\toprule
\textbf{Metric Group} & \textbf{Value} & \textbf{Observation} \\
\midrule
Answer Match & False & The gold answer is \textbf{No}, but the model predicts \textbf{Yes}. \\
Fields Match & True & The model correctly identifies the required fields: \texttt{daily\_high}, \texttt{daily\_low}, and \texttt{daily\_close}. \\
Field Precision & 1.000 & No irrelevant pricing fields are used in the final structured answer. \\
Field Recall & 1.000 & All required pricing fields are covered. \\
Data IDs Match & False & The retrieved/used IDs do not align with the benchmark-required data span. \\
Data ID Precision & 1.000 & The two explicitly cited IDs are valid records, but coverage is extremely narrow. \\
Data ID Recall & 0.069 & Only 2 out of 29 required records are actually used, indicating severe under-retrieval in the final reasoning stage. \\
Queried Rows & 43 & Stage-1 retrieval window is broad enough, but the final reasoning fails to use it faithfully. \\
\bottomrule
\end{tabularx}
\end{center}

\textbf{Structured Failure Diagnosis}

\begin{center}
\tiny
\setlength{\tabcolsep}{3pt}
\begin{tabularx}{\linewidth}{p{3.8cm}Y}
\toprule
\textbf{Failure Mode} & \textbf{Manifestation in This Case} \\
\midrule
Temporal Semantic Drift & The benchmark requires a final-point decision, whereas the model reformulates the task as checking whether a crossover happened at any point in the interval. \\
Retrieval--Reasoning Decoupling & The query plan broadly retrieves the correct time window, but the final answer cites only two data IDs, which is insufficient to reconstruct the required KDJ sequence. \\
Pseudo-precise Reasoning & The model outputs detailed formulas and seemingly exact numerical claims, yet these calculations are not supported by the data explicitly used in the answer. \\
Rule Misexecution & The gold rule requires three constraints to hold jointly, especially the oversold condition at the final point; the model instead relies on a generic crossover template. \\
\bottomrule
\end{tabularx}
\end{center}

}
\end{tcolorbox}

\subsubsection{DI Failure Case: Completeness \& Comparative Analysis}\label{case:DI_Failure_Case}
\begin{tcolorbox}[
  enhanced,
  breakable,
  colback=white,
  colframe=TableDarkPurple,
  colbacktitle=TablePurple,
  coltitle=black,
  title=\textbf{DI Track -- Completeness \& Comparative Analysis},
  fonttitle=\small,
  boxrule=0.6pt,
  arc=1.5mm,
  left=1.5mm,right=1.5mm,top=1mm,bottom=1mm
]
{\footnotesize
\textbf{Evidence}
\begin{innerblock}
\fontsize{7pt}{6pt}\selectfont
{
% 关于储能行业，研报指出：工业和信息化部等八部门发布《新型储能制造业高质量发展行动方案》，要求到2027年，新型储能制造业规模和下游需求基本匹配，培育生态主导型企业3-5家。方案引导行业格局优化，实现高质量发展，并建议关注储能龙头企业及超级电容/钠电/液流产业链。同时，研报在风险提示中明确指出：储能装机需求不及预期，新技术迭代不及预期，行业竞争加剧。
Regarding the energy storage industry, the research report notes that eight departments, including the Ministry of Industry and Information Technology, have released the ``Action Plan for High-Quality Development of the New Energy Storage Manufacturing Industry." The plan requires that by 2027, the scale of the new energy storage manufacturing industry and downstream demand be essentially matched, and that three to five ecologically dominant enterprises be cultivated. The plan aims to guide the optimization of the industry landscape and achieve high-quality development, and it suggests paying attention to leading energy storage companies as well as the supercapacitor, sodium battery, and liquid flow battery industrial chains. Meanwhile, the report explicitly highlights in its risk disclosures that energy storage installation demand may fall short of expectations, the pace of new technology iteration may fail to meet expectations, and industry competition may intensify.
}
\end{innerblock}

\textbf{Question}
\begin{innerblock}
\fontsize{7pt}{6pt}\selectfont
{
% 请基于上述背景事实，进行一项辩证分析：政策引导行业“高质量发展”和“格局优化”的正面预期，与风险提示中“行业竞争加剧”的潜在制约之间，存在何种内在冲突？在何种市场条件下，“竞争加剧”的风险可能显著削弱甚至颠覆“格局优化”的乐观前景？
Based on the factual background presented above, please conduct a dialectical analysis: What inherent conflict exists between the positive expectation of policy-guided ``high-quality development" and ``landscape optimization" and the potential constraint of ``intensified industry competition" as highlighted in the risk disclosures? Under what market conditions might the risk of ``intensified competition" significantly weaken or even overturn the optimistic prospect of ``landscape optimization"?
}
\end{innerblock}

\begin{minipage}[t]{0.49\textwidth}
\textbf{Gold Answer.}
\begin{innerblock}
\fontsize{6pt}{5pt}\selectfont
{
% 政策愿景与市场现实之间存在着典型的“目标-路径”张力，分析如下：\\\\ 1.  冲突的本质：理想化的集中度提升 vs. 现实中的无序竞争。\\   政策目标：通过设定技术标准、鼓励创新和设定培育“生态主导型企业”的目标，旨在推动行业从当前可能存在的同质化、价格战阶段，走向以技术、品牌和规模为核心优势的良性竞争阶段，最终实现“格局优化”（即市场份额向头部集中）。\\   风险现实：“行业竞争加剧”恰恰是当前及未来一段时间行业最显著的特征。大量企业涌入储能赛道，产能快速扩张。在“下游需求”可能不及预期的背景下（另一项风险），供需失衡会加剧价格竞争。企业为了生存和抢占市场份额，可能不得不进行激烈的价格战，侵蚀行业整体利润。\\   内在冲突：政策希望引导企业通过“高质量发展”（如技术升级、产品差异化）来优胜劣汰，但激烈的价格竞争环境可能迫使所有企业，包括潜在龙头，将资源优先用于生存性降价而非长期性研发，这与“高质量发展”的路径背道而驰。\\\\2.  证伪“格局优化”前景的关键市场条件：\\   条件一：需求增速持续低于产能扩张速度。这是最核心的证伪条件。如果储能装机需求（无论是电网侧、发电侧还是用户侧）因经济性、政策落地节奏或电网消纳问题而增长缓慢，行业将长期处于供过于求的状态。在“活下去”的压力下，技术领先的企业可能无法将技术优势转化为定价权和利润，行业整合会延迟，甚至出现“劣币驱逐良币”的现象。\\   条件二：技术路线快速迭代且未形成标准。政策鼓励多种技术路线（超级电容、钠电、液流等）。如果技术迭代速度过快，且没有形成稳定的主流技术标准和供应链，会导致市场碎片化。企业前期投入可能迅速过时，任何一家企业都难以建立持久的技术壁垒和规模效应，从而无法稳定地成长为“生态主导型企业”。\\   条件三：非市场化因素干扰（如地方保护、低价中标）。在激烈的竞争中，一些非技术、非品牌的因素可能扭曲竞争。例如，地方保护主义可能阻碍全国性龙头的市场拓展；一些项目招标过于注重初始投资成本（低价中标），而忽视全生命周期成本和质量，这不利于注重研发和品质的龙头企业胜出。\\\\3.  比较与平衡视角：\\   历史比较：可以参考光伏、动力电池等行业的发展历程。这些行业都经历了政策驱动、产能过剩、价格战、行业洗牌，最终才逐步形成相对稳定的竞争格局。储能行业可能正在重复这一过程，政策是加速器，但无法跳过市场出清的痛苦阶段。\\   论点平衡：政策为行业长期健康发展指明了方向，并可能通过设置准入门槛等方式加速低效产能出清。然而，短期至中期内，“竞争加剧”是确定性很高的现实。投资者需关注企业的真实技术护城河、成本控制能力和现金流状况，而非仅仅依赖政策催生龙头的线性预期。真正的“格局优化”只会发生在那些能够穿越残酷竞争周期、并持续进行有效研发和运营的企业身上。\\\\结论：“格局优化”的乐观前景与“竞争加剧”的风险并非完全对立，但后者是前者的严峻考验。在需求不及预期、技术路线混乱或竞争环境非市场化的条件下，行业可能陷入长期的低利润混战，“格局优化”的进程将被大大延后，甚至出现与政策目标相悖的“逆集中化”现象（众多小企业依靠低价生存）。因此，对储能行业的分析必须动态评估供需关系、技术演进路径和实际竞争强度，而非静态解读政策目标。
There is a typical ``goal-path'' tension between policy vision and market reality. The analysis is as follows:

\textbf{1. Nature of the Conflict: Idealized Concentration vs. Unrestrained Competition in Reality}

\textbf{Policy Goal:} By setting technical standards, encouraging innovation, and setting the goal of cultivating ``ecosystem-dominant enterprises,'' the policy aims to steer the industry away from the current state of homogenization and price wars, toward a stage of healthy competition driven by technology, branding, and scale. Ultimately, this is meant to achieve ``landscape optimization'' (i.e., concentration of market share among leading firms).

\textbf{Risk Reality:} ``Intensified industry competition'' is precisely the most prominent feature of the industry now and for the foreseeable future. A large number of companies are entering the energy storage sector, leading to rapid capacity expansion. Against a backdrop of potentially weaker-than-expected downstream demand (another risk factor), supply-demand imbalances will intensify price competition. To survive and capture market share, companies may be forced into fierce price wars, eroding overall industry profitability.

\textbf{Inherent Conflict:} The policy seeks to guide the industry toward survival of the fittest through ``high-quality development'' (e.g., technological upgrades, product differentiation). However, the intense price competition may force all companies—including potential leaders—to prioritize survival-driven price cuts over long-term R\&D. This runs counter to the path of ``high-quality development.''

\textbf{2. Key Market Conditions That Would Falsify the ``Landscape Optimization'' Outlook}

\textbf{Condition 1: Demand growth persistently lags behind capacity expansion.} This is the most critical falsifying condition. If energy storage installation demand (whether grid-side, generation-side, or user-side) grows slowly due to economic viability, policy implementation pace, or grid absorption issues, the industry will remain in long-term oversupply. Under the pressure to ``stay alive,'' technologically leading companies may fail to translate technical advantages into pricing power and profits. Industry consolidation will be delayed, and there may even be a ``bad money drives out good'' phenomenon.

\textbf{Condition 2: Rapid iteration of technology pathways without standardization.} The policy encourages multiple technology pathways (supercapacitors, sodium-ion, flow batteries, etc.). If technological iteration is too rapid and no stable mainstream technical standards or supply chains emerge, the market will become fragmented. Companies' prior investments may quickly become obsolete, making it difficult for any single firm to establish durable technological barriers and economies of scale—thus hindering stable growth into an ``ecosystem-dominant enterprise.''

\textbf{Condition 3: Distortions from non-market factors (e.g., local protectionism, low-price bidding).} In fierce competition, non-technical and non-brand factors may distort the competitive landscape. For example, local protectionism can hinder the market expansion of national champions. Some project tenders may overemphasize initial investment costs (low-price wins) while neglecting life-cycle costs and quality, which does not favor leading companies focused on R\&D and quality.

\textbf{3. Comparative and Balanced Perspective}

\textbf{Historical Comparison:} Similar evolution can be observed in industries such as photovoltaics and power batteries. Those industries went through policy-driven phases, overcapacity, price wars, and industry shakeouts before gradually forming relatively stable competitive landscapes. The energy storage industry is likely repeating this process. Policy acts as an accelerator but cannot skip the painful phase of market cleansing.

\textbf{Balanced Argument:} Policy charts the direction for long-term healthy industry development and may accelerate the exit of inefficient capacity through measures such as setting access thresholds. However, in the short to medium term, ``intensified competition'' is a highly certain reality. Investors need to focus on companies' genuine technological moats, cost control capabilities, and cash flow—not rely solely on linear expectations of policy-driven champions. Genuine ``landscape optimization'' will only happen for companies that can survive brutal competitive cycles while sustaining effective R\&D and operations.

\textbf{Conclusion}
The optimistic outlook of ``landscape optimization'' and the risk of ``intensified competition'' are not entirely opposed, but the latter poses a severe test for the former. Under conditions of weaker-than-expected demand, chaotic technology pathways, or non-market competition environments, the industry may fall into prolonged low-profit battles. The process of ``landscape optimization'' could be significantly delayed, and even a ``de-concentration'' phenomenon (many small firms surviving on low prices) contrary to policy goals may emerge. Therefore, analysis of the energy storage industry must dynamically assess supply-demand relationships, technology evolution paths, and actual competitive intensity, rather than statically interpreting policy objectives.
}

\end{innerblock}
\end{minipage}
\hfill
\begin{minipage}[t]{0.49\textwidth}
\textbf{Model Answer}
\begin{innerblock}
\fontsize{6pt}{5pt}\selectfont
{
% 1. 政策引导的正面预期\\政策引导下的储能行业高质量发展和格局优化主要体现在以下几个方面：\\- 规模与需求匹配：到2027年，新型储能制造业规模将与下游需求基本匹配，这有助于避免产能过剩或不足的问题，确保行业的可持续发展。\\- 企业培育：培育3-5家生态主导型企业，这些企业在技术、市场等方面具有领先地位，能够引领行业发展，促进技术创新和产业升级。\\- 行业标准与规范：政策引导有助于建立和完善行业标准，提高产品质量和安全性，减少安全隐患，增强消费者信心。\\- 国际合作：通过政策支持，推动储能技术的国际交流与合作，提升中国储能产业在全球市场的竞争力。\\\\2. 风险提示中的行业竞争加剧\\行业竞争加剧可能带来以下挑战：\\- 市场份额争夺：随着更多企业和资本进入储能领域，市场竞争将更加激烈，可能导致价格战，压缩企业利润空间。\\- 技术创新压力：为了在竞争中脱颖而出，企业需要不断投入研发，加快技术迭代速度，增加成本负担。\\- 供应链不稳定：激烈的竞争可能导致供应链管理难度加大，原材料供应紧张或价格上涨，影响生产效率和产品质量。\\- 政策环境不确定性：政策变化可能对行业产生重大影响，如补贴政策调整、环保法规收紧等，增加企业的经营风险。\\\\3. 内在冲突\\正面预期与潜在制约之间的内在冲突主要表现在：\\- 资源分配不均：高质量发展的目标需要大量资源投入，包括资金、人才和技术。如果资源过度集中于少数企业，可能会导致其他中小企业生存困难，从而阻碍整个行业的健康发展。\\- 创新与竞争的悖论：政策鼓励技术创新和高质量发展，但过度的竞争压力可能导致企业过分追求短期利益，忽视长期技术研发，形成恶性循环。\\- 市场准入门槛：政策引导下的行业优化可能设置较高的市场准入门槛，使得新进入者难以突破，进一步固化现有企业的市场地位，抑制新兴力量的成长。\\\\4. 市场条件下的风险放大效应\\在以下市场条件下，“竞争加剧”的风险可能显著削弱甚至颠覆“格局优化”的乐观前景：\\- 宏观经济衰退：经济下行周期可能导致市场需求萎缩，企业盈利能力下降，加剧价格竞争，削弱企业的研发投入能力，延缓技术进步。\\- 政策环境突变：政府对储能行业的支持力度减弱，如取消补贴、收紧环保标准等，将直接打击企业的投资积极性，破坏行业生态平衡。\\- 技术路线分化：储能技术多元化发展，不同技术路径（如锂离子电池、钠离子电池、液流电池等）之间的竞争加剧，可能导致资源分散，无法形成统一的技术标准和市场共识。\\- 供应链断裂：关键原材料供应中断或价格剧烈波动，严重影响生产连续性和成本控制，迫使企业降低产品质量或退出市场，破坏行业稳定。\\\\结论\\政策引导下的储能行业高质量发展和格局优化与行业竞争加剧的风险之间存在复杂的互动关系。在理想情况下，政策能够有效引导资源合理配置，促进技术创新和产业升级；但在现实市场环境中，宏观经济、政策环境和技术路线等因素的不确定性可能放大竞争风险，威胁行业优化成果。因此，需综合考虑多方面因素，制定灵活应对策略，以确保行业健康可持续发展。
\textbf{1. Positive Expectations Under Policy Guidance}

The high-quality development and landscape optimization of the energy storage industry under policy guidance are mainly reflected in the following aspects:

\textbf{Alignment of scale and demand:} By 2027, the scale of the new-type energy storage manufacturing industry will basically match downstream demand, which helps avoid overcapacity or shortage and ensures sustainable industry development.
\textbf{Cultivation of enterprises:} Cultivate 3--5 ecosystem-dominant enterprises. These firms, holding leading positions in technology and markets, can guide industry development, promote technological innovation, and drive industrial upgrading.
\textbf{Industry standards and regulations:} Policy guidance helps establish and improve industry standards, enhance product quality and safety, reduce safety hazards, and boost consumer confidence.
\textbf{International cooperation:} Through policy support, promote international exchanges and cooperation in energy storage technology, enhancing the global competitiveness of China's energy storage industry.

\textbf{2. Intensified Industry Competition as a Risk Factor}

Intensified industry competition may bring the following challenges:

\textbf{Market share battles:} As more firms and capital enter the energy storage sector, market competition will become fiercer, potentially leading to price wars and compressing corporate profit margins.
\textbf{Pressure for technological innovation:} To stand out in competition, companies need to continuously invest in R\&D, accelerate technology iteration, and bear increasing cost burdens.
\textbf{Supply chain instability:} Intense competition may complicate supply chain management, causing shortages or price hikes of raw materials, thereby affecting production efficiency and product quality.
\textbf{Policy environment uncertainty:} Policy changes (e.g., adjustments in subsidy policies, stricter environmental regulations) can significantly impact the industry, increasing operational risks for enterprises.

\textbf{3. Inherent Conflict}

The inherent conflict between positive expectations and potential constraints mainly manifests in the following ways:

\textbf{Uneven resource allocation:} The goal of high-quality development requires substantial resource inputs, including capital, talent, and technology. If resources are overly concentrated in a few firms, other small and medium-sized enterprises may struggle to survive, hindering the healthy development of the entire industry.
\textbf{Innovation vs. competition paradox:} Policy encourages technological innovation and high-quality development, but excessive competitive pressure may drive firms to pursue short-term gains at the expense of long-term R\&D, creating a vicious cycle.
\textbf{Market entry barriers:} Industry optimization under policy guidance may establish high market entry barriers, making it difficult for new entrants to break through, further entrenching incumbents' market positions and suppressing the growth of emerging forces.

\textbf{4. Risk Amplification Effects Under Market Conditions}

Under the following market conditions, the risk of ``intensified competition'' may significantly weaken or even overturn the optimistic outlook for ``landscape optimization'':

\textbf{Macroeconomic recession:} An economic downturn may shrink market demand, reduce corporate profitability, intensify price competition, weaken firms' R\&D investment capacity, and delay technological progress.
\textbf{Sudden policy environment shifts:} Weakened government support for the energy storage industry (e.g., subsidy cuts, stricter environmental standards) would directly dampen corporate investment incentives and disrupt the industry's ecological balance.
\textbf{Technology pathway divergence:} The diversified development of energy storage technologies, with intensified competition among different pathways (e.g., lithium-ion, sodium-ion, flow batteries), may lead to fragmented resources and prevent the formation of unified technical standards and market consensus.
\textbf{Supply chain disruptions:} Interruptions in the supply of key raw materials or sharp price fluctuations would seriously affect production continuity and cost control, forcing firms to lower product quality or exit the market, thereby undermining industry stability.

\textbf{Conclusion}

There exists a complex interaction between the high-quality development and landscape optimization of the energy storage industry under policy guidance, on one hand, and the risk of intensified industry competition, on the other. In ideal circumstances, policy can effectively guide rational resource allocation, promote technological innovation, and drive industrial upgrading. However, in real-world market environments, uncertainties in macroeconomics, policy frameworks, and technology pathways may amplify competitive risks and threaten the gains from industry optimization. Therefore, a comprehensive consideration of multiple factors and the formulation of flexible coping strategies are necessary to ensure the healthy and sustainable development of the industry.
}
\end{innerblock}
\end{minipage}

\vspace{-3pt}
\textbf{Total Score} \scorepill{{54.0}}

\textbf{Subdimension Scores}
\begin{center}
\tiny
\setlength{\tabcolsep}{7pt} 
\begin{tabularx}{0.98\textwidth}{p{1.5cm}p{0.3cm}Y}
\toprule
\textbf{Subdimension} & \textbf{Score} & \textbf{Reason} \\
\midrule
Critical Thinking & {3} & {The response identifies conflict points such as resource allocation, the innovation paradox, and entry barriers, and enumerates risk conditions including macroeconomic downturn, sudden policy changes, technological divergence, and supply chain disruptions, demonstrating the ability to recognize boundaries/risks. However, the depth of analysis is insufficient. Unlike the reference answer, it fails to precisely reveal the core tension of ``idealized concentration increase vs. real-world disordered competition," nor does it deeply examine the key conflict mechanism of ``profit erosion due to price wars, forcing firms to prioritize survival over R\&D."} \\
Comparative perspective & {2} & {The response implies a hedging analysis in the ``technological divergence" part, noting that diversified technology paths may lead to resource fragmentation and difficulty in standard unification, which bears resemblance to ``Condition 2" in the reference answer. Nevertheless, it lacks the explicit historical comparisons found in the reference answer (e.g., the evolution of the photovoltaic and power battery industries), and fails to systematically compare different market conditions (e.g., demand growth, technological iteration, non-market factors) as specific variables that hedge against the prospect of ``landscape optimization."} \\
Argument balance & {3} & {The response structurally balances positive expectations and risk constraints, and its conclusion notes the need to consider multiple factors, reflecting an intention to be objective and neutral. However, in its detailed discussion, the analysis of how ``intensified competition" might concretely overturn ``landscape optimization" (e.g., leading to ``de-concentration" or delayed consolidation) is relatively weak. Unlike the reference answer, it does not clearly emphasize the severity that intensified competition is a highly certain reality and policy cannot bypass the market cleansing phase. The overall balance leans more toward description than critical depth.} \\
\bottomrule
\end{tabularx}
\end{center}
}
\end{tcolorbox}
% \caption{An illustrative failure case from the DI track. The model answer appears fluent, but it receives a low score due to weaknesses in evidence grounding, causal depth, and structured analytical rigor.}

\subsection{Failure Mode Taxonomy}\label{sec:failure_taxonomy}
\subsubsection{Semantic Consistency}

To obtain a finer-grained understanding of the limitations of current large language models on complex financial texts, we conduct a systematic failure-mode analysis on the Semantic Consistency (SC) track. Under the fine-grained human-like evaluation framework of our LLM-as-a-Judge module, we focus on two subtasks: Error Reasoning, which evaluates whether the model can correctly explain the semantic inconsistency, and Corrected Text, which evaluates whether the model can produce a faithful and valid correction. Based on the 0--10 scores of these two subtasks, we categorize model outputs into five representative result patterns: Both Fail, Explain but Not Fix, Fix but Not Explain, Partial Success, and Complete Success. Here, Both Fail indicates that the model neither identifies the inconsistency correctly nor provides a valid correction, while Complete Success denotes high-quality reasoning and correction under both criteria.

Table~\ref{tab:sc_failure_taxonomy} reports the average distribution of failure modes across the three SC dimensions. The \textit{Fact} dimension shows a relatively balanced distribution, with Both Fail (33.2\%) and Complete Success (32.9\%) occurring at nearly the same frequency, alongside a non-trivial proportion of Partial Success (20.9\%). This suggests that models possess some basic capacity for factual relation checking, but still frequently diverge in precise localization and faithful correction. The \textit{Logic} dimension yields the highest Complete Success rate (39.0\%), yet also a relatively high Both Fail rate (31.6\%), indicating strong polarization: top-performing models can handle multi-hop logical constraints in long contexts, whereas weaker models often fail entirely when faced with conflicting conditions. In contrast, Terminology is the most challenging dimension, with the lowest Complete Success rate (16.1\%) and the highest Both Fail rate (48.0\%). This pattern suggests that although models often detect semantic variation near a target term, they struggle to preserve strict local definitions and terminological boundaries in financial discourse, and therefore tend to replace the original term with superficially plausible but structurally mismatched alternatives.

\begin{table}[h]
\centering
\small
\caption{Average distribution of failure modes across the three dimensions of the Semantic Consistency track (\%).}
\label{tab:sc_failure_taxonomy}
\resizebox{\textwidth}{!}{

\begin{tabular}{lccccc}
\toprule
\textbf{Dimension} & \textbf{Both Fail} & \textbf{Complete Success} & \textbf{Fix but Not Explain} & \textbf{Explain but Not Fix} & \textbf{Partial Success} \\
\midrule
Fact        & 33.2 & 32.9 & 9.4  & 3.6 & 20.9 \\
Logic       & 31.6 & 39.0 & 10.1 & 6.4 & 13.0 \\
Terminology & 48.0 & 16.1 & 13.5 & 3.6 & 18.8 \\
\bottomrule
\end{tabular}}
\end{table}

\begin{figure}[h]
    \centering
    \includegraphics[width=0.8\textwidth]{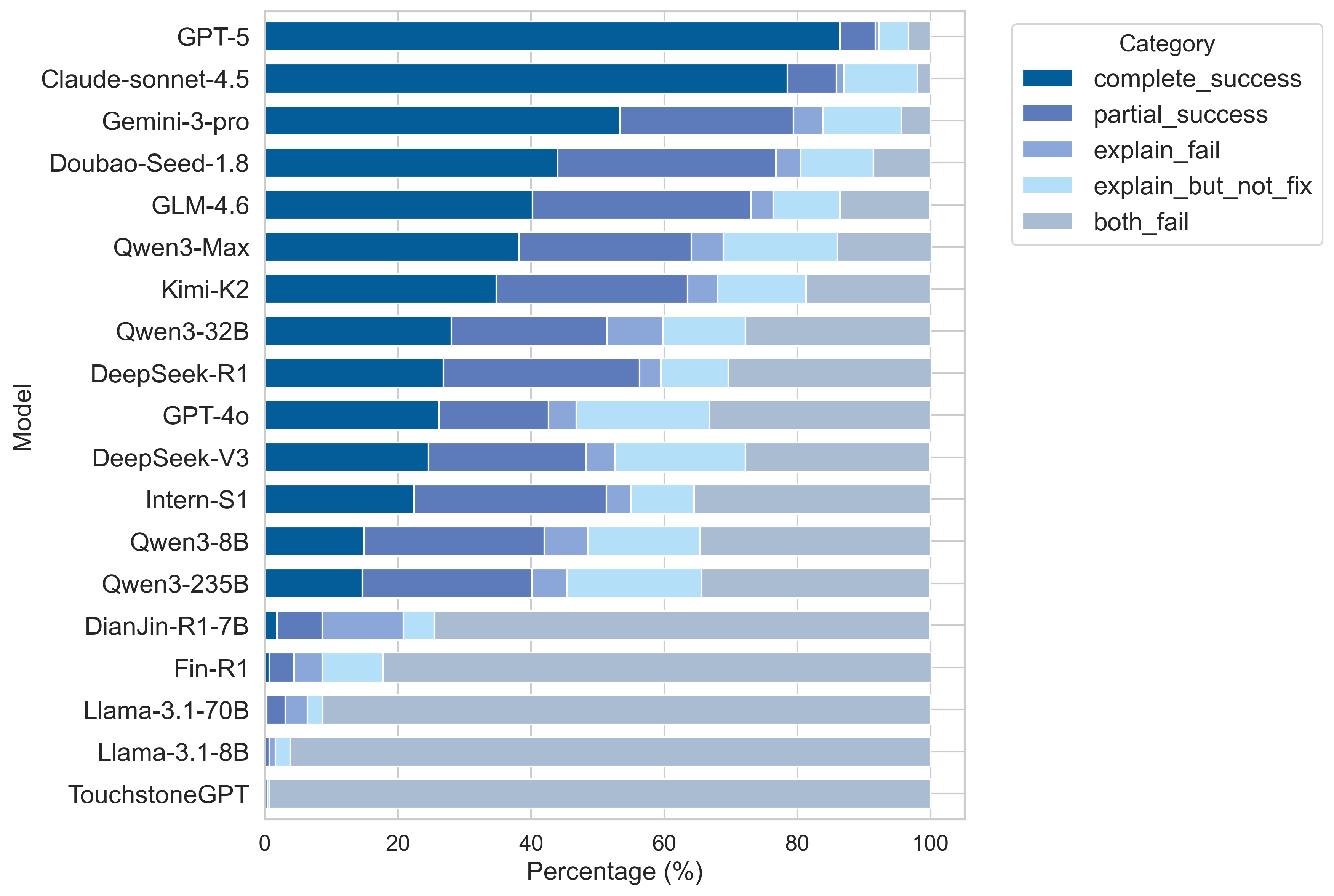}
    \caption{Failure Modes Distribution by Model.}
    \label{fig:failure_modes_model}
%\vspace{-10pt}
\end{figure}

As shown in Figure~\ref{fig:failure_modes_model}, Complete Success is achieved primarily by frontier closed-source models. In particular, GPT-5 reaches a complete-success rate of 91.7\%, substantially outperforming both comparable general-purpose models and most domain-specialized financial models. By contrast, smaller models and several finance-specific models exhibit much higher Both Fail rates, suggesting that the upper bound of performance on SC is still determined largely by the underlying base model's ability to control long-context constraints, preserve local semantic boundaries, and perform high-fidelity revision.

A closer dimension-wise analysis further highlights distinct challenges. For \textit{Terminology}, the combination of the lowest Complete Success rate and the highest Both Fail rate indicates that precise financial terminology requires strict adherence to local definitions, whereas models frequently substitute terms with seemingly reasonable but structurally inconsistent near-synonyms. For \textit{Logic}, the high complete-success rate coexists with severe polarization, suggesting that logical consistency is handled well only by models with sufficiently strong long-context reasoning capabilities. For \textit{Fact}, the distribution is more balanced, reflecting that factual relation checking is neither uniformly easy nor uniformly difficult, but instead highly sensitive to whether the model can maintain alignment between evidence, relation type, and correction target.

It is also worth noting that different models exhibit distinct failure profiles. For example, GPT4o shows a relatively high proportion of Explain but Not Fix (20.0\%). This indicates that the model can often recognize factual inconsistency at the explanation level, but still fails to reintegrate the relevant variables into a structurally coherent and grammatically valid corrected sentence. Such a pattern highlights that the difficulty of SC lies not only in detecting inconsistencies, but also in performing high-fidelity revision without violating the original discourse structure.

\paragraph{Representative case analysis.}
The Factual Relation Error example in Section~\ref{case:SC_Failure_Case} provides a representative illustration of SC failure. The main problem of the model is not merely an occasional misclassification, but a deeper task-level misalignment. The intended task is to detect whether factual relations in the long text have been altered with respect to the provided evidence. However, the model instead replaces this objective with a different decision rule: if a claim is not explicitly supported by the external knowledge base, it is treated as unverifiable or erroneous. This is a form of \emph{knowledge-absence-driven negative reasoning}, rather than genuine factual relation checking. Concretely, the model fails to capture the clearest state-relation error---rewriting ``still under review'' as ``already completed'' in the refinancing process---while incorrectly flagging statements such as ``the review process is progressing as expected,'' which are interpretive rather than hard relational distortions. It also produces ``cannot be confirmed'' as a so-called \texttt{correct\_content}, which is not a legitimate correction target but a meta-level refusal. In addition, its judgments on the SLR-related sentence and the AI-hardware demand sentence exhibit a strong tendency toward over-detection. This case therefore illustrates a core SC failure mode: the substitution of \emph{text-grounded relation checking} with \emph{external factual verification}.

\paragraph{Correlation Analysis: Objective Metrics vs. LLM-as-a-Judge.}
Finally, we examine the correlation between traditional automatic metrics and LLM-as-a-Judge scores in order to validate the latter at scale. On the SC track, Sentence-F1 (localization accuracy) shows a strong positive correlation with the judge's Explanation Correctness score, with Pearson's \(r \approx 0.84\). Meanwhile, semantic repair metrics such as SimCSE exhibit an even stronger correlation with the judge's Correct Content score, with Pearson's \(r \approx 0.91\). These results suggest that traditional objective metrics can capture robust alignment in semantic and positional dimensions, whereas the multi-dimensional scoring paradigm enabled by LLM-as-a-Judge provides a more scalable and interpretable framework for analyzing failure patterns in generated outputs.

\subsubsection{Data Alignment}

To systematically analyze the sources of model failures in this track, we abstract the error patterns beyond single-metric inaccuracies into four answer-alignment families. This classification considers both whether the final answer is correct, and whether the data IDs and fields referenced by the model align with the ground truth.
\paragraph{Distribution of weak dimensions.}
We categorize the failure modes into the following four types:
(1) \textbf{Correct answer with evidence mismatch}, where the model produces the correct answer but the referenced data ID or field deviates from the ground truth;
(2) \textbf{Wrong answer despite partial/full evidence match}, where the model at least partially locates the correct evidence yet still arrives at an incorrect final answer;
(3) \textbf{Wrong answer with evidence mismatch}, where both the answer and the evidence alignment fail;
(4) \textbf{Execution failure}, where the model fails during the execution pipeline and does not produce an evaluable result. Fully correct samples are tallied separately and are not considered part of the failure families.

\begin{figure}[h]
    \centering
    \includegraphics[width=0.8\textwidth]{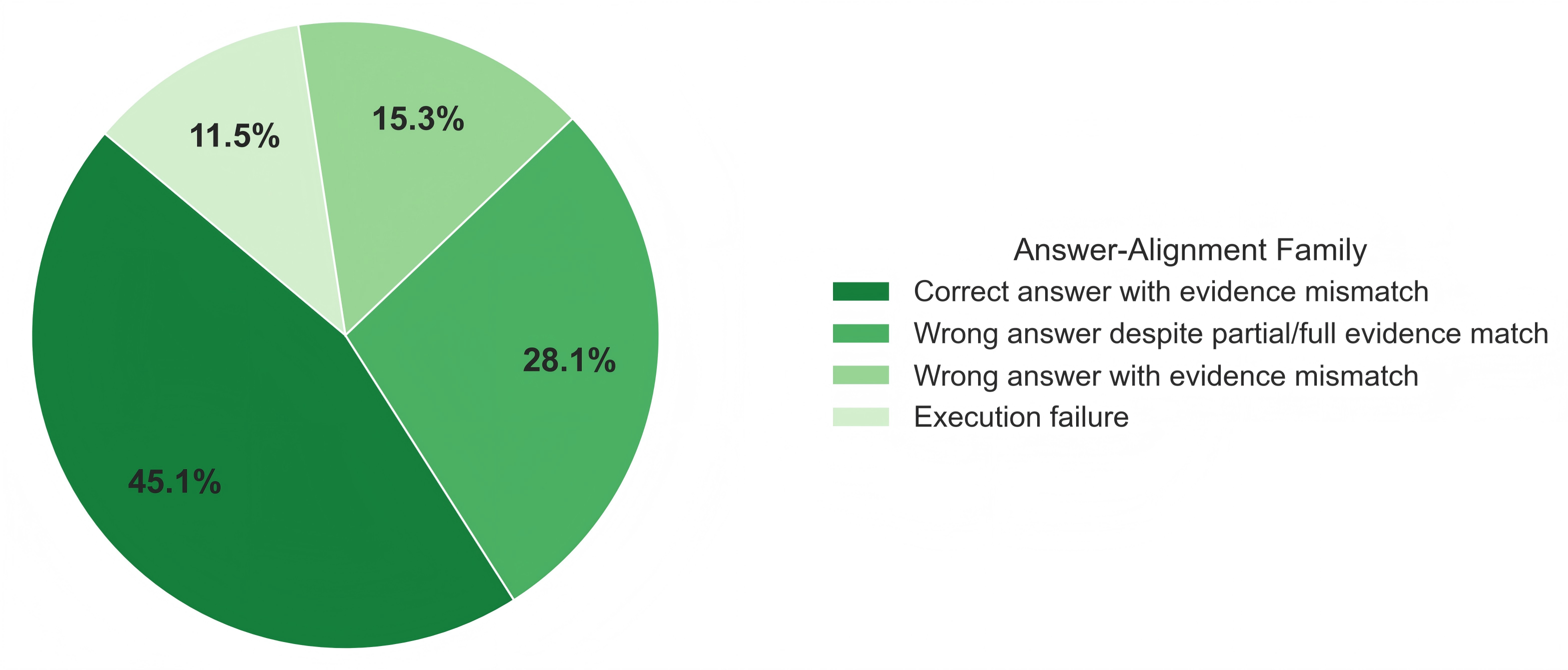}
    \caption{Distribution of failure mode families in the data alignment track.}
    \label{fig:composite_failure_modes_pie}
%\vspace{-10pt}
\end{figure}

The data alignment track comprises a total of 34,200 model-sample evaluation records. As shown in Figure~\ref{fig:composite_failure_modes_pie}, the most prevalent failure type is Correct answer with evidence mismatch, accounting for 11,027 instances, or 32.24\% of all samples. This indicates that a considerable proportion of models are capable of producing correct answers while referencing data IDs or fields that do not perfectly match the ground truth. For financial data question answering, such errors are particularly significant, as the model appears to answer correctly, yet the supporting evidence chain is not fully traceable, potentially compromising the auditability and trustworthiness of the results.

The second largest error category is Wrong answer despite partial/full evidence match, with 6,870 instances, representing 20.09\% of the samples. This type of error suggests that the model has partially or fully identified the correct evidence but still makes mistakes during computation, comparison, or rule-based reasoning. Consequently, the challenge of the data alignment track lies not only in retrieving the correct data, but also in performing stable numerical computation and multi-step reasoning based on the retrieved data. Wrong answer with evidence mismatch accounts for 10.91\% of the samples, representing cases where both the answer and the evidence localization fail. Execution failure constitutes 8.21\%, reflecting model instability in structured output generation, query construction, or execution flow.

\begin{figure}[h]
    \centering
    \includegraphics[width=0.8\textwidth]{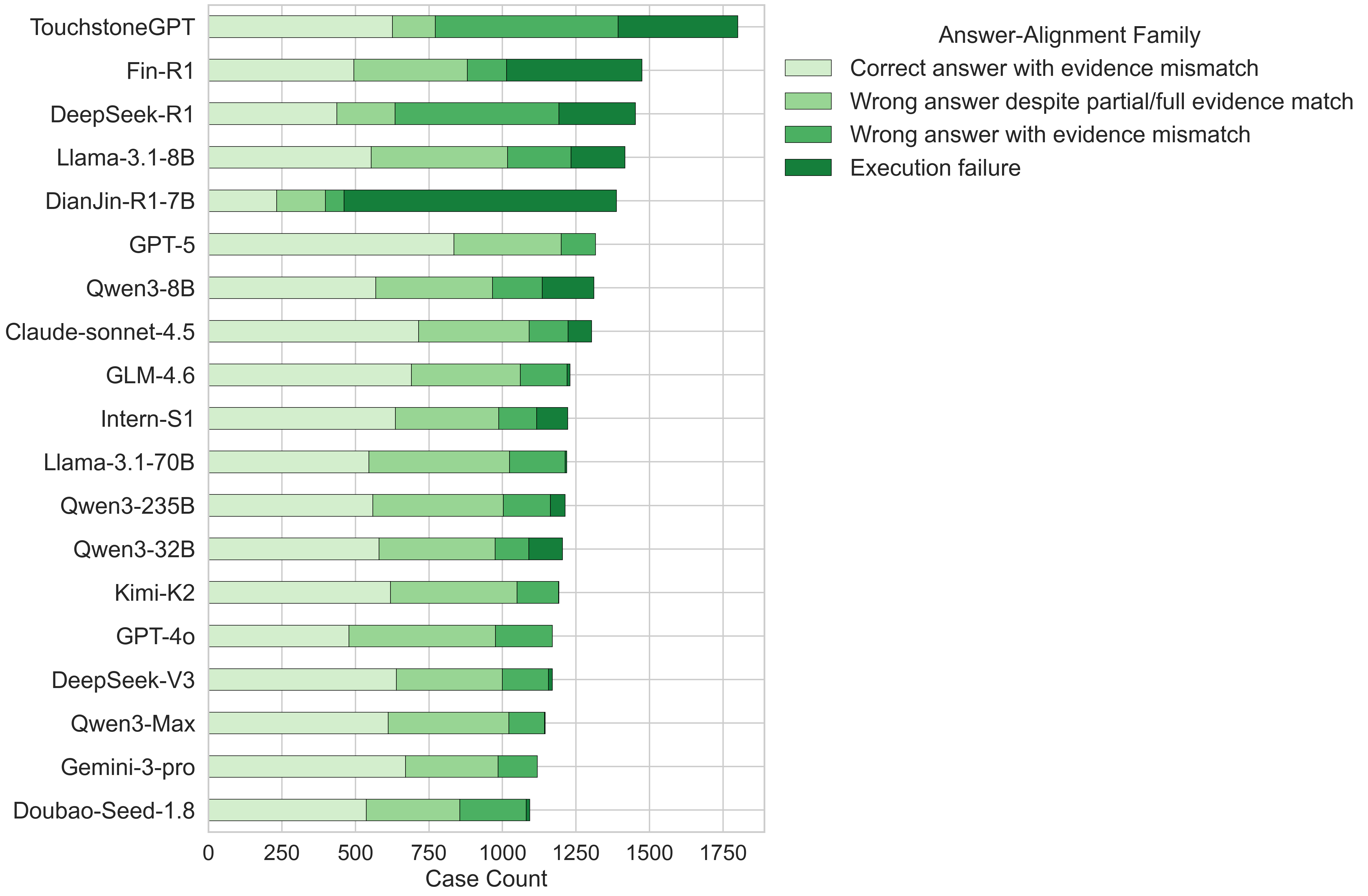}
    \caption{Model-wise counts of failure mode families, excluding fully correct cases.} \label{fig:model_answer_alignment_family_stacked}
%\vspace{-10pt}
\end{figure}

\paragraph{Performance differences across models and tasks.}
From the model perspective, Figure~\ref{fig:model_answer_alignment_family_stacked} presents the distribution of the four failure families across different models. This figure excludes fully correct samples, thereby allowing for a more direct comparison of the failure structures among models. It can be observed that different models exhibit distinct error sources: some models predominantly fall into the evidence mismatch category, where the answer may be correct but the data references are unstable; others have a higher proportion of execution failures or wrong answers with evidence mismatch, indicating notable deficiencies in procedural stability or foundational data retrieval.

\begin{figure}[h]
    \centering
    \includegraphics[width=0.8\textwidth]{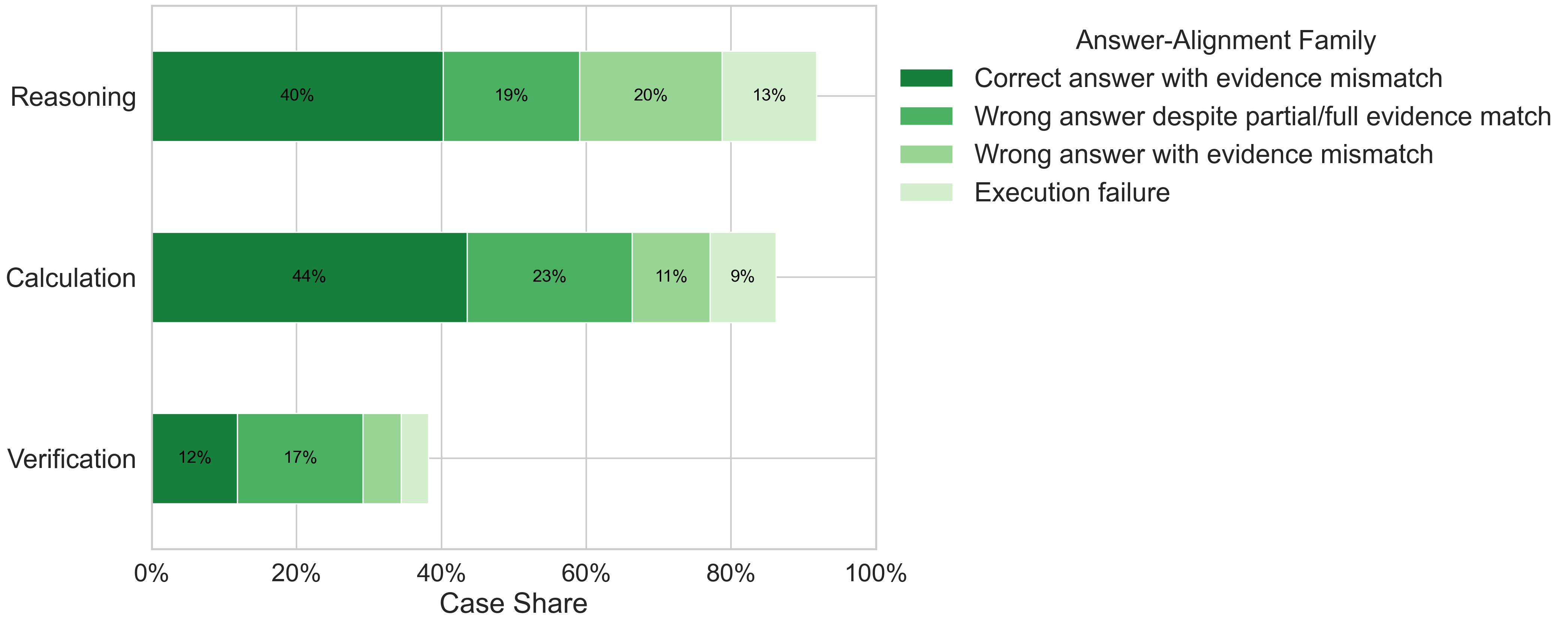}
    \caption{Failure mode distribution across verification, calculation, and reasoning tasks.} \label{fig:answer_alignment_task_distribution}
%\vspace{-10pt}
\end{figure}

From the task perspective, Figure~\ref{fig:answer_alignment_task_distribution} shows that Verification tasks achieve the highest proportion of fully correct samples at 61.76\%, with failures mainly stemming from minor evidence mismatches and answer errors. For Calculation tasks, Correct answer with evidence mismatch accounts for 43.56\% of the samples, suggesting that in complex numerical computation scenarios, models often arrive at correct results but struggle to precisely align the data sources they rely on. Reasoning tasks exhibit a higher risk of dual failures in both reasoning and evidence, with Wrong answer with evidence mismatch comprising 19.70\% and Execution failure 13.05\%, indicating that rule consistency and multi-step conditional judgments simultaneously amplify challenges in evidence selection, reasoning chains, and execution stability.

\paragraph{Representative case analysis.}
We further examine a representative failure case from the Data Alignment track(\ref{case:DA_Failure_Case}), where the model is asked to determine whether a bullish KDJ golden-cross signal is present under the \texttt{KDJ003} rule. At first glance, the response appears highly structured: it selects the relevant fields such as \texttt{daily\_high}, \texttt{daily\_low}, \texttt{daily\_close}, and \texttt{dt}, states the standard RSV, \(K\), \(D\), and \(J\) formulas, and provides a seemingly detailed calculation procedure. A closer inspection, however, shows that the error is not a local numerical mistake. The failure originates from a misalignment between the task semantics, the rule definition, and the actual data used for reasoning.

The gold reasoning treats the task as a target-point rule verification problem. Under \texttt{KDJ003}, the model should verify whether the final decision point satisfies the complete signal definition: the current \(K\) value is below 30, the current \(K\) is greater than \(D\), and the previous-day \(K\) is less than \(D\). The model instead reformulates the problem as an interval-level event search, looking for any date within the queried window where \(K\) crosses above \(D\). It therefore reports a golden cross on 2023-06-29 and answers ``Yes'', while the gold answer is ``No''. This is a temporal semantic drift: a terminal condition check is converted into a historical event retrieval task.

The second layer of failure lies in the separation between retrieval planning and answer generation. In the query-planning stage, the model appears to recognize that KDJ computation requires a historical window, and it expands the date range accordingly. This behavior is reasonable in principle, since KDJ cannot be computed from a single trading day. Yet in the final answer, the explicitly cited \texttt{data\_used} records cover only a small fraction of the sequence required by the gold derivation. The model therefore exhibits retrieval--reasoning decoupling: it plans for a sequence-based calculation, but the final reasoning is not grounded in the full retrieved sequence.

This mismatch is made more problematic by the model's pseudo-precise calculation. The response lists the standard KDJ formulas and gives concrete numerical claims about a specific crossing point, creating the appearance of a rigorous technical analysis. Such precision is misleading. Without using the complete historical sequence needed to reconstruct the rolling RSV values and recursively derive \(K\) and \(D\), the asserted crossing path cannot be reliably verified. The model does not merely calculate incorrectly; it fabricates a detailed computation trace under insufficient data support.

The rule execution is also incomplete. The benchmark rule is not a generic KDJ golden-cross heuristic. It requires the crossing to occur near the oversold region, with \(K<30\), and it also requires a specific previous-day-to-current-day state transition. The model mainly focuses on the crossing relation \(K<D \rightarrow K>D\), while failing to bind the oversold threshold and the target-point constraint to the final decision. In effect, it invokes a loose financial concept rather than executing the exact rule specified by the benchmark.

This case reveals a characteristic Data Alignment failure: the model is well aligned at the level of format, field selection, and formula presentation, but misaligned along the core semantic chain of \emph{question interpretation}--\emph{rule binding}--\emph{temporal scope}--\emph{data usage}--\emph{final judgment}. The response is therefore not a simple calculation error or a formatting failure, but a compound reasoning failure in which several structured components appear correct while the underlying decision process is wrong. For financial analysis, this pattern is particularly risky: the output looks auditable and professionally grounded, yet its conclusion is produced by reasoning over the wrong task.

Overall, this taxonomy reveals two core failure mechanisms in the data alignment track: first, ``correct answer but non-traceable evidence," and second, ``partially correct evidence but reasoning failure." The former underscores that relying solely on final answer accuracy overestimates model reliability in financial scenarios; the latter demonstrates that financial question answering systems require not only accurate data retrieval but also stable computational and rule-based reasoning capabilities. Therefore, LLM evaluation in the financial domain should simultaneously examine answer correctness, data ID alignment, and field alignment, rather than focusing exclusively on whether the final answer matches the ground truth.

\subsubsection{Deep Insight}

The DI track is designed to evaluate whether models can penetrate underlying business mechanisms and produce critically grounded analysis under complex evidence. Our experimental results show that this capability constitutes a key dividing line between frontier foundation models and domain-specialized fine-tuned models. Based on quantitative statistics over the four DI dimensions and their underlying error patterns, we construct a detailed taxonomy of failure modes. Concretely, we identify low-scoring cases from all model outputs using the criterion that either \texttt{total\_score} $<70$ or at least one \texttt{subdimension\_score} $<3$. This yields 7,490 low-scoring instances in total, accounting for 32.9\% of all evaluated DI samples.

\begin{figure}[h]
    \centering
    \includegraphics[width=0.8\textwidth]{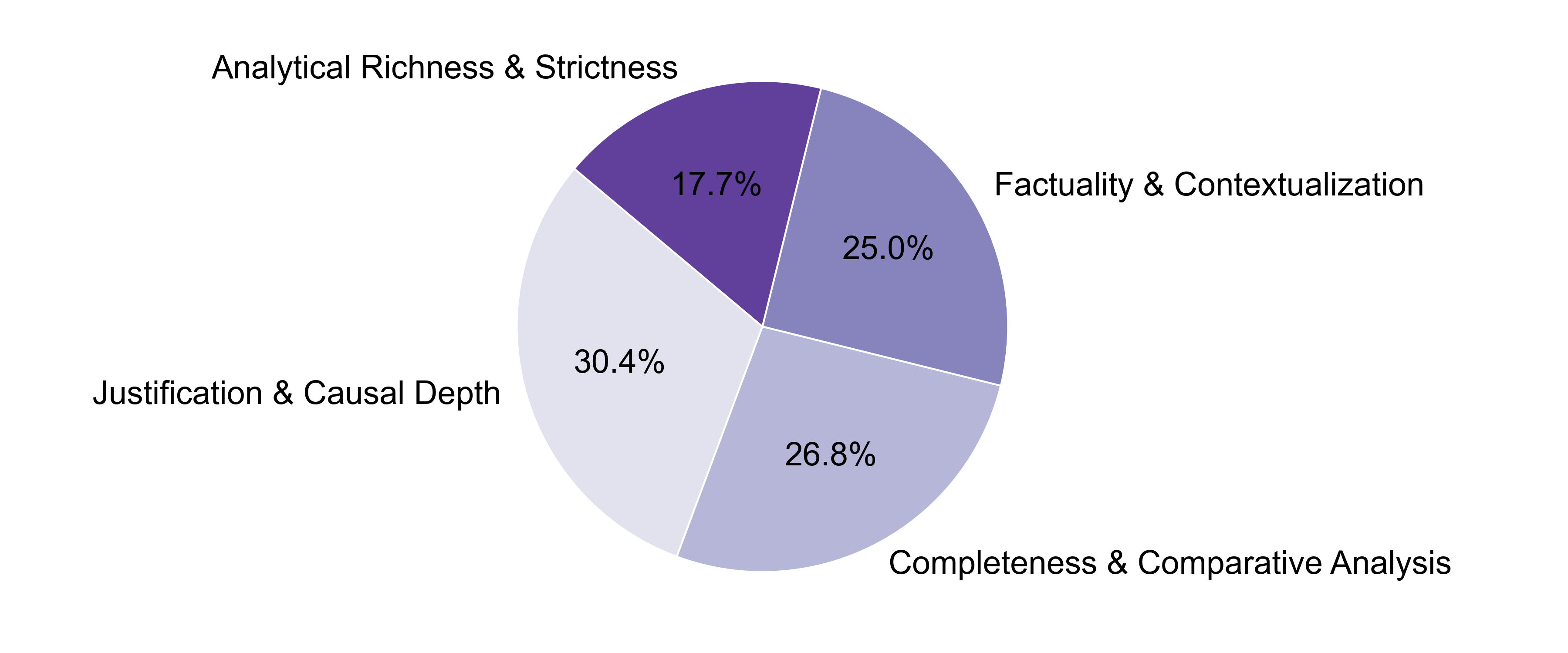}
    \caption{Distribution of weak dimensions.}
    \label{fig:dimension_fail_proportion}
%\vspace{-10pt}
\end{figure}

\paragraph{Distribution of weak dimensions. (Figure~\ref{fig:dimension_fail_proportion})}
Our statistics show that Justification Plausibility \& Causal Depth (30.4\%) and Completeness \& Comparative Analysis (26.8\%) are the two most failure-prone dimensions among low-scoring cases. By contrast, deductions caused by more surface-level features, such as Argument Richness \& Structural Rigor, account for only 17.7\%. This suggests that contemporary LLMs have largely mastered the formal discourse template of financial research reports, yet their reasoning remains highly formulaic: they often fail to couple general analytical rules with the specific business context under discussion.

\begin{figure}[h]
    \centering
    \includegraphics[width=\textwidth]{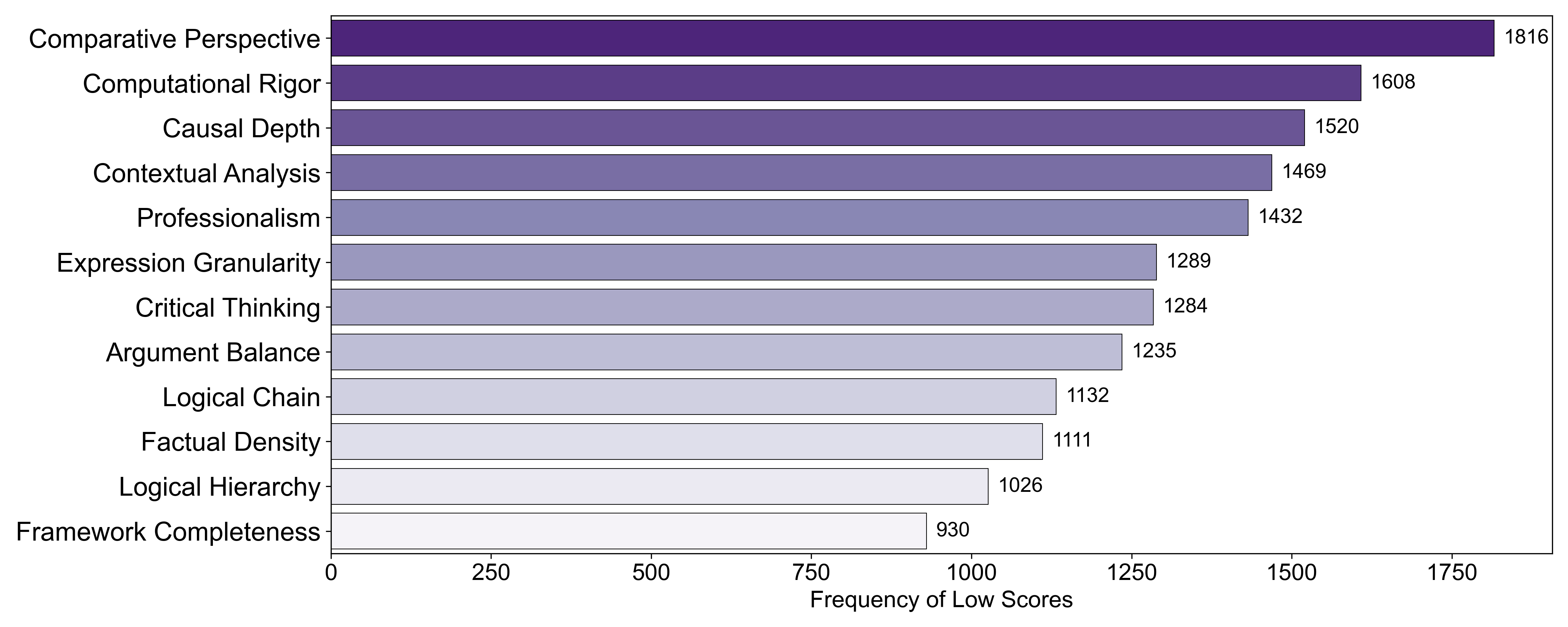}
    \caption{Diagnosis of high-frequency error patterns.}
    \label{fig:error_modes_frequency}
%\vspace{-10pt}
\end{figure}

\paragraph{Diagnosis of high-frequency error patterns. (Figure~\ref{fig:error_modes_frequency})}
Among cases with \texttt{subdimension\_score} $<3$, the two most frequent error types are Missing Comparative Perspective (1,816 occurrences) and Insufficient Computational Rigor (1,608 occurrences). When analyzing a specific industry, models exhibit a severe lack of horizontal mapping ability, either across peer competitors or across historical cycles, and their reasoning often collapses into linear and isolated argumentation. In addition, failures in Causal Depth (1,520 occurrences) and Contextual Analysis (1,469 occurrences) are also highly prominent. Most models remain at the level of superficial associations, failing to construct the finance-specific \emph{mechanism $\rightarrow$ impact} chain that underlies rigorous analytical reasoning. Their responses are often mechanically templated and insufficiently adapted to the particular market constraints described in the research report.

\begin{figure}[htbp]
    \centering
    \includegraphics[width=\textwidth]{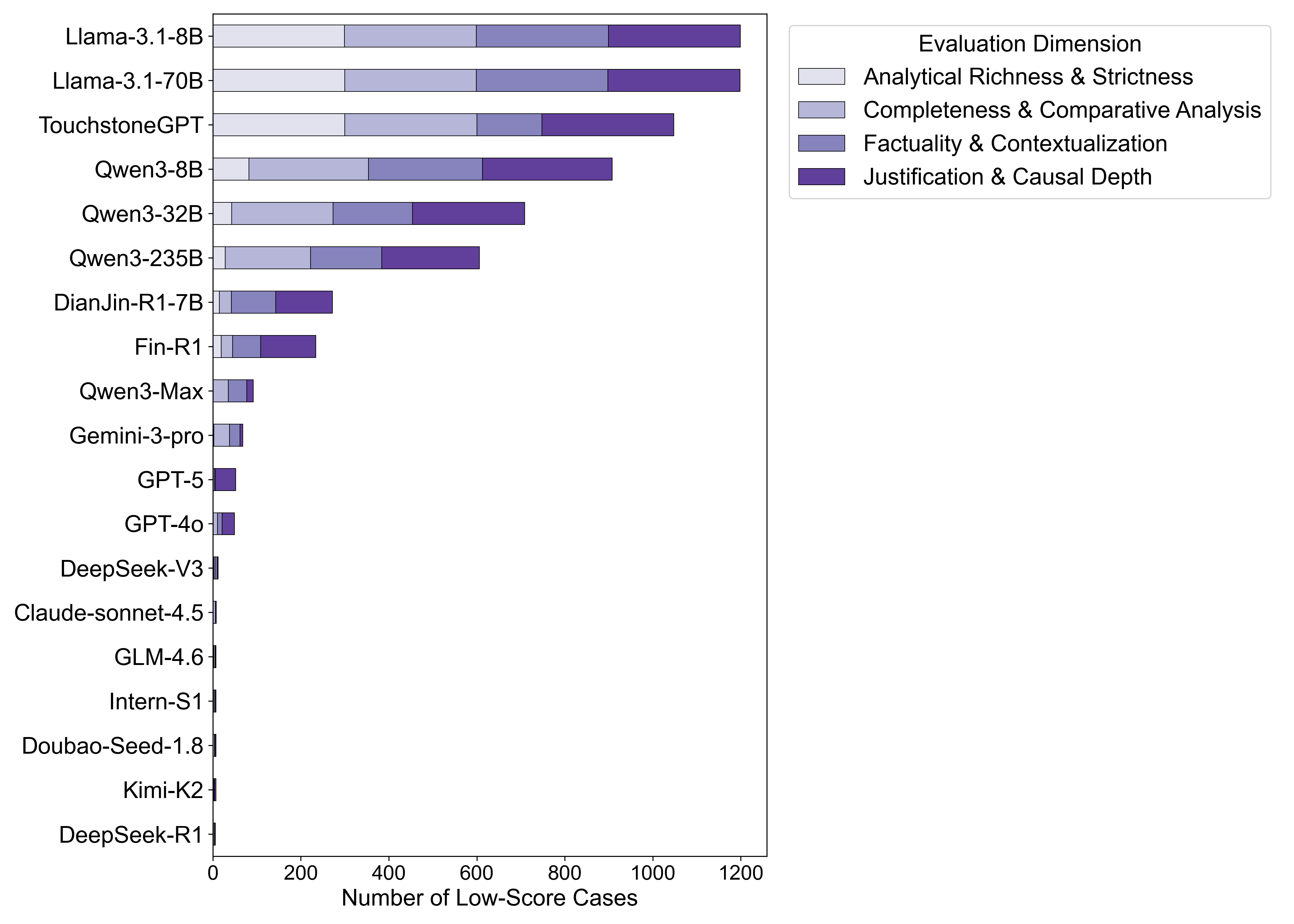}
    \caption{Performance differences across model tiers.}
    \label{fig:model_dimension_stacked}
\vspace{-10pt}
\end{figure}

\paragraph{Performance differences across model tiers. (Figure~\ref{fig:model_dimension_stacked})}
The experimental results reveal a marked pattern of polarization. Frontier foundation models represented by GPT-5, Doubao-Seed-1.8, and Kimi-K2 demonstrate decisively superior logical robustness on the DI task. In sharp contrast, many open-source financial models built on bases such as Llama-3.1-8B or Qwen2.5 (e.g., TouchstoneGPT) perform poorly across the DI track and are nearly uniformly below the passing threshold. The stacked distribution further shows that as overall model capability improves, failures related to Causal Depth and Comparative Analysis shrink the slowest. This finding strongly suggests that, for highly complex financial reasoning tasks, domain-specific supervised fine-tuning (SFT) on financial corpora can at best enhance the professional tone of the generated language, but cannot compensate for deficiencies in the underlying logical inference chain of the base model.

However, it is noteworthy that Fin-R1 and DianJin-R1-7B, despite being far smaller than Qwen3-235B, display significant competitive advantages. Both models adopt reasoning-enhancement strategies similar to those of DeepSeek-R1, aiming to improve logical depth through Chain-of-Thought. Fin-R1 employs reinforcement learning with the GRPO (Group Relative Policy Optimization) algorithm, while DianJin-R1-7B introduces Fin-PRM, a domain-specific process reward mechanism that provides finer-grained feedback on the correctness of each intermediate reasoning step. The success of these two models suggests that, in professional financial analysis, high-quality domain logic distillation and reward mechanisms that explicitly strengthen the reasoning process are more decisive than merely increasing parameter scale.

\paragraph{Representative case analysis.}
To illustrate the above statistical findings more concretely, we select a representative failure case of Fin-R1 in the Completeness \& Comparative Analysis category(\ref{case:DI_Failure_Case}). Although the model can generate structurally well-organized and terminology-dense text, it exhibits clear limitations in Critical Thinking. In particular, it merely enumerates broad notions such as ``uneven resource allocation'' without penetrating the underlying commercial mechanism to identify the core conflict, namely, the crowding-out effect of survival-driven price cuts on long-term R\&D investment.

Moreover, the model also shows substantial weakness in both precise causal derivation (Causal Depth) and cross-cycle comparative reasoning (Comparative Perspective). When explaining the trigger conditions under which competitive risk could overturn the expected market structure, the model tends to resort to generic macro narratives such as ``macroeconomic fluctuations,'' rather than adapting its reasoning to the specific market constraints described in the report. This lack of what may be termed \emph{historical mapping} prevents the model from introducing analogous evidence---for example, capacity-clearing episodes in the photovoltaic or lithium-battery industries---as a basis for comparative argumentation, thereby revealing a broader weakness in long-context relational reasoning. This failure mode directly justifies the inclusion of the DI track in FinReasoning: neither larger model scale nor the mere injection of domain knowledge automatically yields deep analytical capability. Genuine research-grade financial reasoning remains one of the most critical capability gaps between current frontier models and the demands of specialized financial tasks.

\subsection{Full Experiment Results}
\label{sec:detailed_model_results}
We further report the detailed evaluation results under each benchmark track and category in Table~\ref{tab:sc_terminology_results},\ref{tab:sc_fact_results},\ref{tab:sc_logic_results},\ref{tab:da_verification_results},\ref{tab:da_calculation_results},\ref{tab:da_reasoning_results},\ref{tab:di_all_category_results} to provide a more fine-grained view of model performance. Instead of relying only on aggregate scores, these tables decompose model behavior into rubric-level indicators, allows us to examine not only which models achieve higher overall scores, but also where their strengths and weaknesses arise across different financial reasoning requirements. 
% The results therefore provide a more diagnostic comparison of diverse model architectures.

\begin{table}[htbp]
\centering
\scriptsize
\setlength{\tabcolsep}{3pt}
\caption{Semantic Consistency evaluation results of the \textbf{Terminology} category. This table reports model performance on each evaluation rubric and the overall score (Best in bold, second best underlined).}
\label{tab:sc_terminology_results}
\resizebox{\textwidth}{!}{
\begin{tabular}{lcccccccc}
\toprule
\multirow{2}{*}{Models} 
& \multirow{2}{*}{\shortstack{Error Location\\Accuracy}} 
& \multicolumn{3}{c}{Error Explanation} 
& \multicolumn{3}{c}{Correct Content} 
& \multirow{2}{*}{\shortstack{Overall\\Score}} \\
\cmidrule(lr){3-5} \cmidrule(lr){6-8}
& & BERTScore & SimCSE & LLM-as-a-Judge 
& BERTScore & SimCSE & LLM-as-a-Judge & \\
\midrule

\rowcolor{TableYellow}\multicolumn{9}{l}{\textbf{Close-Source}} \\
\midrule
GPT-5 & 41.82 & 67.83 & 73.96 & \textbf{85.09} & \underline{69.81} & \underline{68.68} & \textbf{88.12} & \textbf{64.33} \\
GPT-4o & 11.68 & 64.65 & 63.82 & 48.24 & 30.39 & 28.94 & 50.34 & 35.71 \\
Gemini-3-pro & \underline{43.39} & 69.17 & \textbf{76.09} & 79.00 & 66.42 & 65.10 & 82.17 & 63.13 \\
Claude-sonnet-4.5 & 26.98 & \textbf{69.33} & 73.51 & \underline{80.65} & 53.78 & 52.10 & \underline{83.82} & 54.90 \\
GLM-4.6 & 41.34 & 69.20 & \underline{74.65} & 67.91 & 63.78 & 62.19 & 69.45 & 59.02 \\
Qwen3-Max & 27.20 & 68.41 & 73.88 & 62.20 & 55.24 & 54.32 & 64.78 & 51.16 \\
Kimi-K2 & 26.36 & 65.63 & 68.47 & 63.16 & 52.47 & 51.04 & 65.46 & 49.48 \\
Doubao-Seed-1.8 & \textbf{47.92} & \underline{69.31} & 74.18 & 71.53 & \textbf{71.53} & \textbf{70.77} & 73.69 & \underline{63.86} \\

\midrule
\rowcolor{TableOrange}\multicolumn{9}{l}{\textbf{Open-Source General}} \\
\midrule
DeepSeek-R1 & 32.80 & 69.10 & 72.62 & 58.22 & 53.57 & 52.63 & 59.48 & 51.56 \\
DeepSeek-V3 & 21.12 & 65.95 & 63.82 & 44.87 & 45.40 & 44.12 & 47.97 & 41.72 \\
Intern-S1 & 18.26 & 67.76 & 69.73 & 50.81 & 30.88 & 30.22 & 50.87 & 39.45 \\
Qwen3-8B & 8.37 & 67.23 & 69.32 & 48.81 & 15.94 & 15.18 & 49.66 & 32.36 \\
Qwen3-32B & 15.45 & 67.81 & 70.43 & 51.53 & 28.41 & 27.24 & 50.66 & 38.05 \\
Qwen3-235B & 17.41 & 67.41 & 69.99 & 51.28 & 34.98 & 34.05 & 52.87 & 40.31 \\
Llama-3.1-8B & 3.68 & 63.13 & 59.28 & 13.49 & 8.93 & 7.42 & 11.66 & 19.44 \\
Llama-3.1-70B & 3.45 & 63.22 & 59.57 & 16.39 & 8.85 & 8.02 & 10.97 & 19.71 \\

\midrule
\rowcolor{TableRed}\multicolumn{9}{l}{\textbf{Open-Source Financial}} \\
\midrule
Fin-R1 & 1.84 & 63.78 & 62.11 & 26.73 & 4.57 & 4.13 & 26.04 & 21.43 \\
TouchstoneGPT & 0.73 & 50.81 & 48.45 & 2.88 & 0.73 & 0.50 & 2.36 & 11.99 \\
DianJin-R1-7B & 0.85 & 63.61 & 58.37 & 25.59 & 2.21 & 2.23 & 17.34 & 19.10 \\
\bottomrule
\end{tabular}
}
\end{table}

\begin{table}[htbp]
\centering
\scriptsize
\setlength{\tabcolsep}{3pt}
\caption{Semantic Consistency evaluation results of \textbf{Fact} category. This table reports model performance on each evaluation rubric and the overall score(Best in bold, second best underlined).}
\label{tab:sc_fact_results}
\resizebox{\textwidth}{!}{
\begin{tabular}{lcccccccc}
\toprule
\multirow{2}{*}{Models} 
& \multirow{2}{*}{\shortstack{Error Location\\Accuracy}} 
& \multicolumn{3}{c}{Error Explanation} 
& \multicolumn{3}{c}{Correct Content} 
& \multirow{2}{*}{\shortstack{Overall\\Score}} \\
\cmidrule(lr){3-5} \cmidrule(lr){6-8}
& & BERTScore & SimCSE & LLM-as-a-Judge 
& BERTScore & SimCSE & LLM-as-a-Judge & \\
\midrule

\rowcolor{TableYellow}\multicolumn{9}{l}{\textbf{Close-Source}} \\
\midrule
GPT-5 & 63.47 & 68.33 & 76.77 & 80.14 & 72.43 & 72.41 & 82.38 & 71.43 \\
GPT-4o & 36.41 & 62.69 & 63.68 & 60.08 & 63.88 & 62.01 & 62.74 & 53.81 \\
Gemini-3-pro & 62.59 & 69.41 & 77.84 & \underline{81.43} & 71.19 & 69.08 & \underline{83.35} & 71.12 \\
Claude-sonnet-4.5 & 53.88 & \textbf{70.27} & \underline{78.16} & \textbf{82.12} & 72.17 & 71.19 & \textbf{85.22} & 68.97 \\
GLM-4.6 & \underline{67.04} & \underline{69.88} & 77.39 & 78.74 & \underline{76.97} & \underline{75.43} & 80.78 & \underline{73.37} \\
Qwen3-Max & 57.61 & 69.12 & \textbf{78.97} & 76.00 & 76.65 & 75.15 & 81.62 & 70.04 \\
Kimi-K2 & 55.87 & 65.88 & 71.14 & 72.43 & 71.80 & 70.02 & 74.68 & 65.95 \\
Doubao-Seed-1.8 & \textbf{69.33} & 68.18 & 76.67 & 80.19 & \textbf{77.66} & \textbf{76.87} & 82.36 & \textbf{74.44} \\

\midrule
\rowcolor{TableOrange}\multicolumn{9}{l}{\textbf{Open-Source General}} \\
\midrule
DeepSeek-R1 & 59.70 & 68.97 & 76.27 & 70.46 & 68.41 & 66.85 & 71.61 & 66.85 \\
DeepSeek-V3 & 55.86 & 63.92 & 66.62 & 67.11 & 67.09 & 64.34 & 71.19 & 63.09 \\
Intern-S1 & 44.65 & 67.29 & 73.24 & 64.58 & 55.39 & 53.80 & 65.88 & 57.13 \\
Qwen3-8B & 25.78 & 65.72 & 70.73 & 60.70 & 39.74 & 38.82 & 61.55 & 46.07 \\
Qwen3-32B & 37.48 & 66.45 & 73.94 & 68.10 & 53.33 & 52.03 & 68.09 & 54.93 \\
Qwen3-235B & 39.87 & 66.56 & 72.11 & 61.48 & 67.02 & 65.10 & 67.17 & 57.67 \\
Llama-3.1-8B & 6.60 & 59.22 & 56.28 & 11.94 & 16.22 & 14.98 & 11.30 & 21.08 \\
Llama-3.1-70B & 14.51 & 60.10 & 58.01 & 22.70 & 30.41 & 29.14 & 15.02 & 28.77 \\

\midrule
\rowcolor{TableRed}\multicolumn{9}{l}{\textbf{Open-Source Financial}} \\
\midrule
Fin-R1 & 4.12 & 61.76 & 62.89 & 23.44 & 10.67 & 10.01 & 22.31 & 22.60 \\
TouchstoneGPT & 1.42 & 47.20 & 47.57 & 2.71 & 2.09 & 1.90 & 2.05 & 11.98 \\
DianJin-R1-7B & 4.39 & 60.25 & 58.36 & 31.69 & 9.26 & 8.90 & 22.39 & 22.67 \\
\bottomrule
\end{tabular}
}
\end{table}

\begin{table}[htbp]
\centering
\scriptsize
\setlength{\tabcolsep}{3pt}
\caption{Semantic Consistency evaluation results of the \textbf{Logic} category. This table reports model performance on each evaluation rubric and the overall score (Best in bold, second best underlined).}
\label{tab:sc_logic_results}
\resizebox{\textwidth}{!}{
\begin{tabular}{lcccccccc}
\toprule
\multirow{2}{*}{Models} 
& \multirow{2}{*}{\shortstack{Error Location\\Accuracy}} 
& \multicolumn{3}{c}{Error Explanation} 
& \multicolumn{3}{c}{Correct Content} 
& \multirow{2}{*}{\shortstack{Overall\\Score}} \\
\cmidrule(lr){3-5} \cmidrule(lr){6-8}
& & BERTScore & SimCSE & LLM-as-a-Judge 
& BERTScore & SimCSE & LLM-as-a-Judge & \\
\midrule

\rowcolor{TableYellow}\multicolumn{9}{l}{\textbf{Close-Source}} \\
\midrule
GPT-5 & 37.21 & 64.45 & \underline{72.32} & \underline{87.89} & 45.38 & 45.80 & \underline{90.47} & 57.55 \\
GPT-4o & 14.44 & 60.72 & 55.54 & 67.15 & 21.48 & 21.30 & 68.71 & 37.58 \\
Gemini-3-pro & 28.35 & 64.87 & 71.59 & 87.15 & 28.27 & 26.51 & 88.32 & 50.19 \\
Claude-sonnet-4.5 & 24.37 & \textbf{66.47} & \textbf{72.52} & \textbf{89.76} & 36.55 & 36.37 & \textbf{91.82} & 51.84 \\
GLM-4.6 & \underline{49.83} & 65.29 & 69.70 & 83.23 & \textbf{55.33} & \underline{54.30} & 84.88 & \underline{62.47} \\
Qwen3-Max & 36.80 & 64.57 & 70.72 & 84.29 & 48.97 & 48.46 & 86.74 & 57.13 \\
Kimi-K2 & 45.70 & 61.81 & 64.78 & 82.63 & 51.58 & 50.54 & 83.26 & 59.08 \\
Doubao-Seed-1.8 & \textbf{50.49} & \underline{65.35} & 71.78 & 84.33 & \underline{55.27} & \textbf{55.14} & 86.56 & \textbf{63.32} \\

\midrule
\rowcolor{TableOrange}\multicolumn{9}{l}{\textbf{Open-Source General}} \\
\midrule
DeepSeek-R1 & 29.20 & 65.25 & 68.59 & 67.54 & 32.21 & 31.02 & 68.25 & 46.72 \\
DeepSeek-V3 & 42.25 & 63.11 & 60.43 & 69.79 & 48.24 & 45.93 & 72.05 & 54.03 \\
Intern-S1 & 10.72 & 62.36 & 64.20 & 66.39 & 13.01 & 12.53 & 68.17 & 35.43 \\
Qwen3-8B & 12.99 & 61.98 & 57.01 & 67.46 & 22.00 & 21.09 & 69.97 & 37.61 \\
Qwen3-32B & 6.77 & 62.52 & 65.23 & 70.33 & 10.50 & 10.29 & 68.34 & 34.17 \\
Qwen3-235B & 5.86 & 62.77 & 65.88 & 67.10 & 8.57 & 8.28 & 66.63 & 32.98 \\
Llama-3.1-8B & 2.51 & 56.64 & 50.04 & 17.14 & 5.39 & 5.20 & 10.48 & 16.94 \\
Llama-3.1-70B & 4.37 & 59.59 & 52.87 & 25.10 & 8.41 & 8.22 & 16.65 & 20.44 \\

\midrule
\rowcolor{TableRed}\multicolumn{9}{l}{\textbf{Open-Source Financial}} \\
\midrule
Fin-R1 & 3.29 & 62.18 & 62.80 & 36.39 & 8.06 & 7.82 & 30.97 & 24.23 \\
TouchstoneGPT & 2.88 & 46.93 & 41.91 & 1.93 & 1.91 & 1.45 & 1.95 & 11.64 \\
DianJin-R1-7B & 0.47 & 59.28 & 54.02 & 37.66 & 0.90 & 0.87 & 28.15 & 20.26 \\
\bottomrule
\end{tabular}
}
\end{table}

\begin{table}[htbp]
\centering
\scriptsize
\setlength{\tabcolsep}{10pt}
\caption{Data Alignment evaluation results of the \textbf{Verification} category. This table reports model performance on each evaluation rubric and the overall score (Best in bold, second best underlined).}
\label{tab:da_verification_results}
\resizebox{\textwidth}{!}{
\begin{tabular}{lccccc}
\toprule
Models 
& Success Rate 
& Answer Accuracy 
& Data IDs F1 
& Fields F1 
& Overall Score \\
\midrule

\rowcolor{TableYellow}\multicolumn{6}{l}{\textbf{Close-Source}} \\
\midrule
GPT-5 & \textbf{100.00} & 81.50 & 99.86 & 83.50 & 88.29 \\
GPT-4o & \textbf{100.00} & 81.83 & \textbf{100.00} & 89.35 & 90.39 \\
Gemini-3-pro & \textbf{100.00} & \textbf{87.17} & \textbf{100.00} & \textbf{95.82} & \textbf{94.33} \\
Claude-sonnet-4.5 & 92.83 & \underline{86.54} & 99.84 & 89.27 & 85.30 \\
GLM-4.6 & \textbf{100.00} & 79.17 & \textbf{100.00} & 88.99 & 89.39 \\
Qwen3-Max & \textbf{100.00} & 81.33 & \textbf{100.00} & \underline{95.66} & 92.33 \\
Kimi-K2 & \textbf{100.00} & 82.67 & 99.74 & 88.20 & 90.20 \\
Doubao-Seed-1.8 & \textbf{100.00} & 85.83 & \underline{99.96} & 94.70 & \underline{93.50} \\

\midrule
\rowcolor{TableOrange}\multicolumn{6}{l}{\textbf{Open-Source General}} \\
\midrule
DeepSeek-R1 & \textbf{100.00} & 23.50 & 28.12 & 25.62 & 25.75 \\
DeepSeek-V3 & \textbf{100.00} & 85.33 & 99.83 & 90.32 & 91.83 \\
Intern-S1 & \textbf{100.00} & 79.67 & \textbf{100.00} & 86.69 & 88.79 \\
Qwen3-8B & 97.33 & 77.05 & 96.10 & 84.13 & 83.48 \\
Qwen3-32B & 99.50 & 84.25 & \textbf{100.00} & 91.99 & 91.62 \\
Qwen3-235B & 99.00 & 83.84 & \textbf{100.00} & 89.23 & 90.11 \\
Llama-3.1-8B & 97.33 & 67.47 & 94.52 & 79.55 & 78.37 \\
Llama-3.1-70B & \underline{99.67} & 80.10 & 98.79 & 86.65 & 88.22 \\

\midrule
\rowcolor{TableRed}\multicolumn{6}{l}{\textbf{Open-Source Financial}} \\
\midrule
Fin-R1 & 84.00 & 70.44 & \textbf{100.00} & 63.26 & 65.43 \\
TouchstoneGPT & 75.00 & 48.00 & 0.00 & 47.13 & 23.78 \\
DianJin-R1-7B & 74.67 & 82.37 & 99.33 & 82.92 & 65.86 \\
\bottomrule
\end{tabular}
}
\end{table}

\begin{table}[htbp]
\centering
\scriptsize
\setlength{\tabcolsep}{10pt}
\caption{Data Alignment evaluation results of the \textbf{Calculation} category. This table reports model performance on each evaluation rubric and the overall score (Best in bold, second best underlined).}
\label{tab:da_calculation_results}
\resizebox{\textwidth}{!}{
\begin{tabular}{lccccc}
\toprule
Models 
& Success Rate 
& Answer Accuracy 
& Data IDs F1 
& Fields F1 
& Overall Score \\
\midrule

\rowcolor{TableYellow}\multicolumn{6}{l}{\textbf{Close-Source}} \\
\midrule
GPT-5 & \textbf{100.00} & \underline{72.25} & \textbf{81.32} & \underline{64.74} & \textbf{72.77} \\
GPT-4o & \textbf{100.00} & 48.63 & 76.23 & 58.17 & 61.01 \\
Gemini-3-pro & \textbf{100.00} & \textbf{73.13} & 76.14 & 60.42 & \underline{69.90} \\
Claude-sonnet-4.5 & 95.63 & 64.24 & 78.25 & 60.28 & 64.63 \\
GLM-4.6 & 98.75 & 71.15 & 74.92 & 58.95 & 67.49 \\
Qwen3-Max & \textbf{100.00} & 67.00 & \underline{79.88} & 54.46 & 67.11 \\
Kimi-K2 & \underline{99.88} & 63.69 & 74.65 & \textbf{69.80} & 69.29 \\
Doubao-Seed-1.8 & 98.88 & 67.05 & 73.62 & 61.35 & 66.58 \\

\midrule
\rowcolor{TableOrange}\multicolumn{6}{l}{\textbf{Open-Source General}} \\
\midrule
DeepSeek-R1 & 98.13 & 70.28 & 75.86 & 64.49 & 68.90 \\
DeepSeek-V3 & \textbf{100.00} & 63.50 & 72.04 & 55.56 & 63.70 \\
Intern-S1 & 91.00 & 69.71 & 73.95 & 63.57 & 62.86 \\
Qwen3-8B & 85.25 & 59.22 & 75.88 & 47.36 & 51.85 \\
Qwen3-32B & 86.63 & 61.02 & 76.85 & 54.90 & 55.66 \\
Qwen3-235B & 99.63 & 58.04 & 76.40 & 55.45 & 63.06 \\
Llama-3.1-8B & 87.50 & 53.77 & 67.52 & 47.20 & 49.14 \\
Llama-3.1-70B & \textbf{100.00} & 58.25 & 72.20 & 53.34 & 61.26 \\

\midrule
\rowcolor{TableRed}\multicolumn{6}{l}{\textbf{Open-Source Financial}} \\
\midrule
Fin-R1 & 74.63 & 52.49 & 72.33 & 38.41 & 40.60 \\
TouchstoneGPT & 72.38 & 49.89 & 0.00 & 24.48 & 17.94 \\
DianJin-R1-7B & 33.38 & 62.71 & 64.15 & 53.57 & 20.07 \\
\bottomrule
\end{tabular}
}
\end{table}

\begin{table}[htbp]
\centering
\scriptsize
\setlength{\tabcolsep}{10pt}
\caption{Data Alignment evaluation results of the \textbf{Reasoning} category. This table reports model performance on each evaluation rubric and the overall score (Best in bold, second best underlined).}
\label{tab:da_reasoning_results}
\resizebox{\textwidth}{!}{
\begin{tabular}{lccccc}
\toprule
Models 
& Success Rate 
& Answer Accuracy 
& Data IDs F1 
& Fields F1 
& Overall Score \\
\midrule

\rowcolor{TableYellow}\multicolumn{6}{l}{\textbf{Close-Source}} \\
\midrule
GPT-5 & \textbf{100.00} & \underline{63.00} & 64.67 & \textbf{77.80} & \textbf{68.49} \\
GPT-4o & \textbf{100.00} & 57.00 & 62.31 & 69.85 & 63.05 \\
Gemini-3-pro & \textbf{100.00} & 60.75 & 38.62 & 71.09 & 56.82 \\
Claude-sonnet-4.5 & \underline{99.50} & 60.80 & 52.18 & 75.99 & 62.68 \\
GLM-4.6 & \textbf{100.00} & 55.75 & 46.75 & 72.69 & 58.40 \\
Qwen3-Max & \underline{99.50} & 60.80 & 62.97 & \underline{76.14} & \underline{66.30} \\
Kimi-K2 & \textbf{100.00} & 55.50 & 58.32 & 75.85 & 63.22 \\
Doubao-Seed-1.8 & 99.25 & 49.75 & 53.11 & 65.27 & 55.63 \\

\midrule
\rowcolor{TableOrange}\multicolumn{6}{l}{\textbf{Open-Source General}} \\
\midrule
DeepSeek-R1 & 38.75 & 60.65 & 48.98 & 64.62 & 22.51 \\
DeepSeek-V3 & 96.50 & \textbf{64.51} & 56.04 & 67.13 & 60.37 \\
Intern-S1 & 91.00 & 58.52 & 48.21 & 71.07 & 53.93 \\
Qwen3-8B & 89.50 & 56.42 & 67.28 & 57.99 & 54.21 \\
Qwen3-32B & 99.00 & 62.12 & 67.85 & 67.15 & 65.05 \\
Qwen3-235B & 89.75 & 51.53 & 60.82 & 68.78 & 54.19 \\
Llama-3.1-8B & 83.25 & 47.45 & 67.43 & 38.76 & 42.63 \\
Llama-3.1-70B & 99.25 & 46.10 & \underline{71.66} & 67.60 & 61.32 \\

\midrule
\rowcolor{TableRed}\multicolumn{6}{l}{\textbf{Open-Source Financial}} \\
\midrule
Fin-R1 & 59.75 & 60.67 & \textbf{80.41} & 52.96 & 38.65 \\
TouchstoneGPT & 91.00 & 33.52 & 0.00 & 8.15 & 12.64 \\
DianJin-R1-7B & 16.00 & 51.56 & 43.97 & 34.05 & 6.91 \\
\bottomrule
\end{tabular}
}
\end{table}

\begin{table}[htbp]
\centering
\scriptsize
\setlength{\tabcolsep}{2pt}
\caption{Deep Insight evaluation results across four categories: \textbf{Justification \& Causal Depth}, \textbf{Factuality \& Context}, \textbf{Completeness \& Comparative Analysis}, and \textbf{Architecture Richness \& Strictness}. The abbreviated rubrics are defined as follows: RC = Reasoning Chain, CD = Causal Depth, Prof. = Professionalism, FD = Factual Density, CR = Computational Rigor, CA = Contextual Analysis, CT = Critical Thinking, CP = Comparative Perspective, AB = Argument Balance, FC = Framework Completeness, LH = Logical Hierarchy, and EG = Expression Granularity. This table reports model performance on each rubric and the overall category score (Best in bold, second best underlined).}

\label{tab:di_all_category_results}
\resizebox{\textwidth}{!}{
\begin{tabular}{lcccccccccccccccc}
\toprule
\multirow{2}{*}{Models}
& \multicolumn{4}{c}{\shortstack{Justification \&\\Causal Depth}}
& \multicolumn{4}{c}{\shortstack{Factuality \&\\Context}}
& \multicolumn{4}{c}{\shortstack{Completeness \&\\Comparative Analysis}}
& \multicolumn{4}{c}{\shortstack{Architecture Richness \&\\Strictness}} \\
\cmidrule(lr){2-5} \cmidrule(lr){6-9} \cmidrule(lr){10-13} \cmidrule(lr){14-17}
& Score & RC & CD & Prof.
& Score & FD & CR & CA
& Score & CT & CP & AB
& Score & FC & LH & EG \\
\midrule

\rowcolor{TableYellow}\multicolumn{17}{l}{\textbf{Close-Source}} \\
\midrule
GPT-5 & 81.80 & 32.29 & 24.18 & 24.78 & 84.58 & 32.80 & 25.36 & 26.42 & 86.71 & 35.39 & \underline{24.24} & 27.08 & 92.39 & 36.99 & 27.74 & 27.66 \\
GPT-4o & 76.93 & 31.95 & 21.52 & 23.46 & 77.64 & 31.84 & 22.08 & 23.72 & 74.78 & 31.52 & 19.62 & 23.64 & 79.96 & 32.00 & 23.98 & 23.98 \\
Gemini-3-pro & 80.67 & 33.73 & 21.62 & 25.32 & 79.01 & 32.21 & 22.16 & 24.64 & 76.65 & 32.43 & 20.38 & 23.84 & 82.14 & 32.88 & 24.66 & 24.60 \\
Claude-sonnet-4.5 & \underline{96.17} & 38.13 & \underline{28.54} & \textbf{29.50} & 87.07 & \underline{34.11} & 25.94 & 27.02 & 84.51 & 34.85 & 23.48 & 26.18 & 92.19 & 36.93 & 27.72 & 27.54 \\
GLM-4.6 & 96.06 & \underline{38.16} & \underline{28.54} & \underline{29.36} & \textbf{88.10} & \textbf{34.24} & 25.38 & \textbf{28.48} & \underline{87.33} & \underline{36.03} & 23.88 & \underline{27.42} & 92.58 & 37.04 & 27.78 & 27.76 \\
Qwen3-Max & 82.89 & 35.31 & 22.00 & 25.58 & 75.95 & 31.39 & 21.42 & 23.14 & 75.78 & 31.68 & 20.14 & 23.96 & 85.80 & 34.40 & 25.80 & 25.60 \\
Kimi-K2 & \textbf{96.27} & \textbf{38.19} & \textbf{28.58} & \textbf{29.50} & \underline{87.42} & 33.68 & \textbf{26.82} & 26.92 & \textbf{89.47} & \textbf{36.77} & \textbf{25.04} & \textbf{27.66} & \underline{93.38} & \underline{37.20} & \underline{28.04} & \textbf{28.14} \\
Doubao-Seed-1.8 & 95.73 & \textbf{38.19} & \textbf{28.58} & 28.96 & 87.15 & 33.87 & \underline{26.12} & \underline{27.16} & 86.17 & 35.41 & 23.88 & 26.88 & \textbf{93.52} & \textbf{37.36} & \textbf{28.10} & \underline{28.06} \\

\midrule
\rowcolor{TableOrange}\multicolumn{17}{l}{\textbf{Open-Source General}} \\
\midrule
DeepSeek-R1 & 89.76 & 35.76 & 26.72 & 27.28 & 84.63 & 33.09 & 25.52 & 26.02 & 80.03 & 33.15 & 21.86 & 25.02 & 88.03 & 35.23 & 26.42 & 26.38 \\
DeepSeek-V3 & 86.36 & 34.32 & 25.58 & 26.46 & 81.03 & 32.13 & 23.76 & 25.14 & 77.89 & 32.21 & 21.42 & 24.26 & 81.99 & 32.83 & 24.56 & 24.60 \\
Intern-S1 & 88.00 & 34.88 & 26.02 & 27.10 & 82.08 & 32.40 & 24.30 & 25.38 & 81.59 & 33.57 & 22.70 & 25.32 & 84.36 & 33.76 & 25.30 & 25.30 \\
Qwen3-8B & 57.37 & 27.39 & 14.90 & 15.08 & 51.09 & 23.49 & 13.22 & 14.38 & 53.85 & 23.25 & 12.70 & 17.90 & 69.33 & 29.65 & 22.28 & 17.40 \\
Qwen3-32B & 65.70 & 30.64 & 17.04 & 18.02 & 61.67 & 27.49 & 16.56 & 17.62 & 60.29 & 25.87 & 14.66 & 19.76 & 73.61 & 30.85 & 23.18 & 19.58 \\
Qwen3-235B & 67.48 & 31.12 & 17.38 & 18.98 & 63.79 & 28.19 & 17.22 & 18.38 & 63.65 & 26.99 & 15.88 & 20.78 & 75.38 & 31.28 & 23.48 & 20.62 \\
Llama-3.1-8B & 32.13 & 16.27 & 7.34 & 8.52 & 30.83 & 15.79 & 6.68 & 8.36 & 31.33 & 14.21 & 6.12 & 11.00 & 39.09 & 17.49 & 12.44 & 9.16 \\
Llama-3.1-70B & 36.84 & 17.68 & 8.96 & 10.20 & 36.47 & 18.21 & 8.58 & 9.68 & 34.73 & 15.65 & 7.06 & 12.02 & 42.64 & 18.56 & 13.34 & 10.74 \\

\midrule
\rowcolor{TableRed}\multicolumn{17}{l}{\textbf{Open-Source Financial}} \\
\midrule
Fin-R1 & 71.47 & 31.09 & 18.90 & 21.48 & 71.83 & 30.43 & 19.06 & 22.34 & 73.53 & 31.23 & 18.84 & 23.46 & 77.90 & 31.36 & 23.36 & 23.18 \\
TouchstoneGPT & 34.22 & 16.56 & 8.32 & 9.34 & 28.70 & 7.12 & 2.78 & 4.26 & 30.78 & 14.16 & 6.02 & 10.60 & 38.33 & 16.99 & 12.18 & 9.16 \\
DianJin-R1-7B & 69.14 & 29.92 & 18.42 & 20.80 & 63.61 & 27.01 & 16.66 & 19.94 & 73.27 & 31.01 & 18.96 & 23.30 & 77.14 & 31.04 & 23.16 & 22.94 \\
\bottomrule
\end{tabular}
}
\end{table}

\section{Detailed Instructions and Prompts}
In this section, we provide the complete system of prompts and instructions employed during the construction and evaluation of the FinReasoning benchmark to ensure procedural transparency and methodological rigor. In Section~\ref{sec:data_generation_prompts}, we detail the data generation prompts, covering corpus synthesis, fine-grained error injection, and rule-based QA construction. Section~\ref{sec:evaluation_prompts} presents the standardized evaluation prompts used to probe the reasoning capabilities of various LLMs. Finally, in Section~\ref{sec:judge_prompts}, we specify the instructions for LLM-as-a-Judge evaluation. These documented prompts serve as a standardized reference for reproducing experiments, diagnosing model behaviors, and extending the benchmark to new financial analytical scenarios.

\subsection{Data Generation Instructions and Prompts}
\label{sec:data_generation_prompts}
This subsection presents the prompts used for benchmark data construction. 
The prompts are designed to transform raw financial reports, structured records, and expert-written materials into standardized benchmark instances with explicit supervision signals. 
Depending on the task setting, the generation prompts are used to identify salient evidence spans, introduce controlled perturbations, construct structured question--answer pairs, or produce reference outputs in JSON format. 

\subsubsection{Semantic Consistency}
The prompts in this section are designed for the \textbf{Terminology Confusion} error under the \textbf{Terminology} category. For this track, the prompts used to inject each error type vary slightly in content depending on the category, but their overall structure remains similar.
\begin{promptbox}
You are a financial language expert and a text generation model. Your task is to inject "terminology confusion" type errors into the input long text in a controlled manner while preserving the overall logical coherence, natural readability of the sentences, and the original discourse structure as much as possible. You must not explain the task, provide examples, or supplement any definitions. Simply perform the error injection directly and return the result.

"Terminology confusion" refers to replacing an originally correct professional term in the current context with another term that is semantically similar but differs in definition, scope of application, or conceptual hierarchy, thereby creating conceptual confusion without making the sentence obviously distorted or unnatural.

You must ensure that:
- only the parts involving professional terminology are modified;
- the overall logic, tone, writing style, and linguistic naturalness of the original text are preserved, and no irrelevant polishing, deletion, expansion, or reorganization is allowed;
- 1--3 terminology confusion errors are injected into the text, and these errors should be distributed across different sentences or paragraphs;
- the injected errors must appear superficially plausible and naturally concealed, yet still be identifiable as incorrect by readers with domain expertise;
- except for the local positions where terminology substitutions are made, all other content must remain unchanged as much as possible, ideally word for word;
- the returned result must strictly follow the JSON structure below, and no additional explanation or example may be added:
{{
  "errors": [
    {{
      "error_type": "Terminology Confusion",
      "original_term": "the original correct term or expression",
      "wrong_term": "the injected confusing term or expression",
      "context": "the complete sentence containing the incorrect term",
      "reason": "briefly explain why this term does not fit the current context and what the correct concept should be"
    }}
  ],
  "modified_text": "the full text after error injection"
}}

Original text:
{text}
\end{promptbox}

\subsubsection{Data Alignment}
\label{sec:da_calc_rule}
To ensure that model reasoning is grounded in verifiable financial data, the Data Alignment track implements a suite of predefined executable rules and derived indicator formulas. We list 20 derived indicator calculations and 5 reasoning rules among them below.

\paragraph{Derived Indicator Calculations}

To support composite metric computation in the Data Alignment track, we operationalize a set of derived financial indicators that combine multiple base fields from the A-share market database. Each formula is executed automatically against structured records, ensuring that every computed value is deterministically traceable to the underlying data. These indicators span fundamental analysis, valuation assessment, capital flow measurement, and intraday price dynamics. We present the complete set of 19 derived indicators below, organized by functional category.

\subparagraph{Fundamental and Solvency Indicators}
These indicators assess a firm's financial health, earnings quality, and capital structure using fiscal-period data.

\noindent\textbf{Cash-flow Coverage Ratio.} This metric measures a firm's ability to cover its current liabilities with internally generated operating cash flow. Operating cash flow is proxied by per-share operating cash flow multiplied by total A-share capital, and the result is scaled by current liabilities:
\begin{equation}
\text{Cash-flow Coverage} = \frac{\texttt{s\_fa\_ocfps} \times \texttt{total\_ashares}}{\texttt{tot\_cur\_liab}}.
\end{equation}

\noindent\textbf{Quality of Earnings.} This indicator captures the proportion of reported net profit that is backed by actual operating cash flow, reflecting earnings quality. It is computed as the ratio of operating cash flow to net profit excluding minority interest:
\begin{equation}
\text{Quality of Earnings} = \frac{\texttt{s\_fa\_ocfps} \times \texttt{total\_ashares}}{\texttt{net\_profit\_excl\_min\_int\_inc}}.
\end{equation}

\noindent\textbf{Net Operating Liabilities.} This metric quantifies the net short-term operating obligations of a firm by subtracting current assets from current liabilities:
\begin{equation}
\text{Net Operating Liabilities} = \texttt{tot\_cur\_liab} - \texttt{tot\_cur\_assets}.
\end{equation}

\noindent\textbf{Net Debt to Market Value.} This leverage indicator normalizes net operating liabilities by the firm's total market value, providing a market-based assessment of debt burden:
\begin{equation}
\text{Net Debt to MV} = \frac{\texttt{tot\_cur\_liab} - \texttt{tot\_cur\_assets}}{\texttt{s\_val\_mv}}.
\end{equation}

\noindent\textbf{Core Profit Margin.} This profitability metric isolates the proportion of core operating profit relative to total revenue. It is defined as deducted profit divided by operating revenue:
\begin{equation}
\text{Core Profit Margin} = \frac{\texttt{s\_fa\_deductedprofit}}{\texttt{oper\_rev}}.
\end{equation}

\noindent\textbf{Operating Leverage.} This indicator captures the sensitivity of operating profit growth to revenue growth, reflecting the degree of fixed-cost exposure in a firm's cost structure. It is computed as the quarterly growth rate of operating profit divided by the quarterly growth rate of operating revenue:
\begin{equation}
\text{Operating Leverage} = \frac{\texttt{fin\_growth\_q\_operprofitgrowrate}}{\texttt{fin\_growth\_q\_torgrowrate}}.
\end{equation}

\subparagraph{Valuation and Market-Based Indicators}
These indicators relate financial performance to market valuation, supporting cross-sectional comparison and investment analysis.

\noindent\textbf{Revenue to Market Value.} This indicator inverts the conventional price-to-sales ratio, expressing operating revenue relative to market value. A higher ratio indicates greater revenue per unit of market capitalization.
\begin{equation}
\text{Revenue to MV} = \frac{\texttt{oper\_rev}}{\texttt{s\_val\_mv}}.
\end{equation}

\noindent\textbf{Market-Cap-Weighted ROE.} This indicator scales return on equity by firm size, computed as the product of ROE and total market value. It is designed to support portfolio-level aggregation and cross-sectional weighting:
\begin{equation}
\text{Market-Cap-Weighted ROE} = \texttt{fin\_profit\_roe} \times \texttt{s\_val\_mv}.
\end{equation}

\noindent\textbf{Free-Float Turnover Value.} This metric estimates the total monetary value of shares traded among freely floating shares. It is defined as the product of the turnover rate of free shares, the number of free-float shares, and the share price proxied by total market value divided by total A-share capital:
\begin{equation}
\text{Free-Float Turnover Value} = \texttt{turn\_freeshares} \times \texttt{free\_float\_shares} \times \frac{\texttt{s\_val\_mv}}{\texttt{total\_ashares}}.
\end{equation}

\subparagraph{Operating Asset and Return Proxies}
These indicators construct proxy measures for operating assets and the corresponding return on invested capital, enabling analysis when detailed segment data are unavailable.

\noindent\textbf{Operating Assets Proxy.} This metric approximates the value of operating assets by combining tangible assets (\texttt{s\_fa\_tangibleasset}) with net working capital, defined as the difference between current assets (\texttt{tot\_cur\_assets}) and current liabilities (\texttt{tot\_cur\_liab}):
\begin{equation}
\text{Operating Assets Proxy} = \texttt{s\_fa\_tangibleasset} + (\texttt{tot\_cur\_assets} - \texttt{tot\_cur\_liab}).
\end{equation}

\noindent\textbf{ROIC Proxy.} This indicator estimates return on invested capital by dividing net profit excluding minority interest (\texttt{net\_profit\_excl\_min\_int\_inc}) by the operating assets proxy defined above:
\begin{equation}
\text{ROIC Proxy} = \frac{\texttt{net\_profit\_excl\_min\_int\_inc}}{\texttt{s\_fa\_tangibleasset} + (\texttt{tot\_cur\_assets} - \texttt{tot\_cur\_liab})}.
\end{equation}

\subparagraph{Intraday Price Dynamics Indicators}
These indicators decompose daily price movements using open, high, low, close, and pre-close prices, capturing intraday patterns, overnight effects, and trading efficiency.

\noindent\textbf{Intraday Return.} This metric captures the return generated during regular trading hours, measured from the daily open to the daily close:
\begin{equation}
\text{Intraday Return} = \frac{\texttt{daily\_close} - \texttt{daily\_open}}{\texttt{daily\_open}}.
\end{equation}

\noindent\textbf{Overnight Return (Opening Gap).} This indicator measures the return attributable to overnight information accumulation, computed as the change from the previous close to the current open:
\begin{equation}
\text{Overnight Return} = \frac{\texttt{daily\_open} - \texttt{daily\_preclose}}{\texttt{daily\_preclose}}.
\end{equation}

\noindent\textbf{Full-Day Return.} This metric captures the total daily return from the previous close to the current close, encompassing both overnight and intraday components:
\begin{equation}
\text{Full-Day Return} = \frac{\texttt{daily\_close} - \texttt{daily\_preclose}}{\texttt{daily\_preclose}}.
\end{equation}

\noindent\textbf{Price Efficiency Ratio (Real Body Strength).} This indicator quantifies the proportion of the daily price range represented by the real body, measuring how efficiently price movement translates into a net directional close. It is defined as the absolute distance from open to close scaled by the full high--low range:
\begin{equation}
\text{Price Efficiency Ratio} = \frac{\texttt{daily\_close} - \texttt{daily\_open}}{\texttt{daily\_high} - \texttt{daily\_low}}.
\end{equation}

\noindent\textbf{Upper Shadow Ratio.} This metric measures the relative length of the upper shadow as a proportion of the total daily price range, capturing the extent of intraday selling pressure that reversed from the high:
\begin{equation}
\text{Upper Shadow Ratio} = \frac{\texttt{daily\_high} - \max(\texttt{daily\_open}, \texttt{daily\_close})}{\texttt{daily\_high} - \texttt{daily\_low}}.
\end{equation}

\noindent\textbf{Lower Shadow Ratio.} This metric measures the relative length of the lower shadow as a proportion of the total daily price range, capturing the extent of intraday buying support that reversed from the low:
\begin{equation}
\text{Lower Shadow Ratio} = \frac{\min(\texttt{daily\_open}, \texttt{daily\_close}) - \texttt{daily\_low}}{\texttt{daily\_high} - \texttt{daily\_low}}.
\end{equation}

\noindent\textbf{Close Relative Position.} This indicator locates the closing price within the day's high--low range, providing a normalized measure of closing strength. Values near 1 indicate a close near the high, while values near 0 indicate a close near the low:
\begin{equation}
\text{Close Relative Position} = \frac{\texttt{daily\_close} - \texttt{daily\_low}}{\texttt{daily\_high} - \texttt{daily\_low}}.
\end{equation}

\subparagraph{Volume-Price Interaction Indicators}
These indicators relate price changes to trading volume, offering insight into the intensity and efficiency of market participation.

\noindent\textbf{Average Trade Amount.} This metric estimates the average monetary value per transaction by dividing total daily turnover amount by the number of transactions:
\begin{equation}
\text{Avg.\ Trade Amount} = \frac{\texttt{daily\_amount}}{\texttt{daily\_tnum}}.
\end{equation}

\noindent\textbf{Price Change per Unit Volume.} This indicator measures the marginal price impact per unit of trading volume, computed as the daily price change divided by total daily volume:
\begin{equation}
\text{Price Change per Volume} = \frac{\texttt{daily\_close} - \texttt{daily\_preclose}}{\texttt{daily\_volume}}.
\end{equation}

\paragraph{Executable Reasoning Rules}

The reasoning rules in the Data Alignment track encode established financial analysis practices and industry heuristics into atomic, executable steps. Each rule is grounded in a recognized financial principle, operates over structured database fields, and is decomposed into a sequence of calculator-level operations. All rules produce a binary output that can be objectively verified against the database. Below we present representative examples drawn from the full set of 61 rules spanning capital flow analysis, technical signal detection, valuation assessment, and fundamental quality screening.

\subparagraph{Chaikin Money Flow Zero-Line Crossover}
This rule determines whether the Chaikin Money Flow (CMF) indicator signals net capital inflow over the evaluation window. It is grounded in the zero-line crossover principle of CMF interpretation: $\text{CMF} > 0$ indicates net capital inflow with the market dominated by buying pressure. When the condition holds, the rule concludes that net capital inflow is present, buying power dominates the market, and buyers exhibit strong willingness to initiate purchases. Acceptable phrasings of this conclusion include ``net capital inflow; the market is dominated by buyers,'' ``buying pressure dominates the market,'' ``capital is flowing in on a net basis,'' ``the market is in a state of capital inflow,'' and ``buyers exhibit strong willingness to initiate purchases.''

\textbf{Required Metrics.}
{\ttfamily daily\_high, daily\_low, daily\_close, daily\_volume}

The rule proceeds through the following atomic steps:
\begin{enumerate}[leftmargin=*,itemsep=2pt]
    \item \textbf{Money Flow Multiplier.} For each trading day $t$ in the 20-day window, compute:
    \begin{equation}
    \text{MFM}_t = \frac{(\text{Close}_t - \text{Low}_t) - (\text{High}_t - \text{Close}_t)}{\text{High}_t - \text{Low}_t}.
    \end{equation}
    This multiplier ranges from $+1$ (close at high, strong buying) to $-1$ (close at low, strong selling).

    \item \textbf{Money Flow Volume.} Scale the period's volume by the money flow multiplier:
    \begin{equation}
    \text{MFV}_t = \text{MFM}_t \times \text{Volume}_t.
    \end{equation}

    \item \textbf{Chaikin Money Flow.} Aggregate money flow volume over 20 days and normalize by total volume:
    \begin{equation}
    \text{CMF} = \frac{\sum_{t=1}^{20} \text{MFV}_t}{\sum_{t=1}^{20} \text{Volume}_t}.
    \end{equation}

    \item \textbf{Decision Rule.} Evaluate the zero-line condition:
    \begin{equation}
    \text{CMF} > 0.
    \end{equation}
\end{enumerate}

\subparagraph{RSI Overbought Signal}
This rule determines whether the Relative Strength Index (RSI) indicates an overbought condition over the evaluation window. It is grounded in the traditional RSI threshold principle: $\text{RSI} > 70$ signals that the market may be overbought, prices have risen too rapidly, and there is elevated risk of a short-term pullback. When the condition holds, the rule concludes that the market is overbought with increased probability of a near-term correction. Acceptable phrasings of this conclusion include ``RSI indicates an overbought condition with short-term pullback risk,'' ``the relative strength index signals overbought, cautioning against a technical correction,'' ``the market is overheated in the short term with rising adjustment probability,'' and ``RSI overbought signal warns of near-term risk.''

\textbf{Required Metrics.}
{\ttfamily daily\_close}

The rule proceeds through the following atomic steps:
\begin{enumerate}[leftmargin=*,itemsep=2pt]
    \item \textbf{Price Changes.} For each of the most recent 14 trading days $t$, compute the daily price change relative to the previous close:
    \begin{equation}
    \Delta P_t = \text{Close}_t - \text{Close}_{t-1}.
    \end{equation}

    \item \textbf{Average Gain and Average Loss.} Separate the price changes into gains ($U_t$) and losses ($D_t$), then compute their 14-day exponential moving averages:
    \begin{align}
    U_t &= \max(\Delta P_t, 0), \quad D_t = \max(-\Delta P_t, 0), \\[4pt]
    \text{AvgGain} &= \frac{1}{14}\sum_{i=1}^{14} U_i, \quad \text{AvgLoss} = \frac{1}{14}\sum_{i=1}^{14} D_i.
    \end{align}

    \item \textbf{Relative Strength and RSI.} Compute the relative strength ratio and the RSI value:
    \begin{equation}
    \text{RS} = \frac{\text{AvgGain}}{\text{AvgLoss}}, \qquad \text{RSI} = 100 \times \frac{\text{AvgGain}}{\text{AvgGain} + \text{AvgLoss}} = 100 - \frac{100}{1 + \text{RS}}.
    \end{equation}

    \item \textbf{Decision Rule.} Evaluate the overbought threshold condition:
    \begin{equation}
    \text{RSI} > 70.
    \end{equation}
\end{enumerate}

\subparagraph{KDJ Overbought Signal}
This rule determines whether the KDJ stochastic oscillator indicates an overbought condition over the evaluation window. It is grounded in the traditional stochastic oscillator threshold principle: $K > 80$ and $D > 80$ jointly signal that the indicator is in the overbought zone, suggesting that price appreciation may be excessive and the risk of a short-term pullback is elevated. When the condition holds, the rule concludes that the market is overbought according to KDJ with a heightened probability of near-term correction. Acceptable phrasings of this conclusion include ``KDJ indicates an overbought condition with elevated short-term pullback risk,'' ``the stochastic oscillator signals overbought, cautioning against a price correction,'' ``both K and D values are at elevated levels, signaling an overheated market,'' ``KDJ overbought zone suggests reducing positions or staying on the sidelines,'' and ``technical indicators suggest limited short-term upside potential.''

\textbf{Required Metrics.}
{\ttfamily daily\_high, daily\_low, daily\_close}

The rule proceeds through the following atomic steps:
\begin{enumerate}[leftmargin=*,itemsep=2pt]
    \item \textbf{Raw Stochastic Value (RSV).} For the most recent trading day, compute the RSV over a 9-day lookback window:
    \begin{equation}
    \text{RSV} = 100 \times \frac{\text{Close} - \text{Lowest}_9}{\text{Highest}_9 - \text{Lowest}_9},
    \end{equation}
    where $\text{Lowest}_9$ and $\text{Highest}_9$ denote the minimum low and maximum high over the preceding 9 trading days, respectively.

    \item \textbf{Fast \%K.} Compute the current $K$ value as an exponentially weighted moving average of the RSV sequence. Under the standard 3-period smoothing with factor $1/3$, the recursive formulation is:
    \begin{equation}
    K_t = \frac{2}{3} K_{t-1} + \frac{1}{3} \text{RSV}_t.
    \end{equation}

    \item \textbf{Slow \%D.} Compute the current $D$ value as a smoothed version of the $K$ line under the same weighting scheme:
    \begin{equation}
    D_t = \frac{2}{3} D_{t-1} + \frac{1}{3} K_t.
    \end{equation}

    \item \textbf{Decision Rule.} Evaluate the joint overbought threshold condition:
    \begin{equation}
    K > 80 \quad \land \quad D > 80.
    \end{equation}
\end{enumerate}

\subparagraph{Moving Average Bullish Alignment}
This rule determines whether the 5-day, 10-day, and 20-day simple moving averages (MA5, MA10, MA20) exhibit a bullish alignment pattern. It is grounded in the classical moving average alignment principle: $\text{MA5} > \text{MA10} > \text{MA20}$ indicates that short-term price momentum exceeds medium-term momentum, which in turn exceeds long-term momentum, signaling a strong and stable uptrend with robust bullish continuation. When the condition holds, the rule concludes that the market is in a strong upward trend with firmly established technical strength and a clear directional bias. Acceptable phrasings of this conclusion include ``the market is in a strong uptrend with a powerful bullish continuation signal,'' ``short-, medium-, and long-term moving averages are bullishly aligned, confirming technical strength,'' and``the moving average system exhibits a perfect upward structure with clear trend direction.''

\textbf{Required Metrics.}
{\ttfamily daily\_close}

The rule proceeds through the following atomic steps:
\begin{enumerate}[leftmargin=*,itemsep=2pt]
    \item \textbf{Simple Moving Averages.} For the most recent trading day $t$, compute the SMA values over three rolling windows from the daily closing prices:
    \begin{align}
    \text{MA5}_t  &= \frac{1}{5}  \sum_{i=0}^{4}  \text{Close}_{t-i}, \\[4pt]
    \text{MA10}_t &= \frac{1}{10} \sum_{i=0}^{9}  \text{Close}_{t-i}, \\[4pt]
    \text{MA20}_t &= \frac{1}{20} \sum_{i=0}^{19} \text{Close}_{t-i}.
    \end{align}
    To ensure robustness of the MA20 computation, the evaluation window requires at least 25 prior closing prices, allowing the 20-period SMA to stabilize without relying on the target day as the sole terminal point.

    \item \textbf{Bullish Alignment Check.} Verify that the three moving averages satisfy a strict descending order:
    \begin{equation}
    \text{MA5}_t > \text{MA10}_t > \text{MA20}_t.
    \end{equation}

    \item \textbf{Decision Rule.} Confirm that all pairwise inequalities jointly hold:
    \begin{align}
    \text{MA5}_t  &> \text{MA10}_t, \\
    \text{MA10}_t &> \text{MA20}_t.
    \end{align}
    The conjunction of these two conditions is logically equivalent to the full bullish alignment criterion.
\end{enumerate}

\subparagraph{MACD Golden Cross}
This rule determines whether a MACD golden cross has occurred at the most recent decision point. It is grounded in the classical MACD crossover principle: a golden cross occurs when the DIF line (fast) crosses above the DEA line (slow) from below, signaling that upward momentum is strengthening and a buy signal is triggered. A minimum crossover magnitude of 0.02 is enforced to filter out insignificant crossings caused by noise. When the condition holds, the rule concludes that a buy signal is confirmed, upward momentum is building, and a bullish trend may be initiating. Acceptable phrasings of this conclusion include ``a buy signal is present with strengthening upward momentum,'' ``MACD golden cross confirmed, signaling near-term technical improvement,'' ``upward momentum is beginning to accumulate,'' ``MACD issues a buy signal as bullish forces strengthen,'' and ``an uptrend may be starting.''

\textbf{Required Metrics.}
{\ttfamily daily\_close}

The rule proceeds through the following atomic steps:
\begin{enumerate}[leftmargin=*,itemsep=2pt]
    \item \textbf{Exponential Moving Averages.} For each trading day $t$ in a 30-day window, compute the 12-period and 26-period exponential moving averages from the daily closing prices. Under the standard EMA recursion with smoothing factor $\alpha = \frac{2}{n+1}$, we have:
    \begin{align}
    \text{EMA12}_t &= \frac{2}{13} \cdot \text{Close}_t + \frac{11}{13} \cdot \text{EMA12}_{t-1}, \\[4pt]
    \text{EMA26}_t &= \frac{2}{27} \cdot \text{Close}_t + \frac{25}{27} \cdot \text{EMA26}_{t-1}.
    \end{align}
    A 30-day lookback window is required to ensure stable initialization of both EMA sequences.

    \item \textbf{DIF Line.} Compute the DIF (fast) line as the difference between the short- and long-period EMAs:
    \begin{equation}
    \text{DIF}_t = \text{EMA12}_t - \text{EMA26}_t.
    \end{equation}

    \item \textbf{DEA Line.} Compute the DEA (slow) line as a 9-period EMA of the DIF sequence, with smoothing factor $\beta = \frac{2}{9+1} = 0.2$:
    \begin{equation}
    \text{DEA}_t = 0.2 \cdot \text{DIF}_t + 0.8 \cdot \text{DEA}_{t-1}.
    \end{equation}

    \item \textbf{Crossover Detection.} Check whether DIF has crossed above DEA at the current time point relative to the preceding day:
    \begin{equation}
    \text{DIF}_{t-1} < \text{DEA}_{t-1} \quad \land \quad \text{DIF}_t > \text{DEA}_t.
    \end{equation}

    \item \textbf{Magnitude Filter.} To suppress noise-induced false signals from minuscule crossovers, require that the absolute post-cross separation exceeds a minimum threshold:
    \begin{equation}
    |\,\text{DIF}_t - \text{DEA}_t\,| \geq 0.02.
    \end{equation}

    \item \textbf{Decision Rule.} The golden cross is confirmed only when both conditions are jointly satisfied:
    \begin{equation}
    \bigl(\text{DIF}_{t-1} < \text{DEA}_{t-1} \land \text{DIF}_t > \text{DEA}_t\bigr) \;\land\; \bigl(|\,\text{DIF}_t - \text{DEA}_t\,| \geq 0.02\bigr).
    \end{equation}
\end{enumerate}

\subsubsection{Deep Insight}
\promptboxsetup{colback=TableLightPurple,colframe=TableDarkPurple}
\begin{promptbox}
You are a top-tier financial analyst and also a professional AI benchmark designer. Your task is to transform a piece of original text describing a specific financial event into a high-quality evaluation data that can rigorously test and effectively differentiate downstream AI models in [category].

You must strictly follow all the rules below. The quality of your output will be judged by whether it can enable a strong AI model to achieve a high score in subsequent evaluations.

I. Core Task Definition
Identify the core “event” in the original text, design an open-ended analytical question around that event, and provide a “model ground-truth answer” (Gold Answer). This answer must not only be correct, but also demonstrate outstanding analytical quality.

II. Output Format Rules
1. Your only output must be a valid JSON object containing exactly one key, "qa_pairs", whose value is a list.
2. Each element in the list must be a dictionary containing the three keys: "question", "evidence", and "ground_truth".

III. Content Generation Rules
1. evidence:
   - It must be a long passage copied verbatim from the original text without deletion;
   - The passage should fully describe one event;
   - Its purpose is to provide sufficient yet somewhat redundant context, forcing downstream models to read, understand, and distill the information.

2. question:
   - It must be an open-ended analytical instruction;
   - The question should guide the model to produce a deep, comprehensive, and well-structured response, rather than merely extracting information.

3. gold_answer:
   - It must be a structured, in-depth analytical essay with critical thinking, rather than a simple list of bullet points;
   - It must be a single string;
   - Its internal arguments must be organized using numbered points, such as “1. ... 2. ... 3. ...”.
   - [prompts_for_categories](Table 6)


IV. Quality and Quantity Control
1. A data point should only be created when the event described in the evidence is suitable for deep analysis.
2. For a single report, generating 1 to 3 truly high-quality analytical data points is ideal.
3. Never sacrifice analytical depth, structural completeness, or critical thinking for the sake of quantity.

V. Reference Style Guidance
A high-quality ground_truth should generally have the following characteristics: it first summarizes the nature of the event, then analyzes its positive impacts, mechanisms of influence, and transmission paths to market or financial outcomes in numbered points, and further discusses potential risks, execution uncertainty, or common misinterpretations, before ending with a balanced concluding judgment.

Now, act as this top-tier benchmark designer and begin your work based on the following research report content:

--- Report Begins ---
{report_content}
--- Report Ends ---
\end{promptbox}

\begin{table}[htbp]
\centering
\caption{Dimension-specific gold-answer design principles for the Deep Insight track.}
\label{tab:di_gold_answer_design}
\scriptsize
\setlength{\tabcolsep}{5pt}
\begin{tabular}{p{0.18\textwidth} p{0.76\textwidth}}
\toprule
\textbf{Dimension} & \textbf{Design Principles} \\
\midrule

\textbf{Justification and causal depth}
&
\textbf{Thinking Model:} Multi-stage transmission path analysis. \par
\textbf{Task Objective:} Identify the initial driving variable in the report and require the model to infer its downstream impact on the firm's financial performance or industry position. \par
\textbf{Prompting Technique:} The instruction should force the model to build a long reasoning chain of Variable A $\rightarrow$ Intermediate Mechanism B $\rightarrow$ Financial Outcome C. \par
\textbf{Question Design Requirement:} The background facts should include driving factors, such as policy changes or technological breakthroughs. The instruction should ask how such a factor penetrates the business layer and is ultimately reflected in the financial statements. \par
\textbf{Answer Scoring Rubric (100-point scale):} [40] Logical chain: whether a clear and closed A$\rightarrow$B$\rightarrow$C chain is established without logical jumps. [30] Causal depth: whether the answer explains why the transmission is necessary or highly probable. [30] Professionalism: whether the terminology is precise and consistent with CFA-style analytical reporting.
\\
\midrule

\textbf{Factuality and contextualization}
&
\textbf{Thinking Model:} Data anchoring and quantitative verification. \par
\textbf{Task Objective:} Evaluate the model's ability to extract, understand, and apply hard data from the report. \par
\textbf{Prompting Technique:} The instruction should require the model to use specific numerical values in the background facts, such as gross margin, CAGR, order amount, or market share, for secondary calculation or logical argumentation. \par
\textbf{Question Design Requirement:} The background facts must contain at least 3--5 groups of core financial or operational data. The instruction should include requirements such as ``quantitatively assess the extent to which ... is supported by the above data'' or ``compare and analyze the marginal changes in ...''. \par
\textbf{Answer Scoring Rubric (100-point scale):} [40] Fact density: whether all key numerical values in the question are accurately cited and interpreted. [30] Computational rigor: whether the accounting and quantitative relations across data points are correct. [30] Contextual analysis: whether the answer explains the meaning of the data under the relevant industry context.
\\
\midrule

\textbf{Completeness and comparative analysis}
&
\textbf{Thinking Model:} Dialectical conflict and margin-of-safety analysis. \par
\textbf{Task Objective:} Identify the tension between optimistic expectations and potential constraints in the report, and evaluate the model's critical thinking ability. \par
\textbf{Prompting Technique:} The instruction should force the model to identify risk disclosures, prerequisite conditions, or threats from competitors in the report. \par
\textbf{Question Design Requirement:} The background facts must include both positive and negative information, such as new capacity expansion versus slowing industry demand growth. The instruction should ask the model to evaluate under what conditions the positive logic would be falsified. \par
\textbf{Answer Scoring Rubric (100-point scale):} [40] Critical thinking: whether the answer identifies boundary conditions or downside risks under which the conclusion may fail. [30] Comparative perspective: whether horizontal peer comparison or vertical historical comparison is conducted. [30] Argument balance: whether the answer remains objective and avoids blind optimism.
\\
\midrule

\textbf{Architectural richness and strictness}
&
\textbf{Thinking Model:} Systematic industry analysis framework. \par
\textbf{Task Objective:} Reorganize fragmented facts into a highly professional and structurally rigorous industry analysis. \par
\textbf{Prompting Technique:} The instruction should require the model to use professional analytical frameworks, such as SWOT, Porter's Five Forces, PEST, or the BCG Matrix, for integrated discussion. \par
\textbf{Question Design Requirement:} The background facts should cover multi-dimensional information. The instruction should explicitly ask the model to provide a systematic and in-depth analysis using a professional analytical framework. \par
\textbf{Answer Scoring Rubric (100-point scale):} [40] Framework completeness: whether an appropriate analytical framework is adopted and whether the coverage of dimensions is comprehensive. [30] Logical hierarchy: whether the response follows a clear overall--particular--overall structure with explicit nesting. [30] Granularity of expression: whether concepts are defined with sufficient depth and rigor, reaching the standard of top-tier institutions.
\\

\bottomrule
\end{tabular}
\end{table}

\subsection{Benchmark Evaluation Prompts}\label{sec:evaluation_prompts}
This subsection provides the prompts used to query models during benchmark evaluation. 
Unlike the data generation prompts, these prompts serve as the actual task instructions given to the evaluated models at inference time. 
They are designed to elicit task-specific responses under standardized conditions, including error detection and correction, structured data reasoning, and evidence-grounded financial analysis. 
To improve comparability across models, the prompts explicitly define the required role, the available input context, and the expected output format, often with strict JSON constraints for automatic parsing and scoring. 

\subsubsection{Semantic Consistency}
The prompts in this section are designed for the \textbf{Terminology} category. For this track, the prompts used to evaluate each category vary slightly in content, but their overall structure remains similar, as illustrated below.
\promptboxsetup{colback=TableLightBlue,colframe=TableDarkBlue}
\begin{promptbox}
You are a professional financial fact-checking expert specializing in identifying factual errors in text. Please analyze the text strictly based on the given question and return the result only in valid JSON format. Do not output any additional explanation, commentary, or examples.

If errors are detected, you must strictly return the result in the following format:
{
  "error_exists": "Yes",
  "errors": [
    {
      "error_location": "the complete sentence containing the error",
      "wrong_content": "the incorrect term or expression",
      "correct_content": "the correct term or expression",
      "error_type": "Terminology Confusion Error",
      "reason": "an explanation of why the term does not fit the current context and what the correct concept should be"
    }
  ],
  "explanation": "a comprehensive explanation of all terminology confusion errors in the text",
  "corrected_text": "the fully corrected text"
}

If no terminology confusion errors are detected, you must strictly return the result in the following format:
{
  "error_exists": "No",
  "errors": [],
  "explanation": "No terminology confusion errors were found in the text.",
  "corrected_text": "the original text"
}

Please begin your analysis based on the following question:
{question}
\end{promptbox}

\subsubsection{Data Alignment}
\promptboxsetup{colback=TableLightGreen,colframe=TableDarkGreen}
Prompts for step 1: query data.
\begin{promptbox}
You are a financial data analysis assistant. To answer a question, you need to determine which data should be queried from the database.

Company Information
{company_info_text}

Description of Database Fields
{field_descriptions_text}

Description of Database Table Schema
Main table: stock_jbm_factors
- Main fields: id (row ID), stcode (stock code), dt (trading date), and 90+ other financial and valuation indicator fields
- Each company has multiple records, and each record corresponds to one trading day (field: dt)
- The query will automatically use the current company's stcode: {stcode}

Question
{question}

Task
Please analyze this question and decide what data needs to be queried in order to answer it. Please consider:
1. Which fields (indicators) need to be queried? Please provide a list of field names.
2. What date range or specific dates need to be queried?

Please strictly return your query plan in the following JSON format, without adding any other explanatory text:

```json
{{
  "fields_to_query": ["List of fields to query, for example: [\"fin_valuation_ep_ttm_1y_quantile\", \"fin_profit_roe\", \"s_fa_roe\"]"],
  "date_range": {{
    "start_date": "Start date in YYYY-MM-DD format. Fill this if a date range is needed; otherwise null",
    "end_date": "End date in YYYY-MM-DD format. Fill this if a date range is needed; otherwise null",
    "specific_dates": ["List of specific dates in YYYY-MM-DD format. Fill this if specific dates are needed; otherwise null or an empty array"]
  }},
  "reasoning": "Explain why these data fields are needed and why this date range or these specific dates were chosen"
}}

Important Notes:
- fields_to_query must be an array of strings containing all field names to be queried
- If only specific dates are needed, fill in the specific_dates array (for example: ["2020-01-15", "2020-06-30"]), and set start_date and end_date to null
- If a date range is needed, fill in start_date and end_date (for example: start_date: "2020-01-01", end_date: "2020-12-31"), and set specific_dates to null or an empty array
- You must not use both a date range and a list of specific dates at the same time
- The date format must be YYYY-MM-DD
Please ensure that the returned result is valid JSON and can be parsed directly.
\end{promptbox}

Prompts for step 2: answer questions.
\begin{promptbox}
You are a financial data analysis assistant. Please answer the following question based on the provided data.

Company Information
{company_info_text}

Field Descriptions
{field_descriptions_text}

Data Table
{json.dumps(data_records, ensure_ascii=False, indent=2)}

Question
{question}

Requirements
Please strictly return your answer in the following JSON format, without adding any other explanatory text:

```json
{{
  "answer": "\"Yes\" or \"No\"",
  "calculation_process": "A detailed explanation of the calculation steps, showing each step of the calculation process",
  "formulas_used": [
    {{
      "formula": "Formula expression",
      "description": "Explanation of what the formula means"
    }}
  ],
  "data_used": [
    {{
      "id": "Row ID",
      "field": "Field name",
      "value": "Field value",
      "stcode": "Stock code"
    }}
  ]
}}

Important Notes:
- The answer field must be "Yes" or "No"
- The "calculation_process" field should describe the calculation steps in detail
- "formulas_used" must be an array containing all formulas used
- "data_used" must be an array containing all data used, and each item must include the fields id, field, value, and stcode
- Please ensure that the returned result is valid JSON and can be parsed directly
\end{promptbox}

\subsubsection{Deep Insight}
\promptboxsetup{colback=TableLightPurple,colframe=TableDarkPurple}
\begin{promptbox}
You are a professional financial analyst. Please analyze and answer the given question strictly based on the provided background facts. Your response should draw on the key information in the background facts, demonstrate analytical reasoning, and remain accurate, relevant, and evidence-based. Do not make unsupported extensions beyond the provided facts.

[Background Facts]
{evidence}

[Analytical Question]
{question}
\end{promptbox}

\subsection{LLM-as-a-Judge Evaluation Prompts}\label{sec:judge_prompts}
This subsection describes the prompts used in the LLM-as-a-Judge evaluation framework. 
These prompts are designed to assess model outputs by comparing candidate answers against reference answers under explicitly defined scoring rubrics. 
This structured prompting strategy enables the judge model to produce transparent, fine-grained, and reproducible assessments of analytical quality, factual grounding, critical comparison, and discourse organization.
\promptboxsetup{colback=TableLightBlue,colframe=TableDarkBlue}
\subsubsection{Semantic Consistency}
\begin{promptbox}
You are a professional evaluation expert, specializing in assessing text quality from multiple dimensions. Please carefully analyze the given content and provide a detailed evaluation according to the requirements below. Based on the model's complete output and the reference answer provided below, conduct a comprehensive evaluation of the model.

The model's complete output (including error judgment, error list, error reasons, and corrected text):
{model_json}

Reference answer (including error list, error explanations, and corrected text):
{ref_json}

Description of evaluation dimensions:
Please score the error reasoning and corrected content separately across the following four dimensions (each on a scale of 0–10, rounded to one decimal place), and provide a detailed explanation for each dimension:
1. Correctness: Evaluate whether the model accurately understood the error and whether it provided the correct reason or correction. This measures the degree of alignment with the reference answer.
2. Reasoning Rigor: Examine whether the reasoning process is clear, the logic is coherent, the argumentation is sufficient, and whether the correction is supported by sound reasoning.
3. Relevance: Evaluate whether the explanation or correction directly addresses the specific error, whether it is closely related to the erroneous content, and whether there is any irrelevant or off-topic content.
4. Completeness: Check whether all important error points or correction points are covered, and whether any key information affecting the conclusion is omitted.

Evaluation targets:
1. The model's error reasoning ("explanation"): Evaluate it across the four dimensions by considering both the error list identified by the model and its corresponding explanations.
2. The model's corrected content ("corrected_text"): Evaluate the fully corrected text across the same four dimensions.

Please strictly return the result in the following JSON format, without adding any extra explanation:
{{
"error_reasoning": {{
    "correctness": {{ "score": score, "explanation": "Explanation of the correctness evaluation" }},
    "reasoning_rigor": {{ "score": score, "explanation": "Explanation of the reasoning rigor evaluation" }},
    "relevance": {{ "score": score, "explanation": "Explanation of the relevance evaluation" }},
    "completeness": {{ "score": score, "explanation": "Explanation of the completeness evaluation" }}
}},
"corrected_text": {{
    "correctness": {{ "score": score, "explanation": "Explanation of the correctness evaluation" }},
    "reasoning_rigor": {{ "score": score, "explanation": "Explanation of the reasoning rigor evaluation" }},
    "relevance": {{ "score": score, "explanation": "Explanation of the relevance evaluation" }},
    "completeness": {{ "score": score, "explanation": "Explanation of the completeness evaluation" }}
}}
}}

Please begin the evaluation:
\end{promptbox}

\subsubsection{Deep Insight}
\label{sec:eval-rubric-di}
In this section, we detail the category-specific evaluation criteria used in the Deep Insight track. 
Each category is further decomposed into three sub-dimensions with explicit scoring rubrics, enabling the judge model to perform fine-grained and structured assessment of analytical depth, factual grounding, critical comparison, and discourse organization.
\promptboxsetup{colback=TableLightPurple,colframe=TableDarkPurple}
\begin{promptbox}
You are a strict financial evaluation expert. Score the Candidate Answer by comparing it against the Reference Answer.

Scoring Criteria
{rubric_prompt_template}

Question Information
Question: {question}
Reference Answer: {gold_answer}

Candidate Answer
{model_answer}

Important Notes
- Output JSON only, and do not include any other content
- The explanation must be based on the scoring criteria and should identify the key reasons for the score
- The explanation must not exceed 3 sentences
- The score must be an integer from 0 to 5
\end{promptbox}

\begin{table}[htbp]
\centering
\scriptsize
\caption{Category-specific scoring rubric for Justification and Causal Depth.}
\label{tab:jcd_rubric}
\scriptsize
\setlength{\tabcolsep}{2pt}
\begin{tabularx}{\textwidth}{c c >{\raggedright\arraybackslash}X}
\toprule
\textbf{Sub-dimension} & \textbf{Weight} & \textbf{Scoring Criteria (0--5)} \\
\midrule
Logical chain & 40\% &
0: No logical structure at all; statements are disconnected; reasoning does not hold.\\
& &
1: Only an A$\rightarrow$B pattern appears, but without actual derivation; severe logical jumps; incomplete reasoning.\\
& &
2: A rudimentary A$\rightarrow$B$\rightarrow$C chain appears, but it is loose; key logic is missing; explanation is insufficient.\\
& &
3: A basic A$\rightarrow$B$\rightarrow$C closed loop is formed; the logic is clear but limited in depth.\\
& &
4: The A$\rightarrow$B$\rightarrow$C structure is complete and rigorous; causal progression is natural; no obvious flaws.\\
& &
5: Multi-layer reasoning chains (e.g., A$\rightarrow$B$\rightarrow$C$\rightarrow$D); internally coherent; identifies premises and implicit assumptions; demonstrates a professional reasoning system. \\
\midrule
Causal depth & 30\% &
0: The causal relation is wrong or nonexistent; correlation is mistaken for causation.\\
& &
1: Only phenomena are described; no causal explanation is provided.\\
& &
2: Provides a preliminary causal mechanism, but lacks an argument for necessity; remains at the surface level.\\
& &
3: Clearly explains the mechanism by which A leads to B, but does not analyze conditions, boundaries, or constraints.\\
& &
4: Explains causal necessity; includes triggering conditions, mechanism paths, and boundary conditions.\\
& &
5: Professional-grade causal analysis; explicitly identifies sufficient conditions and uses counterfactual reasoning (e.g., what would happen if A did not occur). \\
\midrule
Professionalism & 30\% &
0: Professional concepts are incorrect; violates basic financial or investment-research common sense.\\
& &
1: Terminology is only partially correct; the reasoning is detached from professional practice.\\
& &
2: Terminology and logic are basically correct, but professional depth is limited.\\
& &
3: Overall consistent with investment-research and financial-statement analysis logic; concepts are used accurately.\\
& &
4: Demonstrates CFA-level rigor; shows risk awareness, clear premises, and disciplined reasoning.\\
& &
5: Reaches the professionalism of an investment institution or sell-side analyst; expression is rigorous; concept definitions are precise; judgments are standardized. \\
\bottomrule
\end{tabularx}
\end{table}

\begin{table}[htbp]
\centering
\scriptsize
\caption{Category-specific scoring rubric for Factuality and Contextualization.}
\label{tab:fc_rubric}
\scriptsize
\setlength{\tabcolsep}{2pt}
\begin{tabularx}{\textwidth}{c c >{\raggedright\arraybackslash}X}
\toprule
\textbf{Sub-dimension} & \textbf{Weight} & \textbf{Scoring Criteria (0--5)} \\
\midrule
Fact density & 40\% &
0: Data are wrong, fabricated, or completely irrelevant to the question.\\
& &
1: Only a very small number of facts are cited; key data are missing; relevance is weak.\\
& &
2: Some key facts are cited, but coverage is incomplete; there are minor inaccuracies or a lack of context.\\
& &
3: The cited information is broadly accurate and covers the main points, but more comprehensive data support is still lacking.\\
& &
4: Factual references are comprehensive and accurate; key financial indicators or core evidence are included; information density is high.\\
& &
5: Data references are precise and rich; they cover core indicators, supporting indicators, trend facts, and so on; the answer reaches the level of a professional analytical report. \\
\midrule
Computational rigor & 30\% &
0: Calculation results are wrong; accounting or logical relations are clearly inconsistent; calculation is essentially absent.\\
& &
1: There is some attempt at calculation or relational judgment, but most of it is not rigorous or lacks necessary steps.\\
& &
2: Calculations are basically correct, but formulas, derivations, or numerical support are insufficient; minor flaws remain.\\
& &
3: The overall calculation process is reasonable; accounting relations are basically consistent; only occasional omissions exist.\\
& &
4: Calculations are rigorous; key accounting chains (e.g., YoY/MoM, profit structure, cash-flow relations) are internally consistent.\\
& &
5: Demonstrates professional-grade rigor; actively validates accounting relations; numerical derivations are solid; there are no logical contradictions. \\
\midrule
Contextual analysis & 30\% &
0: No contextualized explanation is provided; the answer is inconsistent with the industry or firm background.\\
& &
1: Only superficial background information is given; no real contextual linkage is formed.\\
& &
2: Provides a preliminary explanation based on industry characteristics, but the linkage is weak or lacks depth.\\
& &
3: Offers contextual analysis related to the industry or company; the explanation has some depth.\\
& &
4: Contextual analysis is in-depth; it combines data, structure, and industry logic to explain the implications.\\
& &
5: Demonstrates professional-grade contextual reasoning; precisely connects industry mechanisms, corporate strategy, and cyclical logic, and explains the meaning and implications behind the data. \\
\bottomrule
\end{tabularx}
\end{table}
\begin{table}[htbp]
\centering
\caption{Category-specific scoring rubric for Completeness and comparative Analysis.}
\label{tab:ca_rubric}
\scriptsize
\setlength{\tabcolsep}{2pt}
\begin{tabularx}{\textwidth}{c c >{\raggedright\arraybackslash}X}
\toprule
\textbf{Sub-dimension} & \textbf{Weight} & \textbf{Scoring Criteria (0--5)} \\
\midrule
Critical thinking & 40\% &
0: No critical discussion; no risks or assumption limits are identified.\\
& &
1: Merely repeats information mechanically; lacks awareness of risks or boundary conditions.\\
& &
2: Identifies some risks or limitations, but only at a surface level; lacks structured logic.\\
& &
3: Points out key assumptions, boundaries, or sources of risk; shows a basic critical perspective.\\
& &
4: Systematically analyzes risks and limitations; proposes insightful counterexamples or hedging logic.\\
& &
5: Demonstrates professional research- or investment-grade critical thinking; actively examines assumptions, boundaries, and reverse scenarios, and provides rigorous rebuttals and limitation analysis. \\
\midrule
Comparative perspective & 30\% &
0: No comparative or hedging perspective at all.\\
& &
1: Mentions the name of a comparison object only, but provides no comparative logic.\\
& &
2: Some comparison exists, but it remains superficial (e.g., ``higher/lower'') and lacks explanation.\\
& &
3: Provides valid comparative dimensions (e.g., peers, cycles, strategic differences), and the logic is basically clear.\\
& &
4: Analyzes multiple comparative dimensions (e.g., structural, cyclical, hedging strategy); conclusions are reasonable and fairly deep.\\
& &
5: Demonstrates professional multi-angle comparison; identifies structural differences, behavioral differences, and risk-hedging paths, and provides a rigorous explanatory framework. \\
\midrule
Argument balance & 30\% &
0: Obviously biased or one-sided; no balance at all.
1: Presents only a single viewpoint; no counterarguments or neutrality.\\
& &
2: Mentions other viewpoints but does not develop them; balance is insufficient.\\
& &
3: Presents both positive and negative sides; maintains basic objectivity.\\
& &
4: Structurally presents multiple lines of argument; strongly reflects neutrality, balance, and reasonable trade-offs.\\
& &
5: Reaches analyst-level balance; weighs how conclusions change under different scenarios or assumptions, and provides clear conditional judgments. \\
\bottomrule
\end{tabularx}
\end{table}

\begin{table}[htbp]
\centering
\caption{Category-specific scoring rubric for Architectural richness and strictness (ARS).}
\label{tab:ars_rubric}
\scriptsize
\setlength{\tabcolsep}{2pt}
\begin{tabularx}{\textwidth}{c c >{\raggedright\arraybackslash}X}
\toprule
\textbf{Sub-dimension} & \textbf{Weight} & \textbf{Scoring Criteria (0--5)} \\
\midrule
Framework completeness & 40\% &
0: No structure; content is chaotic; no analytical framework is presented.\\
& &
1: There are scattered points, but they are unrelated to one another; the answer lacks even a basic structure.\\
& &
2: A preliminary framework is formed, but coverage is incomplete; key analytical dimensions are missing.\\
& &
3: The framework is relatively clear and covers the main dimensions; the overall structure is understandable.\\
& &
4: The framework is complete and systematic; dimensional coverage is comprehensive; the answer presents a professional analytical structure.\\
& &
5: Reaches the framework quality of consulting or investment-research reports; multiple dimensions are systematically linked; the structure is highly rigorous with clearly differentiated functional sections. \\
\midrule
Logical hierarchy & 30\% &
0: No logical hierarchy; content is presented as an undifferentiated pile.\\
& &
1: There are simple bullet points, but the hierarchy is unclear; paragraphs lack logical connections.\\
& &
2: Some hierarchical structure exists, but progression is weak; the organization remains loose.\\
& &
3: Shows clear layering (e.g., overall-to-specific); logical relations are basically clear.\\
& &
4: The structure is rigorously layered; includes nested hierarchy (e.g., overall-part-subpart); logical progression is natural.\\
& &
5: Demonstrates professional-grade structural design; hierarchical progression and nesting are clear; the overall argument is rigorous and well organized. \\
\midrule
Expression granularity & 30\% &
0: Concepts are vague; expression is rough; there is no meaningful detail support.\\
& &
1: Expression is basically accurate, but granularity is extremely coarse; key definitions or details are missing.\\
& &
2: Conceptual expression is correct but not rigorous enough; detail coverage is limited.\\
& &
3: Granularity is moderate; key concepts are explained correctly; detail support is fairly adequate.\\
& &
4: Expression is rigorous and concept definitions are accurate; details are in place and the argument is solid.\\
& &
5: Reaches the standard of a professional analyst; concepts are defined precisely; details are rich without being excessive; reasoning is finely grained and highly reliable. \\
\bottomrule
\end{tabularx}
\end{table}

\end{document}